\documentclass[11pt]{article}

\usepackage[final]{acl}

\usepackage{times}
\usepackage{latexsym}
\usepackage{xcolor}
\usepackage{colortbl}
\usepackage{arydshln}
\usepackage{enumitem}
\usepackage{graphicx}
\usepackage{booktabs}

\usepackage[T1]{fontenc}

\usepackage[utf8]{inputenc}

\usepackage{microtype}

\usepackage{inconsolata}

\usepackage{graphicx}
\usepackage{algorithm}
\usepackage{algorithmic}
\usepackage{amsmath}
\usepackage{multirow}
\usepackage[most]{tcolorbox}
\tcbuselibrary{skins} 

\title{From Synthesis to Clinical Assistance: A Strategy-Aware Agent Framework for Autism Intervention based on Real Clinical Dataset}

\author{
\textbf{Junhong Lai\textsuperscript{1,2,5}, Shuzhong Lai\textsuperscript{1,2,7}, Yanhao Yu\textsuperscript{1,2,5}, Wanlin Chen\textsuperscript{6}} \\
\textbf{Chenyu Yan\textsuperscript{4}, Haifeng Li\textsuperscript{4}, Lin Yao\textsuperscript{1,2,3,5}\footnotemark[1], Yueming Wang\textsuperscript{2,5}} \\
\small \textsuperscript{1}MOE Frontiers Science Center for Brain and Brain-Machine Integration, Zhejiang University \\[-0.8ex]
\small \textsuperscript{2}Nanhu Brain-Computer Interface Institute \\[-0.8ex]
\small \textsuperscript{3}Department of Neurobiology, Affiliated Mental Health Center and Hangzhou Seventh People's Hospital, \\[-0.8ex]
\small Zhejiang University School of Medicine \\[-0.8ex]
\small \textsuperscript{4}Children's Hospital Zhejiang University School of Medicine \\[-0.8ex]
\small \textsuperscript{5}College of Computer Science and Technology, Zhejiang University \\[-0.8ex]
\small \textsuperscript{6}School of Medicine, Hangzhou City University \\[-0.8ex]
\small \textsuperscript{7}Polytechnic Institute, Zhejiang University \\
}

\begin{document}
\maketitle
\begin{abstract}
The development of AI-assisted Early Intensive Behavioral Intervention (EIBI) for Autism Spectrum Disorder (ASD) is severely constrained by data scarcity. Furthermore, while Applied Behavior Analysis (ABA) serves as the gold standard for clinical intervention, general-purpose Large Language Models (LLMs) struggle to strictly adhere to its standardized procedures, often resulting in interactions that are linguistically fluent but strategically inconsistent. To address these challenges, we introduce \textsc{ASDAgent}, a strategy-aware framework designed to unify high-fidelity intervention dialogue synthesis and clinical decision support. \textsc{ASDAgent} incorporates two specialized components to solve distinct problems: (i) a \textsc{DoctorAgent} equipped with an Observe-Think-Act-Correct (O-T-A-C) reasoning loop, which resolves the issue of strategy collapse in LLMs by making ABA execution explicit and controllable; and (ii) a \textsc{ChildAgent} that utilizes probabilistic behavior modeling to mitigate data homogeneity, simulating diverse and non-deterministic ASD response patterns. Experiments demonstrate that dialogues generated by \textsc{ASDAgent} closely mirror the strategy distribution of human therapists (KL divergence: 0.083). In real autism intervention, \textsc{ASDAgent} achieves nearly 80\% strategic consistency with human experts. Moreover, we show that synthetic data produced by \textsc{ASDAgent} effectively distills professional clinical knowledge into small language models (SLMs), significantly enhancing their therapeutic capabilities.

\end{abstract}

\section{Introduction}
\renewcommand{\thefootnote}{\fnsymbol{footnote}}
\footnotetext[1]{Corresponding author: lin.yao@zju.edu.cn}
\renewcommand{\thefootnote}{\arabic{footnote}}

\begin{figure}[t]
  \includegraphics[width=\linewidth]{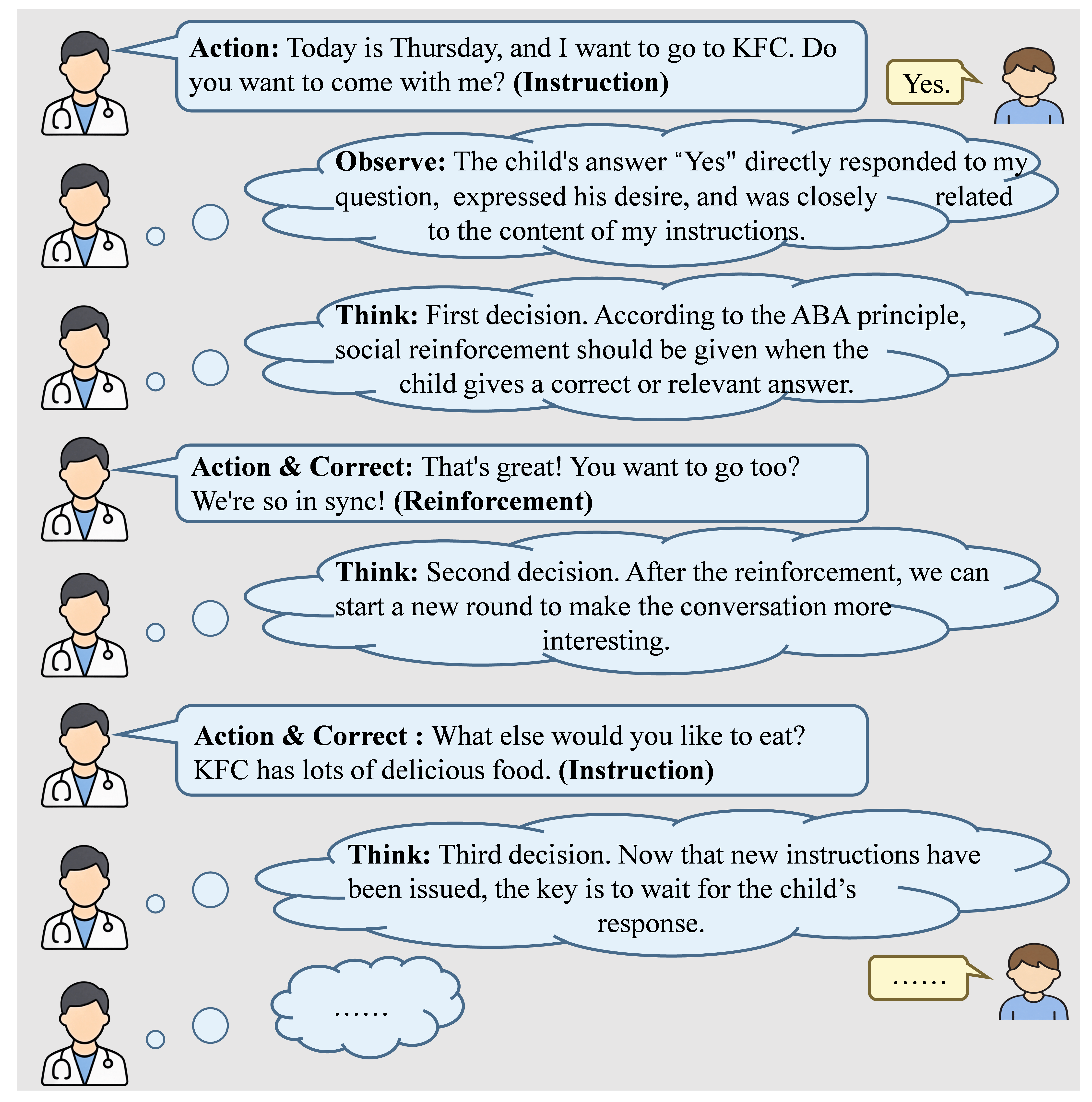}
  \caption{An example of \textsc{DoctorAgent} performing Observe-Think-Action-Correct. In the Observe phase, \textsc{DoctorAgent} categorizes and interprets the child’s responses. In the Think phase, \textsc{DoctorAgent} performs iterative, multi-round reasoning to determine appropriate intervention strategies based on the observed information. After each Think step, \textsc{DoctorAgent} immediately enters the Act and Correct phase, generating a concrete response that executes the selected strategy. This Think–Act-Correct loop may repeat multiple times within a single dialogue turn until an appropriate intervention is completed.}
  \label{fig:example}
\end{figure}
Autism Spectrum Disorder (ASD) is a pervasive neurodevelopmental disorder characterized by persistent deficits in social communication and interaction, alongside restricted, repetitive patterns of behavior, interests, or activities \cite{edition1980diagnostic}. These manifestations impose substantial impediments to social functioning, severely compromising educational attainment and daily living activities for affected individuals \cite{fuller2020effects}.

Evidence suggests that Early Intensive Behavioral Intervention (EIBI), particularly methodologies grounded in Applied Behavior Analysis (ABA) \cite{foxx2008applied, roane2016applied}, yields improved developmental outcomes (e.g., IQ, language, adaptive behavior) for many young children with ASD, although effect sizes vary and evidence quality is occasionally constrained by study design \cite{reichow2012early,virues2010applied,lovaas1987behavioral}. With the global prevalence of autism rising annually to approximately 1\% \cite{zeidan2022global}, the imperative for timely diagnosis and treatment is critical for ameliorating core symptoms \cite{estes2015long}. However, a severe global shortage of qualified providers, coupled with the prohibitive financial burden of long-term therapy, has created a widening chasm between clinical demand and service accessibility \cite{buescher2014costs,zhang2020supply}.



\begin{figure*}[t]
  \includegraphics[width=2.1\columnwidth]{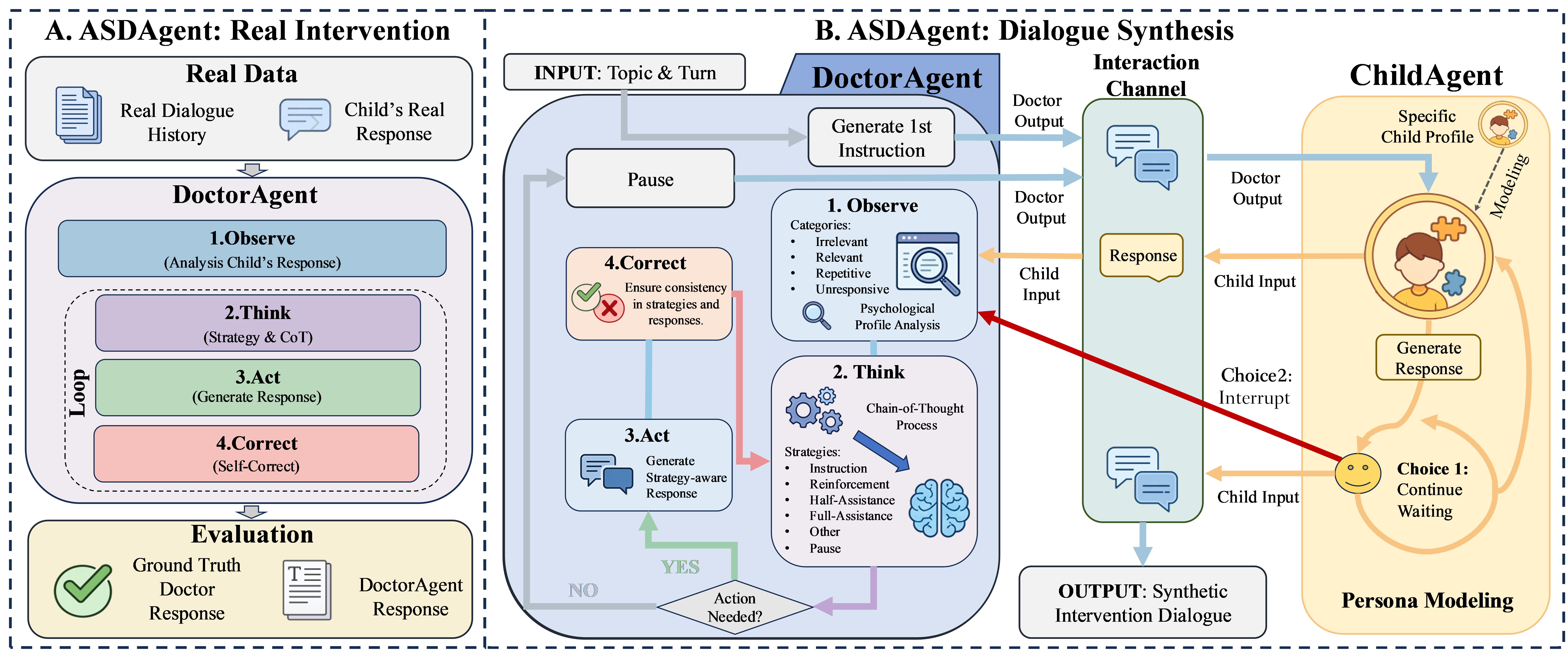}
  \caption{An overview of our framework. \textsc{ASDAgent}, for both Dialogue Synthesis and Real Autism Intervention.}
  \label{fig:overview}
\end{figure*}

Recent advancements in Large Language Models (LLMs) have catalyzed interest in AI-assisted medical diagnosis and intervention \cite{singhal2023large,nori2023capabilities,wang2025capabilities,goh2024large}. Theoretically, LLMs function as tireless "virtual therapists" or training partners. However, the direct deployment of generic state-of-the-art LLMs (e.g., GPT-4o) into ASD intervention is impeded by two critical challenges:

First, the field grapples with Data Scarcity in clinical datasets. High-quality, annotated dialogues of ASD interventions are exceedingly rare due to stringent privacy regulations and practical constraints on sharing clinical records (e.g., HIPAA requirements for protected health information and de-identification) \cite{us2005other,hhs_hipaa_deid}, which limits the development of specialized AI assistants.  Unlike general domains where data is abundant\cite{chapman2011overcoming}, the absence of large-scale clinical transcripts prevents models from learning the complex, implicit logic of professional intervention\cite{mandal2025towards}. As a result, current systems often fail to address the heterogeneous needs of the ASD population\cite{lombardo2019big}, relying instead on generic conversational patterns that lack therapeutic utility\cite{scholich2025comparison, abrams2025using}.

Second, generic models lack Explicit Strategic Reasoning. Effective ABA intervention transcends mere "chatting"; it mandates strict adherence to evidence-based instructional protocols (e.g., Discrete Trial Training, DTT) and transparent control over prompting, reinforcement, and error-correction \cite{baer1968some,smith2001discrete}. Conversely, instruction-tuned generic LLMs often exhibit sycophancy—excessively aligning with a user’s stated beliefs even when factually incorrect—leading to clinically inappropriate over-compliance \cite{sharma2023towards,perez2023discovering}. Moreover, hallucinations remain a well-documented failure mode \cite{huang2025survey}; the generation of false content poses severe ethical and safety risks in real-world clinical scenarios \cite{haltaufderheide2024ethics}.

To address these challenges, we introduce \textsc{ASDAgent}, a \textbf{Strategy-Aware} Agent Framework \textsc{ASDAgent}  integrates \textsc{DoctorAgent} with \textsc{ChildAgent} to close the loop between dialogue synthesis and strategy-aware autism intervention. Our contributions are summarized as follows:
\begin{itemize}
    \item \textbf{Explicit Strategic Reasoning:} We engineer the \textsc{DoctorAgent} with an explicit "Observe-Think-Act-Correct" (O-T-A-C) reasoning loop, inspired by ReAct \cite{yao2022react} and Reflexion \cite{shinn2023reflexion}. This mechanism enables \textsc{DoctorAgent} to transparently output the ABA strategy governing its responses. In real world autism clinical intervention, \textsc{ASDAgent} achieves a strategy consistency of nearly 80\%, representing an improvement of approximately 7\% over vanilla LLMs.
    \item \textbf{High-Fidelity Clinical Intervention Dialogue Synthesis:}  \textsc{ASDAgent} synthesizes clinical-grade dialogues that demonstrate exceptional realism, successfully confusing 89.1\% of LLM judges and 37\% of professional therapists in Turing-like tests.
\end{itemize}

\section{Related Work}

\subsection{LLMs for ASD intervention }

In recent years, the application of LLMs in ASD has expanded from simple screening to complex support systems. Researchers have explored utilizing LLMs to generate social stories for social skills training \cite{feng2025ss} and assist in assessing social reciprocity in ASD via ADOS diagnostic audio \cite{chen2025socialrecnet}. In addressing application of LLMs in autism treatment, ASD-Chat \cite{deng2024asd} employs a design paradigm integrating Verbal Behavior Milestones Assessment and Placement Program (VB-MAPP) \cite{sundberg2008vb} and ChatGPT for topic dialogue interventions, while ASD-iLLM \cite{lai2025asd} employs a fine-tuned LLM to provide dialogue intervention therapy for ASD children .

\subsection{Strategic Reasoning in Medical Agents}

The evolution of LLMs in healthcare is shifting from passive knowledge retrieval to Agentic AI—systems \cite{wang2025survey} capable of autonomous planning, reasoning, and tool use. To overcome the "black box" nature of end-to-end generation, researchers have increasingly adopted cognitive architectures that decouple reasoning from execution. Recent frameworks such as MedAgents \cite{tang2024medagents} demonstrate how multi-disciplinary collaboration and explicit reasoning steps can significantly enhance LLM proficiency in complex clinical tasks. Similarly, prompt engineering techniques like Chain-of-Thought (CoT) \cite{wei2022chain} and Tree of Thoughts (ToT) \cite{yao2023tree} have been successfully adapted to enable agents to "think before speaking," allowing for deliberate decision-making and strategic lookahead in diagnostic scenarios. In the mental health domain, specific frameworks like LLM4CBT \cite{kim2025aligning} have been proposed to align LLMs with Cognitive Behavioral Therapy (CBT) protocols, using internal "reflection" steps to ensure therapeutic adherence.

\section{Methodology}

We propose \textbf{\textsc{ASDAgent}}, a Strategy-Aware Agent framework designed to unify dialogue synthesis and clinical assistance tasks in ASD intervention. As shown in Figure \ref{fig:overview}, the framework consists of two core modules:


\begin{itemize}
    \item \textbf{\textsc{DoctorAgent}}. A doctor agent with an O-T-A-C mechanism, serving as the core intelligence for executing professional ABA interventions.
    \item \textbf{\textsc{ChildAgent}}. A data-driven child simulator based on personalized persona modeling.
\end{itemize}

\subsection{\textsc{DoctorAgent}: A Strategy-Aware Intervention Agent}
The \textsc{DoctorAgent} serves as the core strategy-making entity, executing professional ABA-based interventions through a structured O-T-A-C mechanism, ensuring that every response is clinically grounded and contextually appropriate. Unlike vanilla LLM that generate a single response in one pass, \textsc{DoctorAgent} employs an iterative decision loop, allowing it to execute a sequence of strategic actions (e.g., \textit{Reinforcement} followed by \textit{Instruction}) within a single turn until a termination condition is met.









\subsubsection{Observe}
Firstly, \textsc{DoctorAgent} analyzes the child's response $r_{child}$ to understand their behavioral state. $O_t$ is a structured observation containing Response Type and Related Analysis:


\begin{equation}
    O_t = \text{LLM}_{\text{observe}}(H_t, r_{child}, T \mid \mathcal{I}_{\text{observe}})
\end{equation}

Here, inputs including Dialogue history $H_t$, current topic $T$, the child's latest response $r_{child}$ and prompt $\mathcal{I}_{\text{observe}}$.


\subsubsection{The Loop (Think-Act-Correct)}

\textbf{Think.} At each step $k$, \textsc{DoctorAgent} decides the next immediate strategy $S_k$ and relevant CoT $C_t$ based on the observation $O_t$ and the sequence of actions already taken in this loop ($\mathcal{\pi}_{past} = \{S_1, \dots, S_{k-1}\}$):

\begin{equation}
    (S_t, C_t) = \text{LLM}_{\text{think}}(O_t, H_t, \mathcal{\pi}_{past} \mid \mathcal{I}_{\text{think}})
\end{equation}

Strategy Selection. \textsc{DoctorAgent} selects a strategy $S_t \in \mathcal{S}$ from a predefined set of ABA strategies:


\begin{equation}
\mathcal{S} =
\left\{
\begin{aligned}
&\textit{Instruction}, \textit{Other},  \textit{Full-Assistance},\\
&\textit{Half-Assistance}, \textit{Reinforcement}, \textit{Pause}
\end{aligned}
\right\}
\end{equation}


CoT. To mimic the cognitive process of a professional therapist and ensure decision transparency, we design a structured CoT prompt that guides the \textsc{DoctorAgent} through a four-stage reasoning process $C_t$  before generating any output as illustrated in Figure \ref{fig:PROMPTS FOR DoctorAgent Think}.

Termination Condition. The loop continues until the \textit{Pause} strategy is selected. This usually occurs when \textsc{DoctorAgent} determines it is time to wait for the child's response.

Constraint. If $S_{k-1}$ is \textit{Instruction}, then $S_k$ is forced to be \textit{Pause} to avoid "Instruction Stacking". In addition, $S_k$ cannot be the same as one of the previous strateies $\mathcal{\pi}_{past}$.

\textbf{Act.} Once a non-\textit{Pause}  strategy $S_k$ is selected, \textsc{DoctorAgent} generates the corresponding textual content $A_k$. We employ strategy-specific prompting in Appendix \ref{appendix:PROMPTS FOR DoctorAgent Act}, dynamically selecting a prompt template $\mathcal{I}^{S_k}_{act}$ tailored to the strategy.

\begin{equation}
    A_k = \text{LLM}_{\text{act}}(S_k, H_t \mid \mathcal{I}^{S_k}_{act})
\end{equation}

\textbf{Correct.} \textsc{DoctorAgent} sometimes makes mistakes. To prevent hallucinations where the generated text $A_k$ might drift into other strategies, we apply a self-correction filter, which decomposes $A_k$ into strategy-tagged segments and retains only segments matching $S_k$:

\begin{equation}
    R^{(k)} = \text{LLM}_{\text{correct}}(A_k, S_k \mid \mathcal{I}_{correct})
\end{equation}

This ensures that each component of the final response is pure and clinically precise.



    
    
    
    


\subsection{\textsc{ChildAgent}: Data-Driven Personalized Simulator}
To provide a realistic and diverse intervention  environment for the \textsc{DoctorAgent}, we construct a Data-Driven Child Simulator. Unlike rule-based simulators that follow rigid scripts, our Child Agent is modeled as a probabilistic state machine, where the transition probabilities are derived from real clinical data. 



\subsubsection{Probabilistic Behavioral Modeling}
\textbf{Response Modeling.} We model the child's response $r_t$ at turn $t$ as a sampling process from a categorical distribution conditioned on the interaction history. The core of this model is the Response Type Distribution, denoted as $P(R_t \mid H_t, S_{doc})$, where $R_t \in \{ \textit{Relevant, Irrelevant, UnResponsive, Repetitive} \}$ and $S_{doc}$ is the doctor's strategy at turn $t$.

To capture the sequential dependency characteristic of ASD interactions, we utilize N-gram Transition Matrices including $P_{seq}$ and $P_{last}$.

Sequential Probability $P_{seq}$. Modeling the probability based on the sequence of doctor's strategies:

\begin{equation}
    P_{seq}(r \mid \mathbf{s}_{t-k:t}) \approx \frac{Count(\mathbf{s}_{t-k:t}, r)}{Count(\mathbf{s}_{t-k:t})}
\end{equation}

where $\mathbf{s}_{t-k:t}$ is the sequence of the last $k$ strategies.

Last-Turn Probability $P_{last}$. Modeling the immediate reaction to the doctor's latest action:

\begin{equation}
    P_{last}(r \mid s_t) \approx \frac{Count(s_t, r)}{Count(s_t)}
\end{equation}

\textbf{The Interruption Mechanism.} A defining characteristic of diverse ASD phenotypes is the variance in impulse control. While some children are passive who requiring prompts to speak, others are hyper-active and prone to interrupting the therapist. 

To capture the diverse initiative patterns of ASD children, we explicitly model the Interruption Probability $P_{\text{int}}$. This measures the likelihood of the child initiating a turn immediately after the doctor executes a non-directive strategy, where a response is not explicitly demanded.

Let $\mathcal{S}_{nd} = \{ \textit{Reinforcement}, \textit{Other} \}$ denote the set of non-directive strategies. Let $s_t$ be the doctor's strategy at turn $t$, and $I_{t+1}$ denote the event whether the child speaks at turn $t+1$ (\textit{Interruption}). The probability is estimated as:

\begin{equation}
    P_{\text{int}}(I_{t+1} \mid s_t \in \mathcal{S}_{nd}) \approx \frac{\sum_{s \in \mathcal{S}_{nd}} Count(s, I_{t+1})}{\sum_{s \in \mathcal{S}_{nd}} Count(s)}
\end{equation}

\subsubsection{Personalized Parameter Blending}
A key challenge in modeling specific ASD children is data sparsity—an individual child's historical data might not cover all possible interaction scenarios. To address this, we propose a Personal-Global Blending Mechanism.

Let $\theta_{personal}$ be the probability distribution derived from a specific child's profile, and $\theta_{global}$ be the distribution derived from all real-world data. The final response distribution $\theta_{final}$ is computed as a weighted interpolation:

\begin{equation}
    \theta_{final}(r) = (1 - \alpha) \cdot \theta_{personal}(r) + \alpha \cdot \theta_{global}(r)
\end{equation}

where $\alpha \in [0, 1]$ is a smoothing factor. 

\subsubsection{Child Response Generation}
\textbf{The Interruption Mechanism.} When each doctor completes the action procedure during their turn $t$,the \textsc{ChildAgent} samples a Bernoulli variable $I_t \sim \text{Bernoulli}(P_{\text{int}}(c))$.

If $I_t = 1$, \textsc{ChildAgent} interrupts the conversation and immediately samples a response type probabilistically, generates a consistent response, and inserts it into the dialogue flow, forcing the \textsc{DoctorAgent} to handle the interruption in the next turn of the conversation. Otherwise, \textsc{ChildAgent} waits for the \textsc{DoctorAgent}'s cue.


\textbf{Response Generation.} Once the response type $y_t \in R_t$ is sampled from $\theta_{final}$, \textsc{ChildAgent} generates the textual content. We employ type-specific prompting to ensure the generated text matches the sampled response type in Appendix \ref{appendix:PROMPTS FOR ChildAgent Act}.

\begin{equation}
    R^\text{c}_t = \text{LLM}_{\text{gen}}(y_t, \text{Profile}_c, \text{T} \mid \mathcal{I}^{y_t}_{gen})
\end{equation}

where $\mathcal{I}^{y_t}_{gen}$ is a prompt template specific to the response type $y_t$.

\section{Experiment}

\subsection{Datasets}
We created a multi-turn dialogue dataset for interventions between doctors and children with ASD, named \textbf{ASDAgent-Dataset}. We transcribed 2071 instances of multi-turn dialogues. After data cleaning, we obtained 764 high-quality, authentic multi-turn dialogues from 83 children with ASD on 10 topics, which we denote as $\mathcal{D}_{golden}$.

For more information about ASDAgent-Dataset please see the Appendix \ref{appendix:sec:dataset}.


\subsection{Experiment Instructions}
In $\mathcal{D}_{golden}$, a total of 46 dialogues were sampled from 10 different dialogue topics using stratified sampling to form the test set. For hyperparameters, we set $\alpha$ to 0.3. Detailed experiment instructions can be found in Appendix \ref{appendix:Experiment Instructions}.

\subsection{Evaluation}
To comprehensively evaluate the capabilities of our proposed \textsc{ASDAgent}, we design three evaluation: Quality of dialogue synthesis, Clinical intervention effect, Data efficacy and O-T-A-C efficacy.




\noindent\textbf{Evaluation 1: Quality of dialogue synthesis.}
This task evaluates the capacity of \textsc{ASDAgent} to autonomously generate coherent, and clinically valid intervention sessions through the interaction between \textsc{DoctorAgent} and \textsc{ChildAgent} compared to $\mathcal{D}_{golden}$. In this task, \textsc{ASDAgent} synthesizes intervention dialogues that match the dialogue topics and number of turns of the test set in $\mathcal{D}_{golden}$.

\noindent\textbf{Evaluation 2: Clinical intervention effect.}
This task evaluates the \textsc{DoctorAgent}'s utility of making strategy. Instead of interacting with \textsc{ChildAgent}, the \textsc{DoctorAgent} predicts the next intervention response given a real-world clinical context. In this task, for the test set, we use a sliding window approach to generate responses turn by turn, meaning that the \textsc{DoctorAgent} independently generates the output for the current turn based on the existing dialogue history.

\noindent\textbf{Evaluation 3: Data efficacy.}
To strictly evaluate the efficacy of our proposed dialogue synthesis framework, we conducted comparative experiments across four representative SLM families: Qwen3-4B-Instruct \cite{qwen3technicalreport}, Qwen2.5-3B-Instruct \cite{qwen2} and Hunyuan-4B-Instruct using datasets of identical size sourced from: (1) Vanilla GPT-4o ("Common"), (2) Our \textsc{ASDAgent}, and (3) Real Clinical Dialogues ("Real"). We compared their performance against the non-finetuned "Base" models on a \textbf{held-out real-world test set}. 

\noindent\textbf{Evaluation 4: O-T-A-C efficacy.}
To comprehensively evaluate the architectural necessity of the O-T-A-C framework, we conducted two specific validation setups, which focus on  computational complexity, clinical effectiveness and the necessity of the Correct Module.



\subsection{Baselines}


\noindent\textbf{Baselines with Evaluation 1}. To demonstrate that our \textsc{ASDAgent} generates higher-quality dialogue than baselines, we compare \textsc{ASDAgent} against two baseline configurations. We chose \texttt{GPT-4o} \cite{hurst2024gpt} as the backbone for dialogue synthesis. 


\noindent\textbf{Baselines with Evaluation 2}. To demonstrate the effectiveness of DoctorAgent in real-world autism interventions, we selected \texttt{ASD-iLLM} \cite{lai2025asd}, \texttt{GPT-4o-mini} and \texttt{GPT-4o} \cite{hurst2024gpt} as baselines.

\noindent\textbf{Baselines with Evaluation 4}. To demonstrate the effectiveness of O-T-A-C framework, we selected Tree-of-Thoughts \cite{yao2023tree} as baseline.



\subsection{Evaluation Metrics}
We employ various metrics for automatic, manual and LLM-based evaluation purposes. Importantly, to measure the ability of \textsc{ASDAgent} for explicit strategic reasoning, we propose a metric for strategy temporal consistency. Detailed metrics
explanations can be found in Appendix \ref{appendix:evaluate metric}.

\section{Result and Analysis}

\subsection{Quality of Dialogue Synthesis}

\textbf{Automatic Evaluation.}
Table \ref{tab:kl_js_combined} shows the KL and JS divergence to real distribution for doctor strategies and child response types.



Removing \textsc{DoctorAgent} results in a significant increase in Strategy KL divergence (0.259), indicating a severe deviation from authentic clinical protocols (e.g., strategy collapse). Similarly, removing \textsc{ChildAgent} not only yields a higher Child Response divergence (KL 0.039) but, critically, exacerbates the doctor's strategic misalignment (KL rising to 0.325). This suggests that an unrealistic child simulator fails to elicit appropriate therapeutic responses, destabilizing the interaction. In contrast, the full \textsc{ASDAgent} framework achieves the lowest divergence across all metrics (Strategy KL: 0.083, Response KL: 0.007), demonstrating that the synergistic operation of both agents best reproduces realistic clinical interaction patterns and serves as the most reliable source for high-quality synthetic dialogues.

\begin{table}[htbp]
    \centering
    \caption{KL and JS Divergence to Real Distribution for Doctor Strategies and Child Responses.}
    \label{tab:kl_js_combined}
    \resizebox{\linewidth}{!}{
    \begin{tabular}{llcccc}
        \hline
        \textbf{Doctor} & \textbf{Child} 
        & \multicolumn{2}{c}{\textbf{Doctor Strategy}} 
        & \multicolumn{2}{c}{\textbf{Child Response}} \\
        \cline{3-6}
        & & \textbf{KL} $\downarrow$ & \textbf{JS}$\downarrow$ & \textbf{KL}$\downarrow$ & \textbf{JS}$\downarrow$ \\
        \hline
        DoctorAgent & ChildAgent 
        & \textbf{0.083} & \textbf{0.019} 
        & \textbf{0.007} & \textbf{0.002} \\

        DoctorAgent & GPT-4o 
        & 0.325 & 0.072 
        & 0.039 & 0.009 \\

        GPT-4o & ChildAgent 
        & 0.259 & 0.118 
        & 0.024 & 0.006 \\
        \hline
    \end{tabular}
    }
\end{table}



\begin{figure}[htbp]
  \includegraphics[width=\columnwidth]{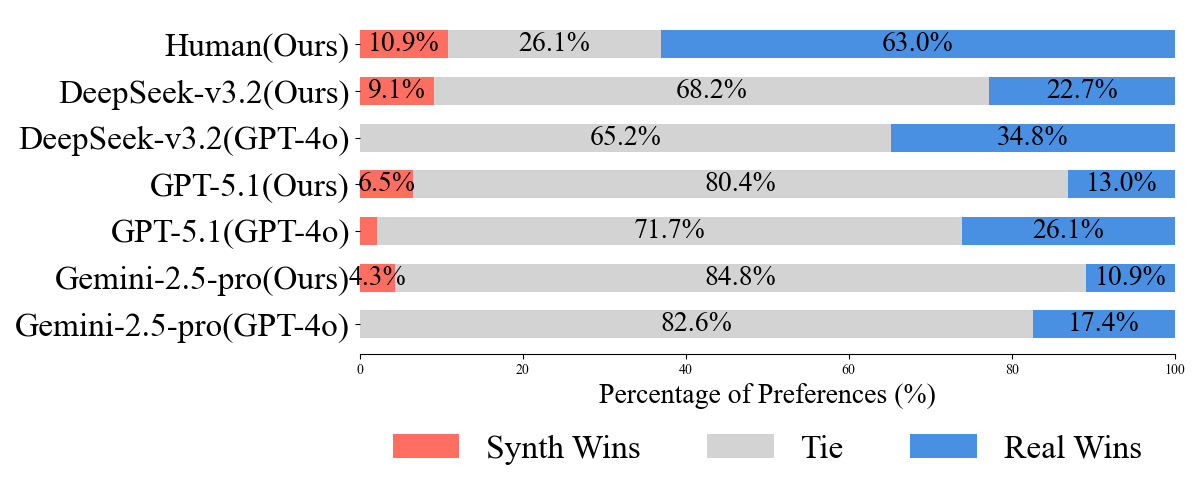}
  \caption{Human and LLM-based Preference Evaluation between Real Data and Synthetic data.}
  \label{fig:human_llm_eval_comparison}
\end{figure}

\textbf{Human and LLM Evaluation.}
We compared \textsc{ASDAgent} against a GPT-4o baseline using Turing-like preference tests. In the preference analysis (Figure \ref{fig:human_llm_eval_comparison}), notably, human experts rated \textsc{ASDAgent} as tying or surpassing real clinical sessions in 37\% of cases. Regarding automated judges, while the baseline also elicited high tie rates due to evaluator bias, it failed to secure significant win rates (e.g., 0\% with DeepSeek-v3.2). In contrast, \textsc{ASDAgent} consistently achieved higher win rates and reduced the preference for real data across all evaluators, demonstrating superior synthesis fidelity.


\begin{figure*}[htbp]
  \includegraphics[width=\linewidth]{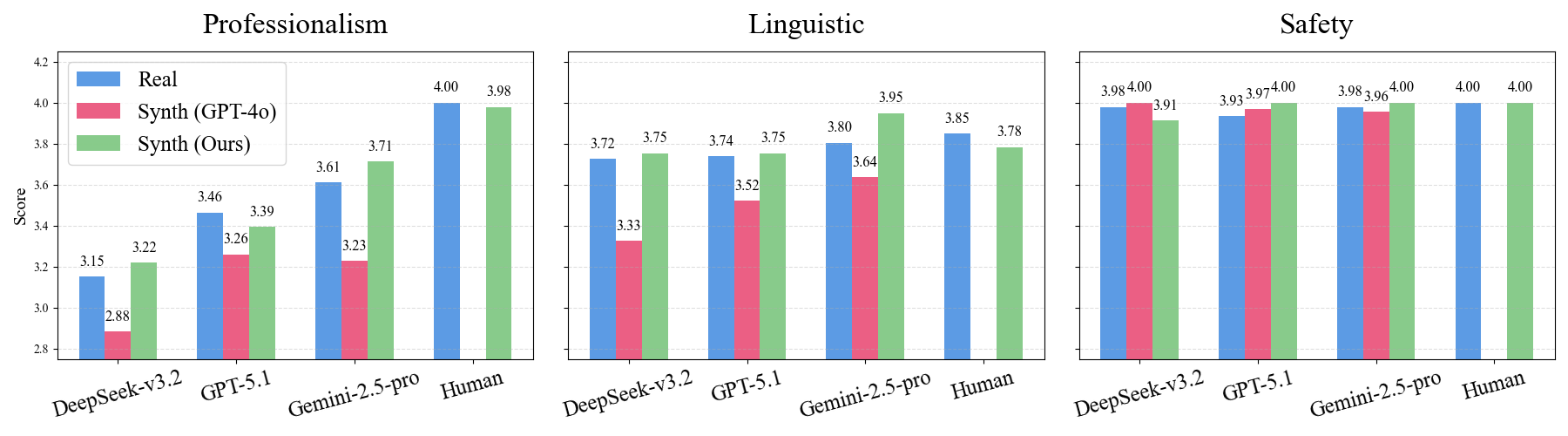}
  \caption{Human and LLM-based Scoring between Real Data and Synthetic data.}
  \label{fig:metrics_comparison}
\end{figure*}

Crucially, Figure \ref{fig:metrics_comparison} underscores the \textsc{ASDAgent}'s clinical validity, particularly in Professionalism. While the generic \textsc{GPT-4o} baseline consistently lags behind real clinical standards across automated evaluators, \textsc{ASDAgent} effectively bridges this gap. Human experts rated \textsc{ASDAgent}'s adherence to ABA protocols at 3.98/4.00, closely approximating the gold standard of real therapists (4.00). This alignment validates that the \textsc{DoctorAgent}'s explicit O-T-A-C reasoning effectively replicates professional therapeutic logic, addressing the strategic deficiencies observed in vanilla LLMs. Furthermore, \textsc{ASDAgent} maintains parity with real data in Linguistic (3.78 vs. 3.85) and Safety (4.00), demonstrating its capability to generate data that is not only textually natural but clinically rigorous.



\subsection{Clinical Intervention Effect}

\begin{figure}[htbp]
    \includegraphics[width=\linewidth]{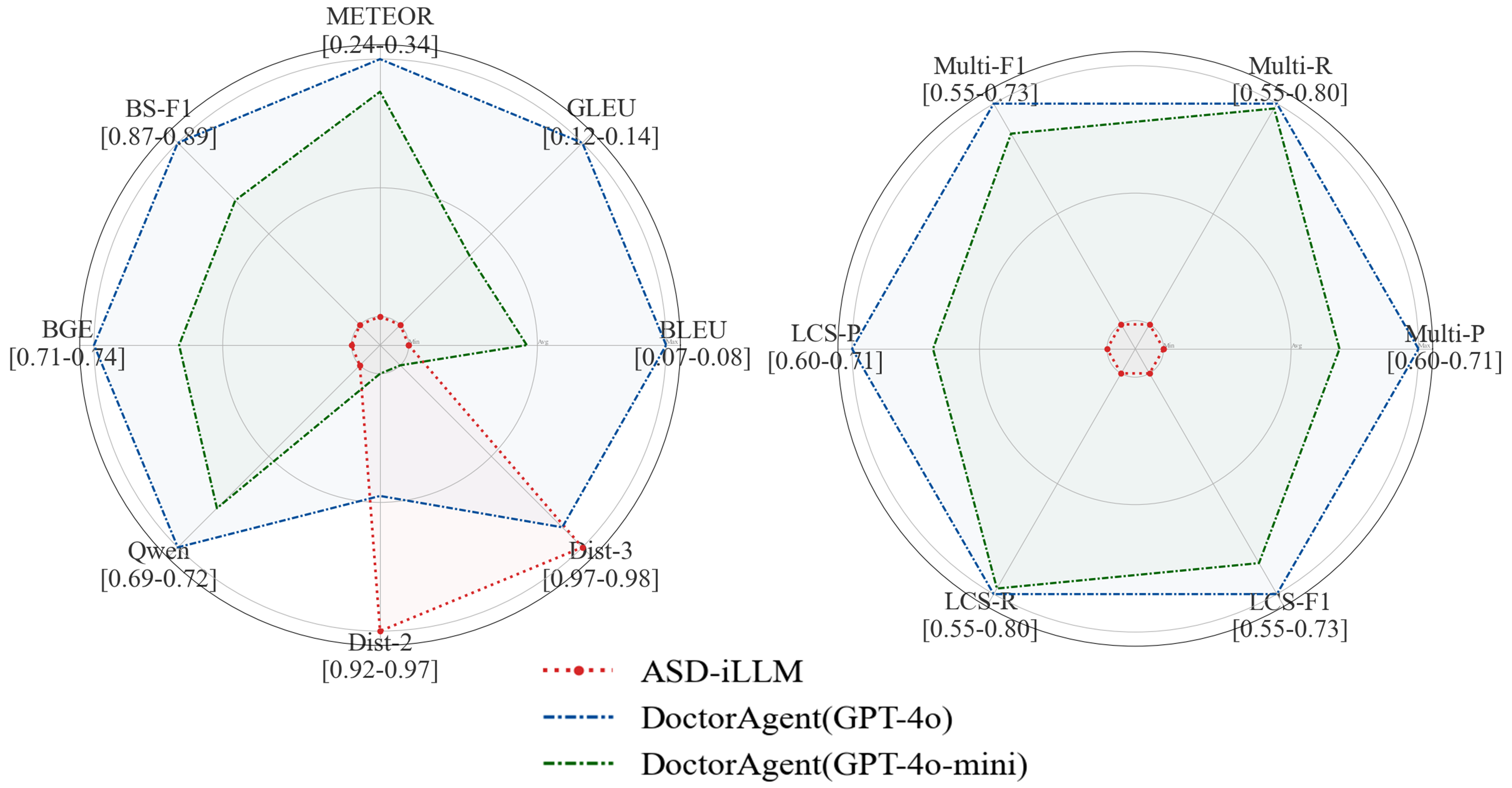}
  \caption {Evaluation on Real Intervention Dialogues. The left-hand graph shows semantic metrics, and the right-hand graph shows strategy temporal consistency.}
  \label{fig:radar}
\end{figure}

\textbf{Automatic Evaluation.} 
As shown in Figure \ref{fig:radar}, the evaluation on real intervention dialogues demonstrates that \textsc{DoctorAgent}(GPT-4o) achieves the best balance between semantic similarity and strategy temporal consistency, closely approximating real clinician behavior. \textsc{DoctorAgent}(GPT-4o-mini) provides a reasonable lightweight alternative with moderate performance degradation. In contrast, ASD-iLLM, despite exhibiting high lexical diversity, shows substantial misalignment in semantic content, strategy temporal consistency, limiting its suitability for realistic ASD intervention settings.


\begin{figure}[htbp]
  \includegraphics[width=\linewidth]{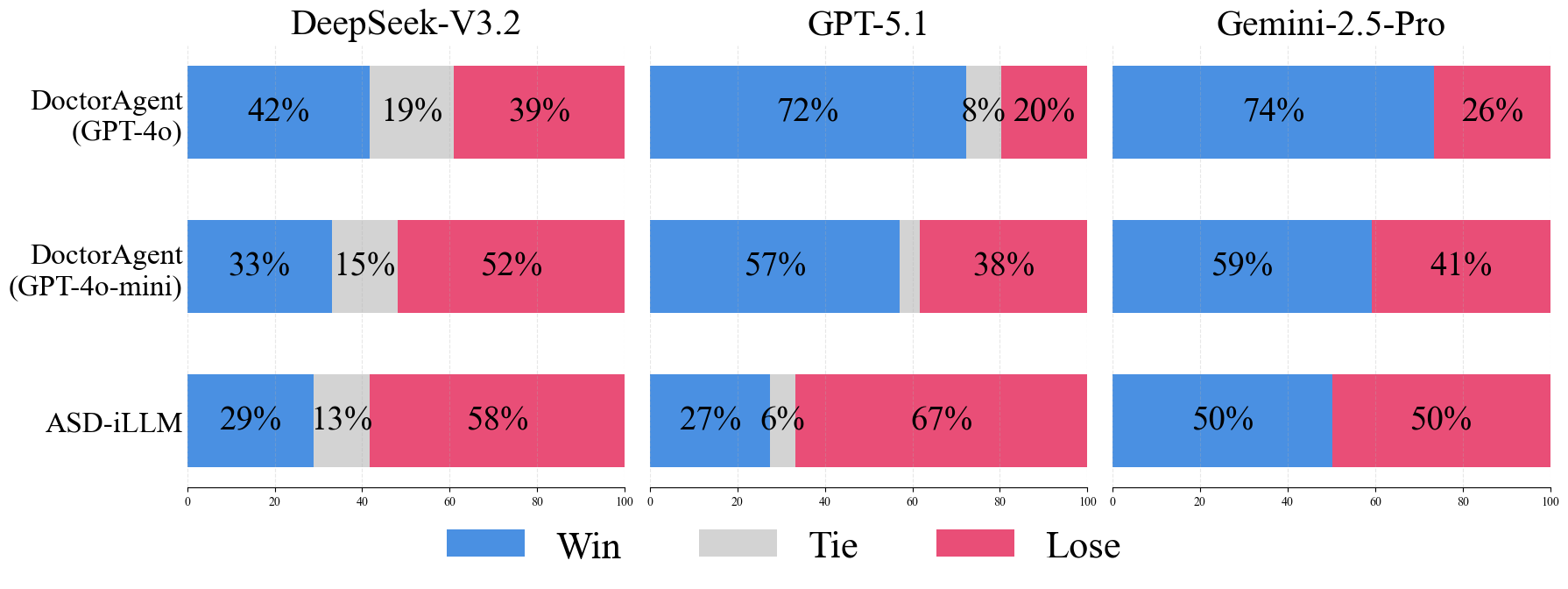}
  \caption{Win–Tie–Lose Comparison Between Model-Generated and Human Doctor Responses Across Different Evaluators.}
  \label{fig:pairwise_comparison_judges}
\end{figure}

\textbf{LLM Evaluation.} 
As shown in Figure \ref{fig:pairwise_comparison_judges}, real-world intervention dialogue assessment based on LLM showed that DoctorAgent (GPT-4o) performed best in paired comparisons with responses from real doctors during real-world dialogue interventions. DoctorAgent (GPT-4o-mini) provides a reasonable lightweight alternative, while ASD-iLLM shows substantial limitations under realistic clinical conditions.

\begin{table}[htbp]
\centering
\resizebox{\columnwidth}{!}{ 
\begin{tabular}{lccc}
\toprule
\textbf{Model} & \textbf{Correct Times} & \textbf{Total} & \textbf{Trigger Rate (\%)} \\ 
\midrule
GPT-4o-mini & $339$ & $1208$ & $\mathbf{28.06}$ \\
GPT-4o & $266$ & $1120$ & $\mathbf{23.75}$ \\
\bottomrule
\end{tabular}
}
\caption{Statistics on the triggering of the Correct phase during real-world clinical interventions.}
\label{tab:correct_ablation}
\end{table}

\subsection{Data Efficacy}
\begin{figure}[t]
  \includegraphics[width=\linewidth]{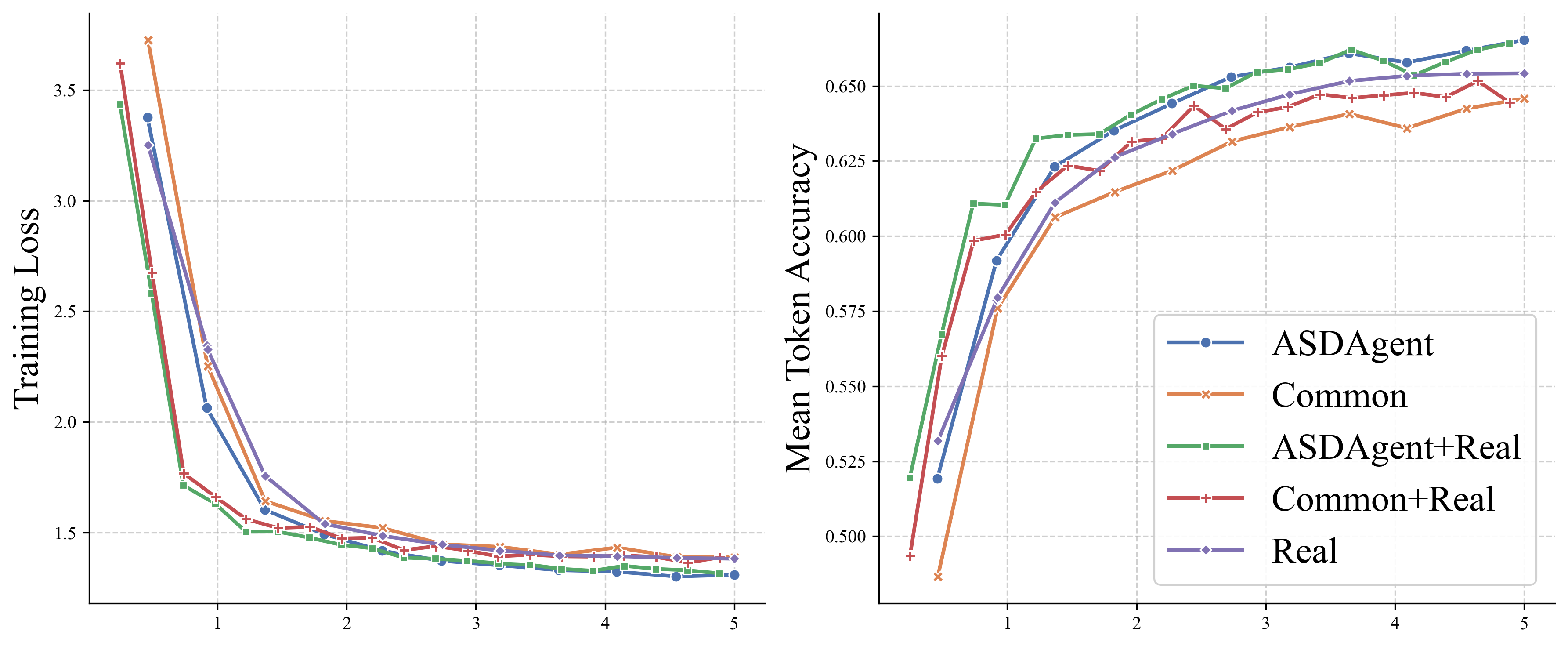}
  \caption{Training dynamics of Qwen3-4B during Supervised Fine-Tuning. (Left) Training loss convergence and (Right) mean token accuracy curves across different data sources. The x-axis represents the number of training epochs.}
  \label{fig:train_curve}
\end{figure}

As illustrated in Figure \ref{fig:train_curve}, the training trajectories reveal the superior quality and learnability of our synthesized data, particularly when utilized for data augmentation. The model fine-tuned on the augmented dataset (\textsc{ASDAgent}+Real) exhibits the most efficient convergence, maintaining the lowest training loss early in the process and achieving the highest mean token accuracy throughout the SFT process. Furthermore, even as a standalone training source, \textsc{ASDAgent} closely mirrors this augmented performance, consistently surpassing the standalone Real clinical data. Notably, both \textsc{ASDAgent} configurations significantly outperform their generic counterparts (Common and Common+Real), which suffer from slower convergence, lower accuracy, and higher final loss. This discrepancy is likely due to the stochastic noise and ``chitchat bias'' inherent in generic LLM outputs. These learning dynamics suggest that our O-T-A-C framework successfully distills the core therapeutic logic into a cleaner, more structurally consistent format. It not only substitutes scarce clinical records but also acts as a highly effective catalyst for knowledge transfer when combined with real data. 

Based on the results presented in Tables \ref{tab:sft_results} and \ref{tab:policy_alignment_detailed}, fine-tuning SLMs on data synthesized by \textsc{ASDAgent} yields superior performance across both linguistic quality and strategic alignment, consistently outperforming the generic Common baseline (vanilla GPT-4o) and effectively approaching or even exceeding the Real clinical data upper bound when used for data augmentation. Linguistically, \textsc{ASDAgent} demonstrates robust semantic fidelity; notably, when augmenting real data (\textsc{ASDAgent}+Real) on Qwen3-4B, it achieves the highest BERTScore (88.71) and BGE (75.40), surpassing both the Real data alone (88.59 and 74.97) and the Common+Real baseline (88.55 and 75.20). Most remarkably, augmenting real data with our synthetic framework breaks the strategy ceiling: on Hunyuan-4B, \textsc{ASDAgent}+Real achieves a Strategy Multi-F1 of 69.82\%, significantly outperforming the model trained on Real data alone (67.52\%). These results confirm that our framework effectively distills both the semantic nuances and the rigorous O-T-A-C therapeutic logic into deployable models, offering not only a privacy-preserving alternative to scarce clinical records but also a powerful data augmentation mechanism.

\begin{table*}[htbp]
    \centering
    \small
    \caption{Performance comparison of Small Language Models (SLMs) fine-tuned on different datasets. \textbf{Base}: Zero-shot performance. \textbf{Common}: SFT on GPT-4o synthesized data. \textbf{ASDAgent}: SFT on our synthetic data. \textbf{Real}: SFT on real clinical data. For each model, the best result is highlighted in \textbf{bold}, and the second best is \underline{underlined}.}
    \label{tab:sft_results}
    \setlength{\tabcolsep}{5pt}
    \begin{tabular}{l|lccccc}
        \hline
        \textbf{Model} & \textbf{Training Data} & \textbf{BLEU} $\uparrow$ & \textbf{GLEU}$\uparrow$ & \textbf{METEOR}$\uparrow$ & \textbf{BERTScore (F1)}$\uparrow$ & \textbf{BGE}$\uparrow$ \\
        \hline
        \multirow{6}{*}{\textbf{Qwen3-4B-Instruct}} 
        & Base & 8.04 & 11.24 & \textbf{36.44} & 87.03 & 74.71 \\
        & Common & 12.46 & 14.82 & 32.32 & 88.42 & 74.14 \\
        & ASDAgent & 12.59 & 14.98 & 33.21 & \underline{88.60} & 73.94 \\
        & Real & 14.13 & 16.07 & 34.75 & 88.59 & 74.97 \\
        & Common+Real & \underline{14.13} & \textbf{16.43} & 34.88 & 88.55 & \underline{75.20} \\
        & ASDAgent+Real & \textbf{14.47} & \underline{16.33} & \underline{35.24} & \textbf{88.71} & \textbf{75.40} \\
        \hline
        \multirow{6}{*}{\textbf{Qwen2.5-3B-Instruct}} 
        & Base & 8.88 & 12.43 & \textbf{33.78} & 88.18 & 73.56 \\
        & Common & 10.64 & 13.16 & 31.17 & 88.17 & 72.72 \\
        & ASDAgent & 10.75 & 13.68 & 31.57 & \underline{88.52} & 73.18 \\
        & Real & 11.81 & 14.04 & 31.71 & 88.06 & 73.50 \\
        & Common+Real & \underline{12.22} & \underline{14.40} & 32.53 & 88.25 & \textbf{73.91} \\
        & ASDAgent+Real & \textbf{12.29} & \textbf{14.54} & \underline{32.67} & \textbf{88.54} & \underline{73.83} \\
        \hline
        \multirow{6}{*}{\textbf{Hunyuan-4B-Instruct}} 
        & Base & 5.91 & 8.32 & 30.01 & 86.58 & 71.35 \\
        & Common & 9.97 & 13.04 & 30.96 & 88.11 & 72.49 \\
        & ASDAgent & 9.81 & 12.79 & 29.77 & \textbf{88.38} & 72.71 \\
        & Real & 11.74 & 14.18 & 31.80 & 88.15 & 73.51 \\
        & Common+Real & \underline{12.22} & \underline{14.51} & \underline{31.96} & \underline{88.32} & \underline{73.67} \\
        & ASDAgent+Real & \textbf{12.32} & \textbf{14.73} & \textbf{32.65} & 88.29 & \textbf{73.97} \\
        \hline
    \end{tabular}
\end{table*}

\begin{table*}[htbp]
\centering
\small
\caption{Strategy Alignment Analysis on Strategy Consistency Metrics (in \%). We evaluate the alignment of fine-tuned models against the ground truth strategies using Multiset (Strategy Selection) and LCS (Temporal Consistency) metrics. \textbf{Base}: Zero-shot baseline. \textbf{Common}: SFT on GPT-4o synthesized data. \textbf{ASDAgent}: SFT on our synthetic data. \textbf{Real}: SFT on real clinical data. For each model, the best result is highlighted in \textbf{bold}, and the second best is \underline{underlined}.}
\label{tab:policy_alignment_detailed}
\setlength{\tabcolsep}{4.5pt}
\begin{tabular}{l|lcccccc}
\hline
\textbf{Model} & \textbf{Training Data} & \textbf{Multi-P} $\uparrow$ & \textbf{Multi-R} $\uparrow$ & \textbf{Multi-F1} $\uparrow$ & \textbf{LCS-P} $\uparrow$ & \textbf{LCS-R} $\uparrow$ & \textbf{LCS-F1} $\uparrow$ \\
\hline
\multirow{6}{*}{\textbf{Qwen3-4B-Instruct}} 
& Base & 47.73 & 68.45 & 52.91 & 47.63 & 68.28 & 52.78 \\
& Common & 60.44 & 69.24 & 62.79 & 60.44 & 69.24 & 62.79 \\
& ASDAgent & 65.68 & 70.86 & 66.04 & 65.68 & 70.86 & 66.04 \\
& Real & \textbf{67.68} & \textbf{74.75} & \textbf{69.01} & \textbf{67.68} & \textbf{74.75} & \textbf{69.01} \\
& Common+Real & 66.23 & 73.77 & 67.82 & 66.23 & 73.77 & 67.82 \\
& ASDAgent+Real  & \underline{67.16} & \underline{74.04} & \underline{68.32} & \underline{67.16} & \underline{74.04} & \underline{68.32} \\
\hline
\multirow{6}{*}{\textbf{Qwen2.5-3B-Instruct}} 
& Base & 58.48 & 68.01 & 60.58 & 58.48 & 68.01 & 60.58 \\
& Common & 60.09 & 67.79 & 61.85 & 60.09 & 67.79 & 61.85 \\
& ASDAgent & 65.85 & 71.24 & 66.20 & 65.85 & 71.24 & 66.20 \\
& Real & \textbf{67.27} & \textbf{74.48} & \textbf{68.59} & \textbf{67.27} & \textbf{74.48} & \textbf{68.59} \\
& Common+Real & 66.30 & \underline{73.57} & \underline{67.76} & 66.30 & \underline{73.57} & \underline{67.76} \\
& ASDAgent+Real & \underline{66.56} & 72.56 & 67.44 & \underline{66.56} & 72.56 & 67.44 \\
\hline
\multirow{6}{*}{\textbf{Hunyuan-4B-Instruct}} 
& Base & 57.73 & 70.01 & 60.53 & 57.63 & 69.85 & 60.42 \\
& Common & 60.20 & 66.83 & 61.66 & 60.20 & 66.83 & 61.66 \\
& ASDAgent & \underline{66.64} & 71.46 & 66.66 & \underline{66.64} & 71.46 & 66.66 \\
& Real & 66.47 & \underline{72.92} & \underline{67.52} & 66.47 & \underline{72.92} & \underline{67.52} \\
& Common+Real & 65.19 & 72.26 & 66.68 & 65.19 & 72.26 & 66.68 \\
& ASDAgent+Real & \textbf{68.86} & \textbf{75.14} & \textbf{69.82} & \textbf{68.86} & \textbf{75.14} & \textbf{69.82} \\
\hline
\end{tabular}
\end{table*}

\subsection{O-T-A-C Architecture Analysis}

\paragraph{Comparison with ToT Baseline}
Table \ref{tab:tot_comparison} details the performance trade-offs between the ToT and O-T-A-C frameworks. While ToT's complex multi-path reasoning yields higher linguistic diversity (e.g., higher BLEU and BERTScore), it suffers from severe \textit{Strategy Collapse}. For instance, on GPT-4o, ToT's Strategy Multi-F1 drops significantly to $55.72\%$, compared to O-T-A-C's $72.95\%$. Without explicit structural constraints, ToT tends to generate overly elaborate responses that mix conflicting ABA strategies (e.g., stacking new \textit{Instructions} immediately after \textit{Reinforcement}), directly violating the ``Atomic Action'' requirement of ABA therapy. Furthermore, ToT's search mechanism introduces prohibitive latency (e.g., surging from $23.89$s to $60.24$s on GPT-4o-mini), which easily causes ASD children to lose focus and breaks the real-time therapeutic loop. Thus, O-T-A-C explicitly injects domain constraints to ensure clinical safety and strategy alignment at a fraction of the computational cost.

\begin{table*}[htbp]
\centering
\resizebox{\textwidth}{!}{
\begin{tabular}{lccccccccc}
\toprule
\textbf{Model} & \textbf{Method} & \textbf{Total Time (s) $\downarrow$} & \textbf{Avg Time (s) $\downarrow$} & \textbf{BLEU $\uparrow$} & \textbf{BERTScore F1 $\uparrow$} & \textbf{BGE $\uparrow$} & \textbf{Dist-2 $\uparrow$} & \textbf{Multi-F1 $\uparrow$} & \textbf{LCS-F1 $\uparrow$} \\
\midrule
\multirow{2}{*}{GPT-4o-mini} & Ours & $\mathbf{23.89_{\pm 13.84}}$ & $\mathbf{11.88_{\pm 5.02}}$ & $7.39$ & $88.05$ & $72.83$ & $92.49$ & $\mathbf{70.47}$ & $\mathbf{70.38}$ \\
 & ToT & $60.24_{\pm 31.43}$ & $25.86_{\pm 13.03}$ & $\mathbf{9.94}$ & $\mathbf{89.54}$ & $\mathbf{78.82}$ & $\mathbf{95.93}$ & $56.11$ & $56.03$ \\
\midrule
\multirow{2}{*}{GPT-4o} & Ours & $\mathbf{34.47_{\pm 19.49}}$ & $\mathbf{17.58_{\pm 10.12}}$ & $8.29$ & $88.60$ & $73.77$ & $\mathbf{94.54}$ & $\mathbf{72.95}$ & $\mathbf{72.95}$ \\
 & ToT & $70.29_{\pm 42.19}$ & $23.27_{\pm 10.81}$ & $\mathbf{11.88}$ & $\mathbf{89.41}$ & $\mathbf{78.90}$ & $93.80$ & $55.72$ & $55.56$ \\
\bottomrule
\end{tabular}
}
\caption{Performance and latency comparison between ToT and O-T-A-C. While ToT exhibits higher textual diversity, it severely fails in strictly adhering to clinical ABA strategies (Multi-F1 and LCS-F1) and introduces unacceptable latency compared to our framework.}
\label{tab:tot_comparison}
\end{table*}

\paragraph{Efficacy of the Correct Module}
\label{app:correct_ablation}
To evaluate the efficacy of the ``Correct'' phase, we compiled its triggering statistics during real-world interventions as shown in Table \ref{tab:correct_ablation}. The module actively modified generated responses in $23.75\%$ (GPT-4o) to $28.06\%$ (GPT-4o-mini) of dialogue turns. This intervention rate aligns with our expectations, effectively preventing \textsc{DoctorAgent} from executing strategy-inconsistent statements (e.g., inappropriately appending the \textit{Instruction} ``What else do you want?'' immediately after a \textit{Reinforcement}), which indicates that while \textsc{DoctorAgent} generally maintains consistency between the selected strategy and the generated response, the Correct module plays an indispensable role in self-filtering and correction, which acts as an adaptive safety filter, dynamically adjusting its intervention based on the generation quality to ensure strict adherence to ABA protocols to a certain extent.

\section{Conclusion}

In this work, we address two critical bottlenecks impeding the advancement of AI-assisted ASD intervention: the scarcity of clinical dialogue scenarios, and the inherent struggle of general-purpose LLMs to adhere to standardized ABA protocols. We introduce \textsc{ASDAgent}, a unified strategy-aware framework designed to simultaneously tackle high-fidelity dialogue synthesis and clinical decision support. Specifically, our framework incorporates a \textsc{DoctorAgent} that operationalizes rigorous ABA procedures via an explicit O-T-A-C reasoning loop, coupled with a probabilistic \textsc{ChildAgent} that simulates diverse, non-deterministic patient phenotypes. This multi-agent synergy establishes a robust closed-loop environment, enabling the synthesis of clinical-grade intervention dialogues that effectively distill professional therapeutic knowledge into deployable SLMs.

\section*{Limitations}

Despite the promising results demonstrated in our simulation and evaluation, several limitations should be acknowledged to contextualize our findings and guide future research.

\textbf{Absence of Real-World Clinical Validation.} First and foremost, as ASDAgent has not yet been deployed in direct clinical interventions with children diagnosed with ASD, its practical efficacy remains theoretically grounded but empirically unproven in in vivo settings.The system currently serves best as a training tool for therapists or a decision support system, rather than an autonomous intervention agent.

\textbf{Restriction to Textual Modality.} Our current framework operates exclusively within the textual modality. However, EIBI heavily relies on multimodal cues, including prosody (tone of voice), facial expressions, eye contact, and body language—factors that are critical for assessing engagement and emotional regulation in children with ASD. By relying solely on text, ASDAgent abstracts away these non-verbal signals, potentially limiting its ability to detect subtle behavioral triggers or reinforce non-verbal communication milestones.

\textbf{Simplification of Longitudinal Dynamics.} While our ChildAgent simulates session-level behaviors (e.g., turn-taking, impulsivity), it does not yet fully model the long-term developmental trajectory of a child. In real therapy, a child's skills and interests evolve over months or years. 

\section*{Ethical Considerations}

\textbf{Data Privacy and Protection.} The protection of participant privacy is paramount, particularly given the sensitive nature of clinical data involving children with ASD. Throughout the dataset construction process, we implemented a rigorous, multi-layered de-identification protocol. This involved an initial pass of automated PII (Personally Identifiable Information) scrubbing, followed by manual verification to ensure the complete removal or obfuscation of sensitive attributes, including names, locations, and institutional references. Our dataset is released strictly for non-commercial research purposes under a license that prohibits any attempt to re-identify individuals.

\textbf{Ethics of Synthetic Data Generation.} We acknowledge the ethical complexities inherent in simulating the behaviors of neurodivergent populations. A primary concern is the potential for algorithmic stereotyping, where the generative model might oversimplify ASD phenotypes into repetitive or remaining silent, ignoring the high-functioning or "masking" traits often seen in real scenarios. To mitigate this, our ChildAgent utilizes a probabilistic behavioral mechanism rather than fixed, caricature-like personas. However, users must recognize that these synthetic dialogues are statistical approximations and not substitutes for the lived experiences of real children. To ensure transparency and prevent misinformation, all synthesized data is explicitly watermarked or metadata-tagged to distinguish it from authentic clinical records.

\textbf{Clinical Applicability and Safety Scope.} While ASDAgent demonstrates high fidelity in simulating intervention scenarios, we explicitly caution against its immediate deployment in unsupervised clinical settings. The system lacks validation through longitudinal clinical trials and does not possess the legal or ethical authority to act as an autonomous therapist. Therefore, ASDAgent should be utilized strictly as a Clinical Decision Support System (CDSS) or a training simulator. Any application in a real intervention loop must adhere to a "Human-in-the-Loop" framework, where professional therapists review all AI-generated suggestions to ensure safety, efficacy, and ethical compliance.

\section*{Acknowledgements}
We thank all volunteers for their participation in the study. This work was supported in part by STI 2030—Major Projects under Grant 2021ZD0200400, in part by the National Natural Science Foundation of China under Grant 62336007, in part by the Starry Night Science Fund of Zhejiang University Shanghai Institute for Advanced Study under Grant SN-ZJU-SIAS-002, in part by the Fundamental Research Funds for the Central Universities, in part by the Project for Hangzhou Medical Disciplines of Excellence, and in part by the Key Project for Hangzhou Medical Disciplines.

\bibliography{main}

\appendix


\section{Language Clarification}
We confirm that all experiments, datasets, and model interactions were conducted in Chinese to align with the native language of the collected clinical data. The prompts and case studies presented in the paper were translated into English solely for the readability of the audience.

\section{Case Study}



\subsection{Case Study in Dialogue Synthesis}
Figure \ref{fig:qual_color_case} shows examples of real and synthetic dialogues on the same topic and with the same number of rounds.

\tcbset{
  dialogpanel/.style={
    colback=gray!12,
    colframe=black!40,
    boxrule=0.4pt,
    arc=2pt,
    left=3pt,right=3pt,top=3pt,bottom=3pt,
    enhanced,
    equal height group=A, 
  }
}

\definecolor{instrblue}{RGB}{212,222,239}   
\definecolor{assistgreen}{RGB}{219,239,196} 
\definecolor{feedbackyellow}{RGB}{253,242,205} 
\definecolor{myred}{RGB}{241,195,189}

\newcommand{\instr}[1]{\colorbox{instrblue}{\parbox{\linewidth}{#1}}}
\newcommand{\assist}[1]{\colorbox{assistgreen}{\parbox{\linewidth}{#1}}}
\newcommand{\feedback}[1]{\colorbox{feedbackyellow}{\parbox{\linewidth}{#1}}}
\newcommand{\child}[1]{\colorbox{myred}{\parbox{\linewidth}{#1}}}

\newcommand{\D}{\textbf{Doctor:}\ }
\newcommand{\C}{\textbf{Child:}\ }

\begin{figure*}[htbp]
\centering
\setlength{\tabcolsep}{6pt}

\begin{minipage}[htbp]{0.48\linewidth}
\begin{tcolorbox}[dialogpanel]
\small
\setlength{\parskip}{2pt}

\instr{\D {Shall we go buy fruit together today?} }
\child{\C [Child Unresponsive] \emph{(No Response)}}
\instr{\D Now, imagine we've walked into a fruit shop. Behind the counter stands a smiling auntie who asks, ``Little friend, which fruit would you like to take home today?'' }
\child{\C I want Coke. \emph{(Irrevelant Response)}}
\feedback{\D Coke is a beverage, but today we're here specifically to buy fruit! }
\assist{\D Look at the shelves again, is there any fruit you want to eat? }
\child{\C [Child Unresponsive] \emph{(No Response)}}
\assist{\D That's okay, let's look again—are there any other fruits you'd like to buy? }
\child{\C Grapes. \emph{(Relevant Response)}}
\feedback{\D So Dudu likes grapes. Teacher also wants to buy some fruit, like kiwis and oranges. }
\child{\C Hey, hey, hey. \emph{(Irrevelant Response)}}
\instr{\D Now it's Auntie's turn—what does Auntie want to buy? }
\child{\C What a long orange! \emph{(Irrevelant Response)}}
\instr{\D We just said we wanted to buy grapes. What kind of grapes did we buy? }
\child{\C They're in bunches. \emph{(Relevant Response)}}
\feedback{\D Right! Grapes are in bunches. }
\instr{\D What color are they? }
\child{\C Purple. \emph{(Relevant Response)}}
\feedback{\D Great! }
\instr{\D What do grapes taste like? }
\child{\C [Child Unresponsive] \emph{(No Response)}}
\assist{\D Are they sweet or sour? }
\child{\C Sweet. \emph{(Relevant Response)}}
\feedback{\D That's right! Grapes are sweet and juicy. }
\end{tcolorbox}
\end{minipage}
\hfill
\begin{minipage}[htbp]{0.48\linewidth}
\begin{tcolorbox}[dialogpanel]
\small
\setlength{\parskip}{2pt}

\instr{\D {Hello Dudu, today we're going to learn how to buy fruit. Are you ready?} }
\child{\C Mm, heyhey. \emph{(Irrevelant Response)}}
\feedback{\D Haha, Dudu, you're in a really good mood today!}
\assist{\D What fruit do you most want to buy today? }
\child{\C Apples. \emph{(Relevant Response)}}
\feedback{\D Great! Dudu wants to buy apples, that's wonderful! }
\instr{\D What color are apples?}
\child{\C Where did the little bear go? \emph{(Irrevelant Response)}}
\feedback{\D Oh, the little bear, he's probably playing outside. }
\instr{\D What color are the apples we just talked about?}
\child{\C I'm Superman. \emph{(Irrevelant Response)}}
\feedback{\D Oh, Dudu is Superman, then Superman must be very powerful, right? }
\assist{\D Are apples red or yellow?}
\child{\C They're red. \emph{(Relevant Response)}}
\feedback{\D That's right! Apples are red, Dudu answered very well!}
\instr{\D So how does Dudu like to eat apples? Raw or juiced? }
\child{\C Raw. \emph{(Relevant Response)}}
\feedback{\D Wow! Dudu likes to eat apples raw, what a big apple lover!}
\instr{\D Oh, so how does Dudu usually cut apples since he likes to eat them raw? }
\child{\C I want to play with building blocks. \emph{(Irrevelant Response)}}
\feedback{\D Wow, building blocks are fun too, I understand you want to play now. We can play with building blocks later, let's talk about apples first, okay? }
\child{\C Apples are red. \emph{(Relevant Response)}}
\feedback{\D Great answer! Apples are indeed red, that's great! }
\child{\C I want to go to the amusement park. \emph{(Irrevelant Response)}}
\feedback{\D Oh, Dudu wants to go to the amusement park! Amusement parks are so much fun!}
\assist{\D There are lots of fruit shops in the amusement park! For example, there are big, sweet watermelons there. Is there any fruit that Dudu particularly wants to eat?}
\child{\C Banana. \emph{(Relevant Response)}}
\feedback{\D Dudu, that's great! You like bananas, that's wonderful! You're a little fruit expert!  }
\instr{\D Besides bananas, what other fruits does Dudu like?}
\end{tcolorbox}
\end{minipage}

\caption{Case Study in Dialogue Synthesis with Topic "Buy fruit" and the same Turns. The left side shows a real intervention dialogue, while the right side shows a synthetic intervention dialogue. \colorbox{instrblue}{Blue} indicates Instruction, \colorbox{assistgreen}{green} denotes Assistance including Half-Assistance and Full-Assistance, \colorbox{feedbackyellow}{yellow} signifies Reinforcement and Acknowledgement, and \colorbox{myred}{red} represents the child's responses.}
\label{fig:qual_color_case}
\end{figure*}

\subsection{Case Study in Real Autism Intervention}
\begin{figure*}[htbp]
  \includegraphics[width=2.1\columnwidth]{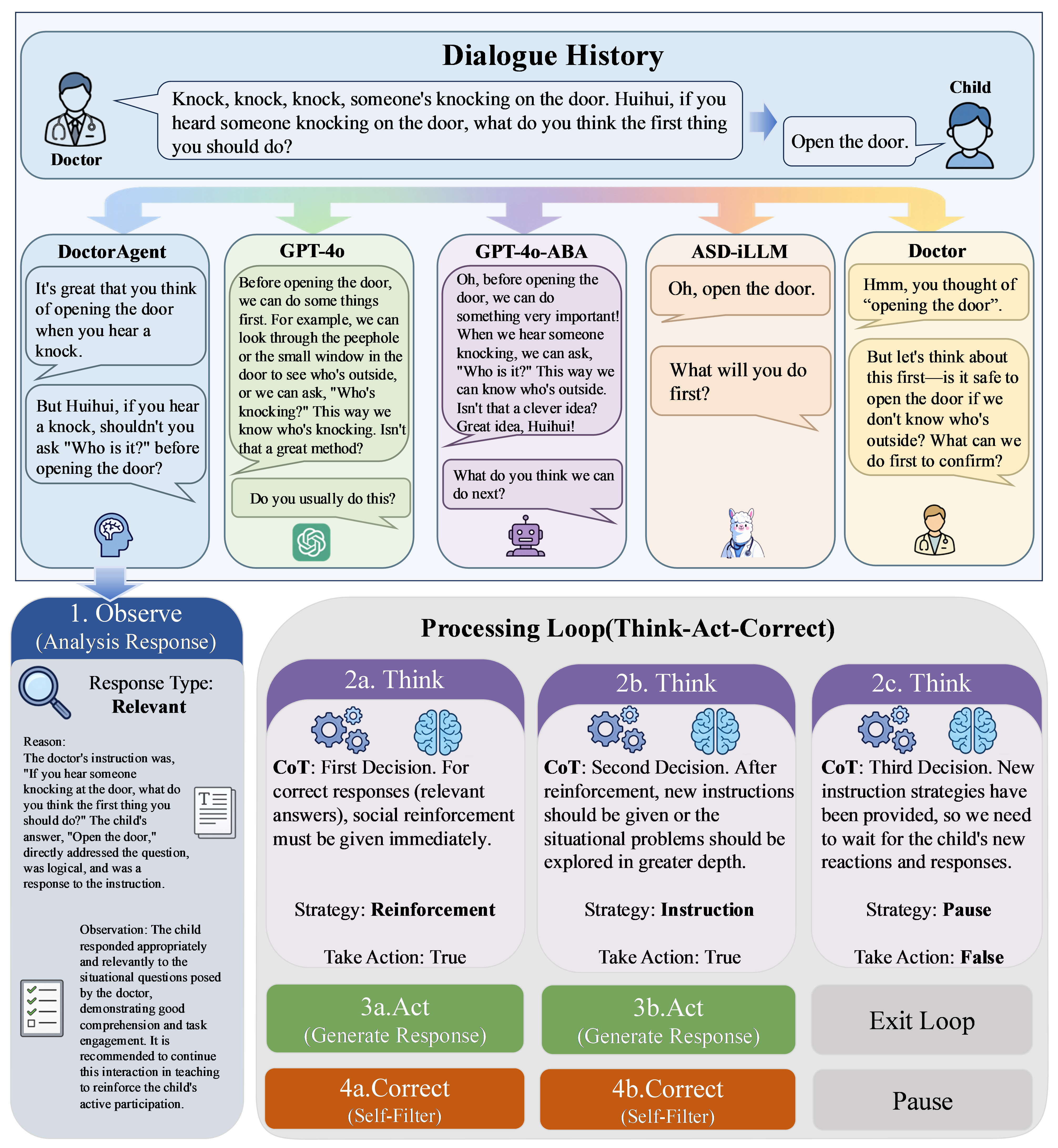}
  \caption{Case Study in Real Autism Intervention. The diagram above illustrates the intervention responses of \textsc{DoctorAgent}, the real Doctor, and other models based on a realistic intervention dialogue. The diagram below shows how \textsc{DoctorAgent} completes the O-A-T-C process.}
  \label{fig:case study3}
\end{figure*}

Figure \ref{fig:case study3} illustrates a case that the different responses of different models to real autism clinical interventions and how \textsc{DoctorAgent} performs O-T-A-C.

\subsection{Case Study in Comparison to ToT}
\begin{figure*}[t]
  \includegraphics[width=2.1\columnwidth]{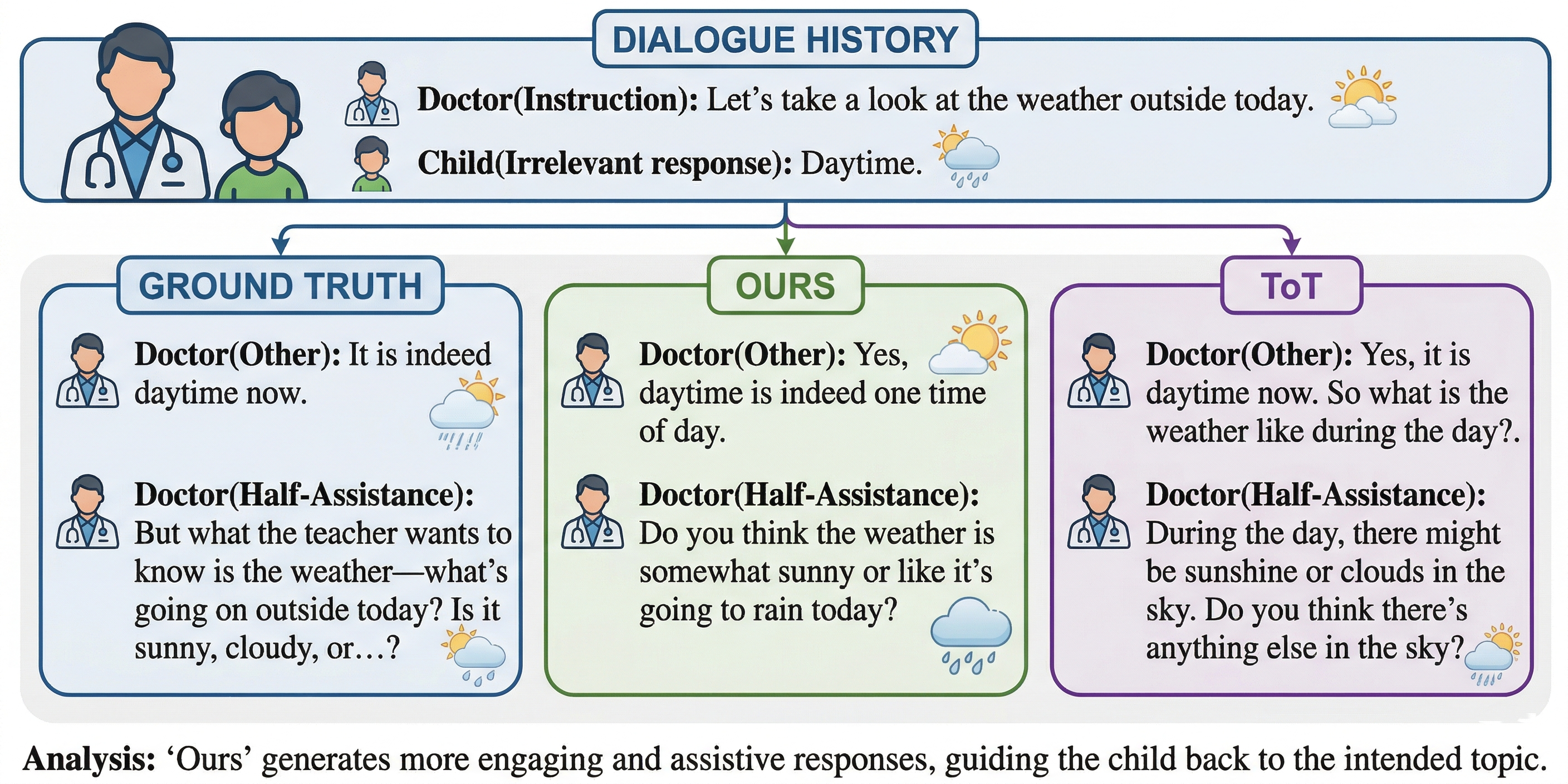}
  \caption{Qualitative comparison between our O-T-A-C framework and the Tree-of-Thought (ToT) baseline. While ToT exhibits conversational fluency, it critically fails in clinical adherence by introducing strategy deviation (instruction stacking) and topic drift. Our framework safely guides the child back to the intended topic.}
  \label{fig:case study4}
\end{figure*}

To intuitively illustrate the clinical limitations of the ToT baseline, Figure \ref{fig:case study4} presents a qualitative case study where the child provides an irrelevant response (``Daytime'') to a weather-related instruction. According to ABA protocols, the doctor should first acknowledge the child's response before providing further assistance. 

Through this case study, we observe that while both our framework and ToT generate responses that subjectively resemble the natural conversational tone of real doctors, the ToT generation suffers from two critical clinical flaws:

\begin{itemize}
    \item \textbf{Strategy Deviation:} During the initial acceptance phase (\textit{Other}), ToT inappropriately appends a new instruction (``So what is the weather like during the day?''). Acceptance should strictly acknowledge the child without immediately demanding a new cognitive task. This flaw precisely demonstrates the necessity of our \textit{Correct} module. While the foundational generation in our framework might occasionally make similar instruction-stacking errors, the O-T-A-C loop effectively detects and filters them out, ensuring the atomicity of the strategy.
    
    \item \textbf{Topic Deviation:} The doctor's initial target concept was ``weather.'' However, when executing the \textit{Half-Assistance} intervention, the ToT response drifts from the core topic, instead asking an open-ended, vague question (``Do you think there's anything else in the sky?''). Such topic deviation easily distracts ASD children and strictly violates the precise targeting requirements of clinical EIBI interventions. Conversely, our framework generates a highly engaging and assistive prompt (``sunny or like it's going to rain'') that safely guides the child back to the intended topic.
\end{itemize}

\subsection{Case Study in Correct Module}
\begin{figure*}[t]
  \includegraphics[width=2.1\columnwidth]{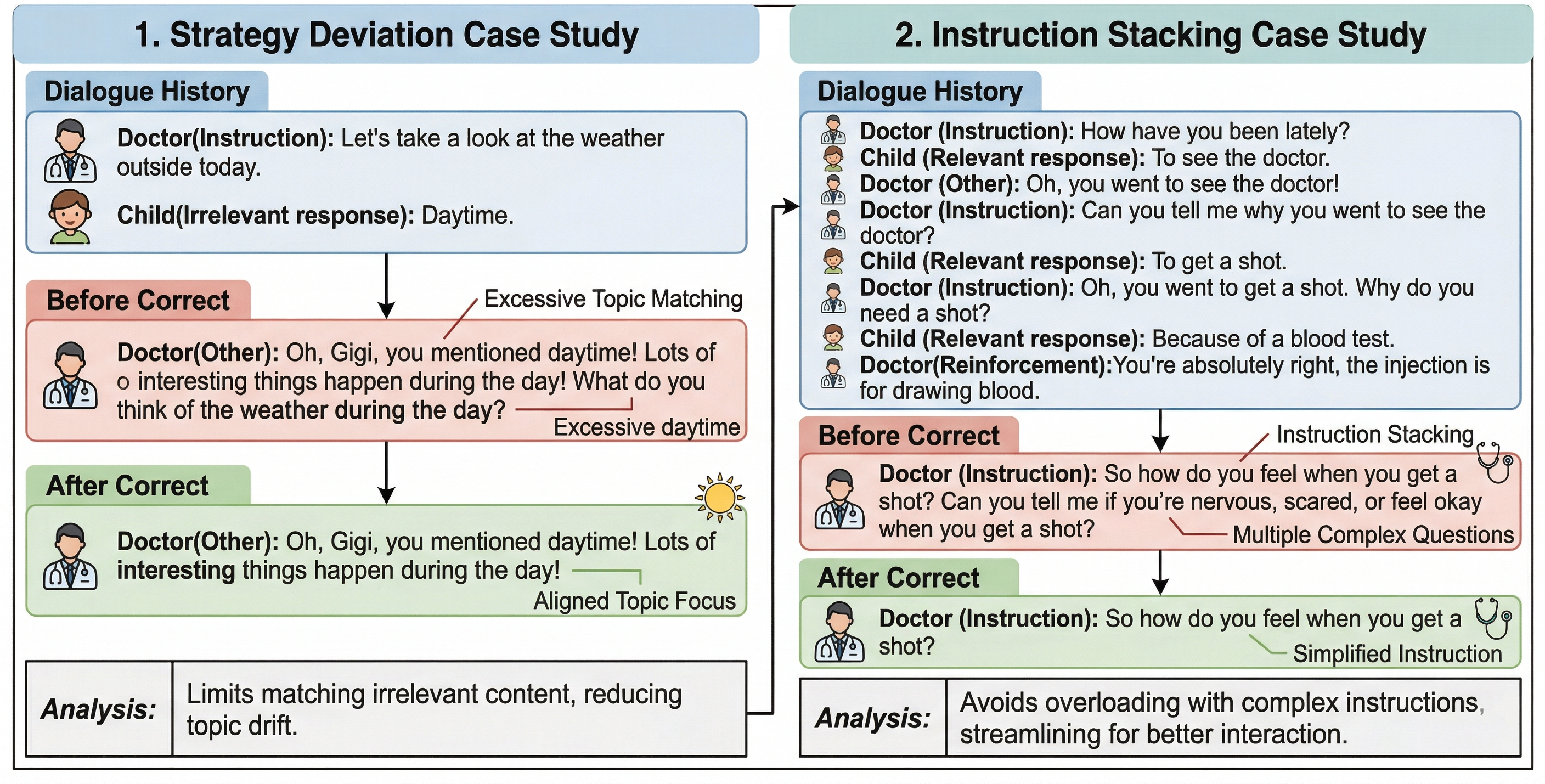}
  \caption{Qualitative case studies demonstrating the efficacy of the \textit{Correct} module. The module actively rectifies strategy deviation by preventing topic drift (left) and resolves instruction stacking by streamlining complex, overloaded questions into clinically appropriate atomic instructions (right).}
  \label{fig:case study5}
\end{figure*}

To intuitively illustrate the impact of the \textit{Correct} module, Figure \ref{fig:case study5} presents two typical intervention scenarios. In the \textbf{Strategy Deviation} case (Figure \ref{fig:case study5}, left), the child provides a seemingly irrelevant but tangentially related response (``Daytime''). While the doctor appropriately attempts to acknowledge this, the uncorrected generation improperly appends a new instruction (``What do you think of the weather during the day?'') during the acceptance phase. The \textit{Correct} module successfully excises this excessive topic matching, ensuring the response aligns with the intended strategy without causing topic drift.

Furthermore, the \textbf{Instruction Stacking} case (Figure \ref{fig:case study5}, right) highlights a critical clinical safety mechanism. Before correction, the agent stacks multiple complex questions into a single conversational turn. Given the severe social and cognitive communication difficulties faced by autistic children, ABA intervention protocols strictly prohibit delivering complex or consecutive instructions, which can easily cause cognitive overload and break the therapeutic loop. The \textit{Correct} module effectively filters out the redundant questions, streamlining the utterance into a single, atomic instruction that is clinically safe and manageable for the child to process.

\section{ABA Strategy and Response Type}
\subsection{ABA Strategy}
ABA is a structured approach commonly used as a behavioral therapy in treating autism . Specifically, doctors integrate Discrete Trial Teaching (DTT) and Natural Environment Teaching (NET) methods from ABA to intervene with autistic children. The doctor's strategies are categorized as: \textit{Instruction}, \textit{Reinforcement}, \textit{Half-Assistance}, \textit{Full-Assistance}, \textit{Other} and \textit{Pause}. The child's response types are categorized as: \textit{Irrelevant}, \textit{Relevant}, \textit{Repetitive}, and \textit{Unresponsive}. 

\textbf{Instruction} are issued by the doctor, who ensures they are concise and easy for the child to comprehend. Through these instructions, the doctor guides the child in understanding language and learning social skills.

\textbf{Reinforcement} involves providing stimuli when a child responds to an instruction. The purpose of reinforcement is to encourage the continued occurrence of appropriate behaviors, while inappropriate behaviors diminish or disappear due to a lack of reinforcement. Reinforcement can be physiological, such as favorite foods or toys, or social, such as praise. In social dialogue interventions, we emphasize social reinforcement, enhancing the child's socialization through verbal praise and empathy.

\textbf{Assistance} refers to the support provided by therapists when autistic children have difficulty responding. This support can take the form of physical, visual, or verbal prompts.
Assistance helps children build confidence, reduce frustration, and gradually understand the meaning of instructions. Assistance needs to be timely and appropriate to avoid causing feelings of failure in the child or creating dependence on the prompts. In thematic conversation intervention, Assistance usually takes the form of verbal prompts, such as rephrasing questions, breaking down questions, or providing hints to the answer.

Assistance can be further categorized into Half-Assistance and Full-Assistance. 

{Half-Assistance} refers to providing limited hints, such as keyword reminders, selective prompts, or guiding questions, when the child already has some understanding or a tendency to respond, helping the child complete the response based on their existing understanding. 

\textbf{Full-Assistance}, on the other hand, involves the therapist directly providing clear demonstrations or complete answers when the child cannot understand the instructions or shows no response, guiding the child to imitate or repeat the correct response. By flexibly using partial and Full-Assistance at different stages, therapists can ensure the success rate of the intervention while gradually improving the child's independent response ability.

\textbf{Pause} refers to the brief interval between each trial, allowing the child time to reflect on and internalize their response and the reinforcement.

\subsection{Child Response Type}
\label{appendix:sec:dataset:child response type}
\textbf{Relevant} responses refer to children's answers that semantically or functionally match the instructions or questions given by the doctor, indicating that the child understands the current topic and can respond appropriately;

\textbf{Irrelevant} responses refer to children's answers that have no clear connection to the current instructions or topic, possibly reflecting attention shifts, comprehension difficulties, or language organization problems;

\textbf{Repetitive} responses refer to children simply repeating the doctor's words or their own previous expressions without providing new information or independent responses, usually reflecting imitative behavior or limitations in response strategies;

\textbf{Unresponsive} responses refers to the child not giving any verbal or non-verbal response within a reasonable waiting time, which may be related to comprehension difficulties, avoidance behavior, or emotional state.

\section{Details for ASDAgent-Dataset}
\label{appendix:sec:dataset}
Currently, there are no publicly available datasets for ASD dialogue intervention. Therefore, we created a multi-turn dialogue dataset for interventions between doctors and children with ASD, named \textbf{ASDAgent-Dataset}.


\subsection{Data Collection}
To ensure the authenticity and quality of the data, we collaborated with five treatment centers for autistic children after obtaining ethical approval. With full informed consent from both parents and children, audio recordings were collected during topic-based dialogue interventions using a portable recording device (H1-Pro, iFlytek Inc., China). To ensure clear audio capture, the recorder was placed in the chest pocket of the doctor’s coat.

Given that autistic children often experience delays in language development, chronological age does not necessarily reflect actual language ability. Therefore, only children with a language developmental age greater than 24 months were included in the study. Previous studies have shown that topic-based dialogue interventions can effectively alleviate social impairments in autistic children \cite{dekker2019social, hanrahan2020pilot, van2022dialogic}. Accordingly, all recordings were conducted in the form of structured topic dialogues, with each recording focusing on a single predefined topic. All audio recordings were sampled at 16,000 Hz and stored in WAV format.

\subsection{Data Processing}
We employed a three-stage processing method to transcribe the original audio recordings into multi-dialogue text and annotate the doctors' strategies and the children's response types.

\textbf{Automatic Transcription} First, we utilized existing automated transcription tools SEED-ASR\cite{bai2024seed} to convert the original recordings into multi-turn dialogues.

\textbf{Manual Transcription} Our goal is to improve the quality of multi-turn dialogue text through manual transcription. Building upon \cite{lai2025asd}, we annotated the data using crowdsourcing. Details about crowdsourcing can be found in the Appendix \ref{appendix:Crowdsourcing}.

\begin{figure}[htbp]
  \includegraphics[width=\columnwidth]{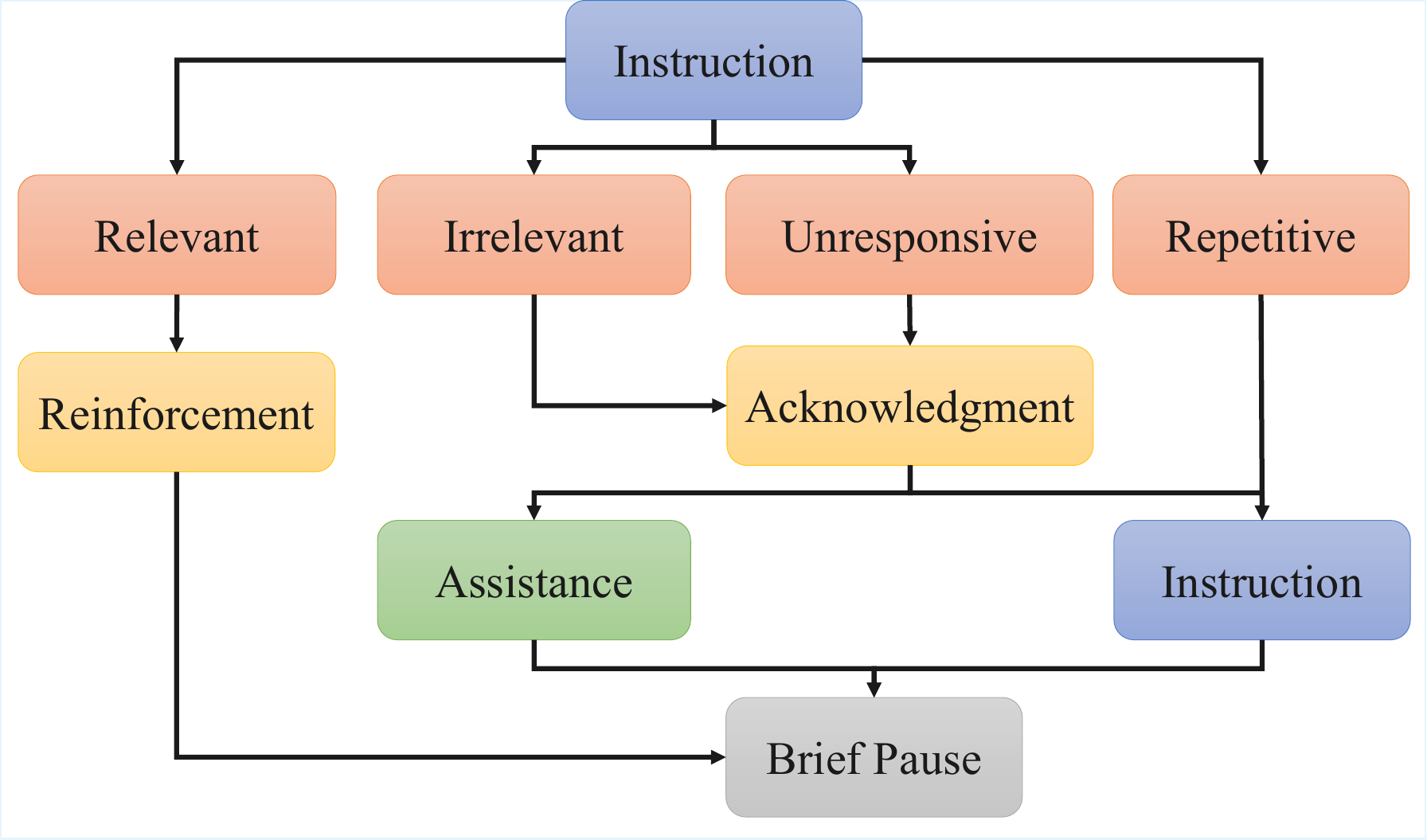}
  \caption{The standard workflow of Discrete Trial Training (DTT) derived from ABA literature \cite{packetdiscrete}, illustrating the structured interaction cycle. Doctors can adjust their treatment strategies as needed, based on the actual intervention situation.}
  \label{fig:DTT}
\end{figure}

\textbf{State Annotation} According to the ABA\cite{foxx2008applied, roane2016applied}, we performed more detailed data annotation on the selected high-quality dialogues, including annotating the doctor's strategies and the child's response types using ABA and DTT. The basic flow of DTT is illustrated in Figure \ref{fig:DTT}. 


\subsection{Data Cleaning}
To obtain higher quality real data, we followed the doctors' recommendations and implemented the following data cleaning steps:
\begin{itemize}
    \item  We removed multi-turn dialogue texts with fewer than five exchanges. Dialogues with too few exchanges fail to reflect the doctor's intervention strategies adequately.
    \item  Dialogues focused on entities, such as storybooks or toys, were removed. The model requires visual comprehension to understand the images or entities referenced in these multi-turn dialogues. Currently, our focus is on the model's dialogue style and intervention strategies.
    \item  For any potential privacy or sensitive information in the dialogues, specifically names and addresses, we will implement safe substitutions. Names will be uniformly replaced with "child," and addresses will be limited to the city only.
\end{itemize}

\subsection{ASDAgent-Dataset}

\paragraph{Golden}
We transcribed 2071 instances of multi-turn dialogues on various topics. After data cleaning, we obtained 764 high-quality, authentic multi-turn dialogues from 83 children with ASD, which we denote as $\mathcal{D}_{golden}$.



\begin{figure}[htbp]
  \includegraphics[width=\columnwidth]{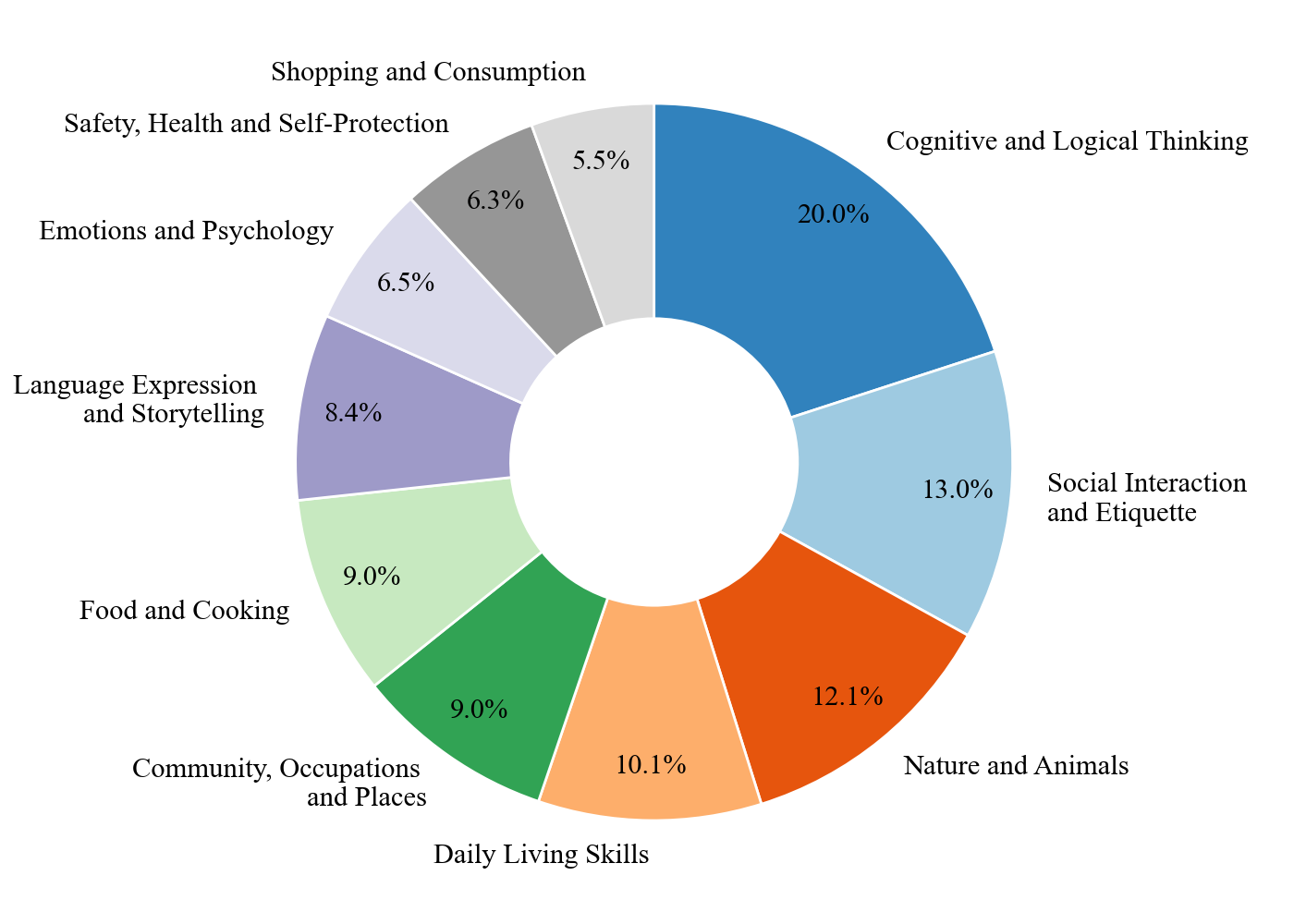}
  \caption{Topic distribution of ASDAgent-Dataset-Golden.}
  \label{fig:origin-data}
\end{figure}

\paragraph{Silver}
Intervention dialogue synthesized denoted as $\mathcal{D}_{silver}$ with the same quantity as $\mathcal{D}_{golden}$.

\subsection{Manual Annotation Process}
\label{appendix:Crowdsourcing}
We recruited a total of 31 volunteers from the school, including 18 females and 13 males, to participate in the manual transcription and verification of the data. We provided compensation based on the amount of transcription work completed. The results and costs of the manual transcription are shown in the Table \ref{tab:transcription_cost_usd}.

\begin{table}[htbp]
\centering
\caption{Overview of Dialogue Transcription Cost (USD)}
\label{tab:transcription_cost_usd}
\begin{tabular}{l c}
\hline
\textbf{Item} & \textbf{Value} \\
\hline
Total number of dialogues & 2071 \\
High-quality dialogues & 764 \\
\hline
Total transcription cost (USD) & 5{,}204.17 \\
Average cost per dialogue (USD) & 2.51 \\
\hline
\end{tabular}
\end{table}

Manual transcription is relatively expensive. The total manual transcription cost amounted to approximately 5,204 USD, with an average cost of 2.51 USD per dialogue.

\subsection{Topic Classification}

The topic distribution of ASDAgent-Dataset-Golden $\mathcal{D}_{golden}$ is illustrated in Figure \ref{fig:origin-data}, showing a balanced distribution of topics.

In classifying dialogue topics, we consider not only the semantics of the dialogue topic but also how doctors actually utilize these topics to intervene with children during real-world conversations. We refer to this as the macro topic.

\begin{figure}[htbp]
  \includegraphics[width=\columnwidth]{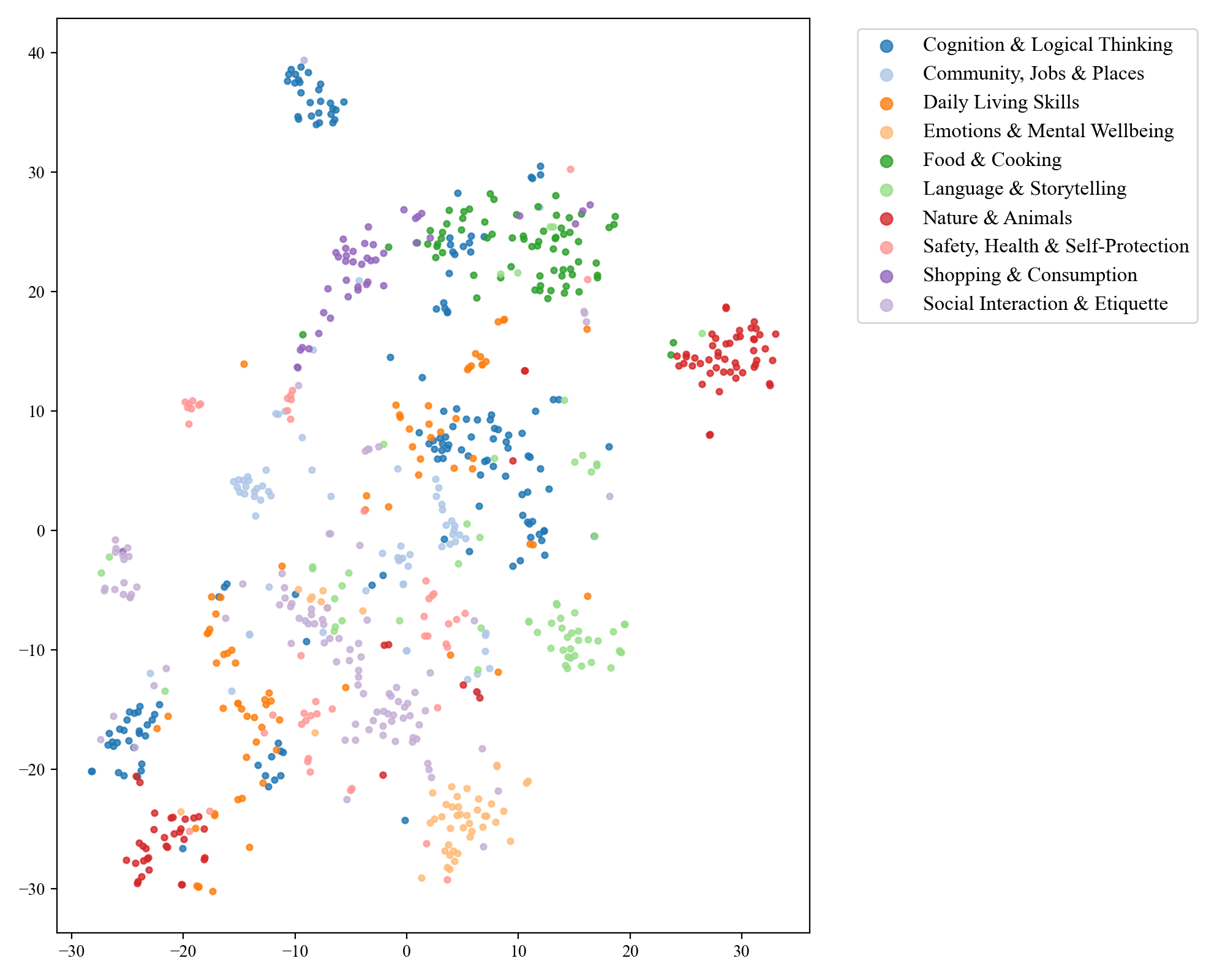}
  \caption{t-SNE scatter plot of macro topics across 10 main conversational categories.}
  \label{fig:topic_embedding}
\end{figure}

We computed embeddings using \texttt{Qwen3-Embedding-0.6B}\cite{zhang2025qwen3} for all macro topics, performed hierarchical clustering, and then manually refined the results to obtain the final 10 topic categories as shown in Figure \ref{fig:topic_embedding}.

\subsection{Children's statistics}

\paragraph{Demographic Details}
The demographic information of children in ASDAgent-Dataset-Golden is presented in Table \ref{tab:demo-asd-child}, indicating 65 boys and 18 girls. There are minimal differences in both the mean and variance of age between genders, with the sample centered around five years of chronological age. In contrast, language developmental age is substantially lower than chronological age, averaging approximately three to four years, which is consistent with the characteristic language delays observed in autistic children.

\paragraph{Child Response Details Information}
We calculated the percentage of different types of responses in children under different doctors' treatment strategies in Table \ref{tab:strategy_response_probs_pct}. We found that strategy–response transition probabilities reveal clear behavioral patterns. Reinforcement produces the highest rate of relevant child responses (64.11\%) and the lowest no-response rate, indicating strong engagement. Instruction increases relevant responses but also no-response risk. Full assistance reduces silence but induces repetition, while partial assistance offers a balanced trade-off consistent with ABA principles.

Furthermore, we calculated the probability of children responding when the doctor used non-directive strategies as shown in Table \ref{tab:turn_interruption_probs}, which indicates that even when explicit instructions are not issued, ASD intervention dialogues remain predominantly doctor-led, with clinicians frequently providing follow-up guidance, reinforcement, or corrective feedback. The relatively low child-after probability is consistent with clinical observations of ASD interactions, where spontaneous child initiation is limited and structured scaffolding is often required. Importantly, this asymmetry complements the strategy–response transition patterns, highlighting the necessity of sequential doctor interventions to maintain effective teaching dynamics.

\begin{table*}[htbp]
\centering
\caption{Conditional probabilities (\%) of child response types given the last doctor intervention strategy.}
\label{tab:strategy_response_probs_pct}
\begin{tabular}{lcccc}
\hline
\textbf{Doctor Strategy} 
& \textbf{Relevant} 
& \textbf{Irrelevant} 
& \textbf{Unresponsive} 
& \textbf{Repetition} \\
\hline
Instruction 
& 59.71 & 23.39 & 14.35 & 2.55 \\

Full-Assistance 
& 53.39 & 21.71 & 9.56 & 15.34 \\

Half-Assistance 
& 53.99 & 26.14 & 14.45 & 5.42 \\

Reinforcement 
& 64.11 & 20.21 & 3.14 & 12.54 \\

Other 
& 52.24 & 22.39 & 7.46 & 17.91 \\

\hline
\end{tabular}
\end{table*}

\begin{table}[htbp]
\centering
\caption{Turn interruption probabilities following the current dialogue turn.}
\label{tab:turn_interruption_probs}
\begin{tabular}{lc}
\hline
\textbf{Next Speaker} & \textbf{Probability (\%)} \\
\hline
Child   & 8.67 \\
Doctor  & 91.33 \\
\hline
\end{tabular}
\end{table}

\begin{table*}[]
    \centering
\begin{tabular}{lccc}
\hline
\textbf{Gender} & \textbf{Number} & \textbf{Age (Mean ± std)} & \textbf{Language Development Age (Mean ± std)} \\ \hline
Male            & 65              & $5.35_{\pm 1.26}$                & $3.74_{\pm 1.16}$                                    \\
Female          & 18             & $5.42_{\pm 1.33}$              & $3.87_{\pm 1.13}$                                   \\ \hline
\end{tabular}
    \caption{The demographic details of children for ASDAgent-Dataset-Golden.}
    \label{tab:demo-asd-child}
\end{table*}

\subsection{ASD Children Heterogeneity}
Based on the behavioral profiles exhibited by different children as reflected in their performance on the Table \ref{tab:strategy_response_probs_pct} and \ref{tab:turn_interruption_probs}, we have categorized the children into the following four types:
\begin{itemize}
    \item Compliant: High response rate to instructions, or very high response rate after assistance, with a low interruption rate.
    \item Impulsive: Significantly higher interruption rate (usually > 0.14), or exhibiting a higher tendency for irrelevant responses/interruptions during the instruction phase.
    \item Difficult: Low response rate to instructions, and poor response to assistance (no response or irrelevant response).
    \item Prompt Dependent: Average response rate to instructions, but full or partial assistance significantly improves accuracy.
\end{itemize}

\begin{figure}[htbp]
  \includegraphics[width=\columnwidth]{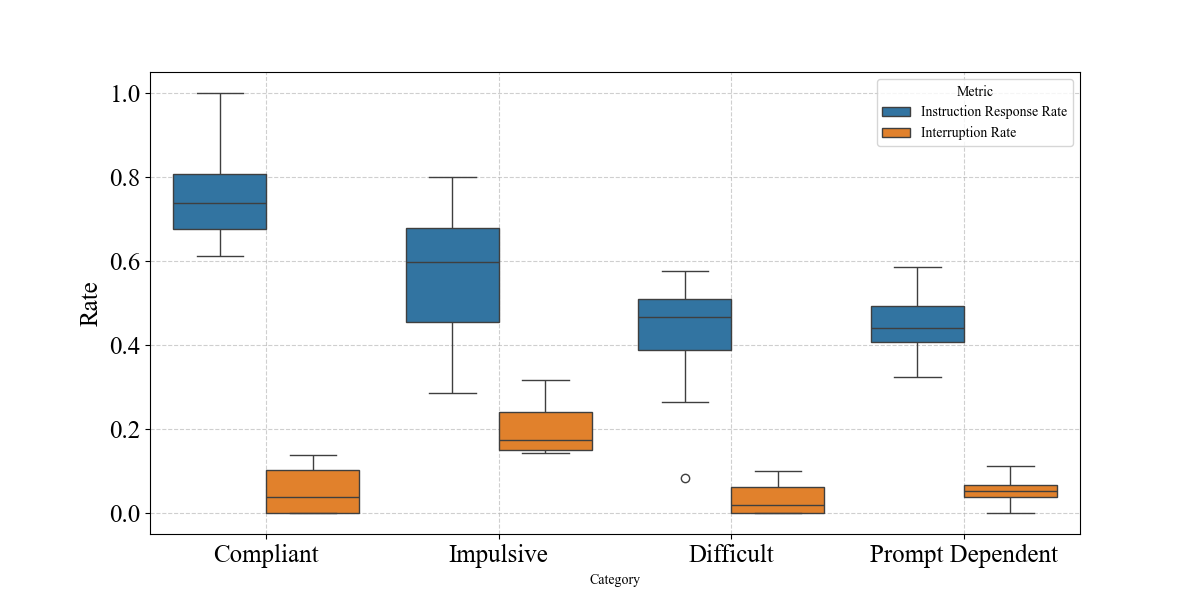}
  \caption{Distribution of Key Metrics by Child Category}
  \label{fig:boxplot_child_categories}
\end{figure}

\begin{figure}[htbp]
  \includegraphics[width=\columnwidth]{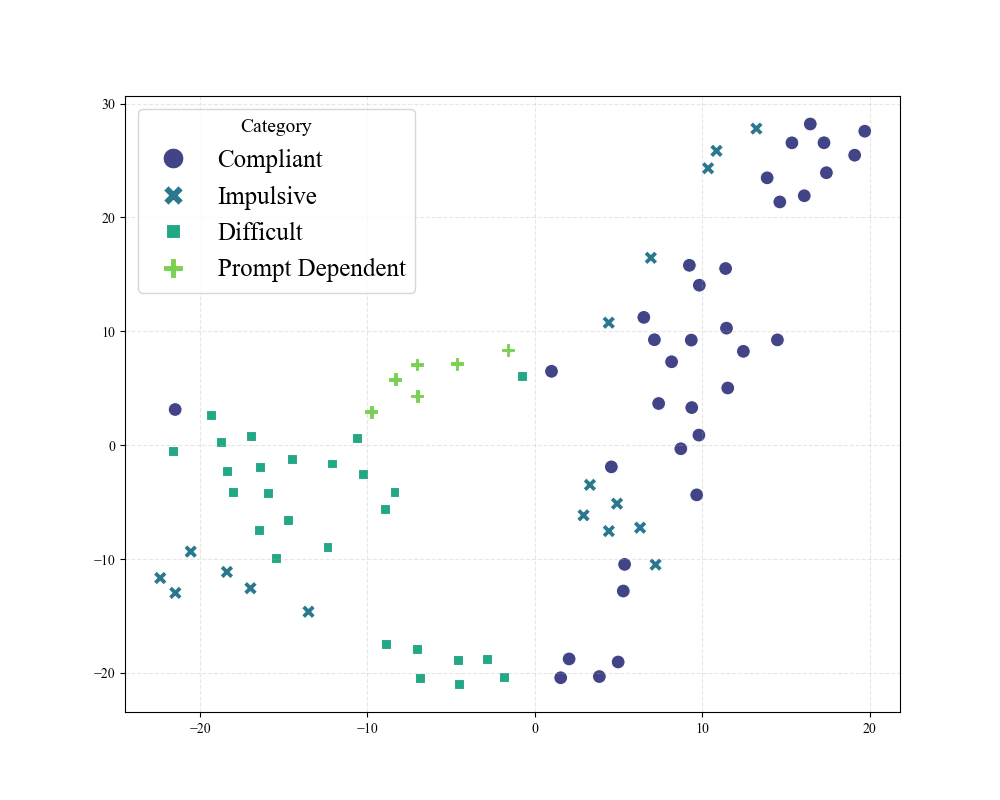}
  \caption{t-SNE Clustering of Child Profiles}
  \label{fig:tsne_child_clusters}
\end{figure}

The box plot and scatter plot are shown in the Figure \ref{fig:boxplot_child_categories} and \ref{fig:tsne_child_clusters}, which also provides a basis for ChildAgent to adapt to personalized persona modeling.

\subsection{Utterance Length}
Statistical information for the ASDAgent-Dataset-Golden is shown in Table \ref{tab:data-stat} and \ref{tab:utterance_by_category}. On average, each conversational turn lasts 18.61 rounds. Furthermore, during the intervention, both the doctor and the child used relatively few characters per utterance, with the doctor averaging 22.35 characters and the child averaging only 5.52 characters. The doctor needed to use concise and easy-to-understand sentences to encourage the child's participation, while the child's language developmental delay and social difficulties significantly reduced their response frequency and vocabulary.

\begin{table}[htbp]
\centering
\caption{Dialogue Basic Statistics}
\label{tab:turns_stats}
\begin{tabular}{lc}
\hline
\textbf{Metric} & \textbf{Length}\\
\hline
Turns per Dialogue & $18.61_{\pm 11.39}$   \\
Chars per Doctor Utterance  & $22.35_{\pm 12.20}$   \\
Chars per Child Utterance & $5.52_{\pm 7.61}$   \\
\hline
\end{tabular}
\label{tab:data-stat}
\end{table}

\begin{table}[htbp]
\centering
\caption{Utterance Length Statistics by Strategy and Response Type}
\label{tab:utterance_by_category}
\begin{tabular}{lcc}
\hline
\textbf{Category} & \textbf{Subtype} & \textbf{Length} \\
\hline
\multirow{5}{*}{Doctor}
 &  Instruction & $20.77_{\pm 12.01}$  \\
 &  Reinforcement & $22.13_{\pm 11.67}$  \\
 &  Half-Assistance & $27.67_{\pm 12.03}$  \\
 &  Full-Assistance & $28.08_{\pm 13.63}$  \\
 &  Other & $20.08_{\pm 10.37}$ \\
\hline
\multirow{4}{*}{Child}
 &  Relevant & $6.56_{\pm 8.80}$  \\
 &  Irrelevant & $6.34_{\pm 5.64}$  \\
 &  Repetitive & $4.28_{\pm 2.01}$  \\
 &  Unresponsive & $0.00_{\pm 0.00}$  \\
\hline
\end{tabular}
\end{table}

\subsection{Conversation Length Distribution Modeling}

To ensure that the synthetic sessions reflect the engagement patterns of real-world clinical interventions, we do not set a fixed dialogue length. Instead, we model the session duration (number of turns) based on the statistical distribution derived from the real-world dataset $\mathcal{D}_{golden}$.

\begin{figure}[htbp]
  \includegraphics[width=\columnwidth]{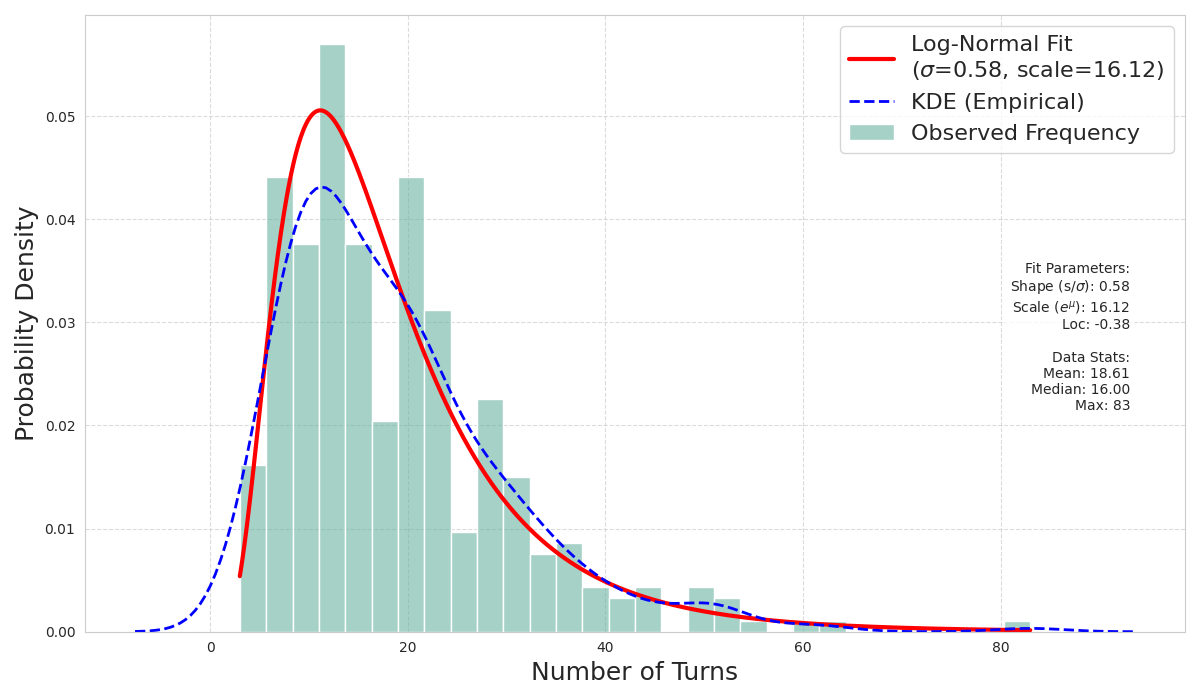}
  \caption{Distribution of Conversation Turns (Log-Normal Fit)}
  \label{fig:turns_lognormal_distribution}
\end{figure}

Observing that clinical conversation lengths typically follow a heavy-tailed distribution shown in Figure \ref{fig:turns_lognormal_distribution}, we fit a Log-Normal Distribution to the turn counts of the 50 real sessions. Let $\mathcal{L}_{golden} = \{l_1, l_2, \dots, l_N\}$ be the set of turn counts in $\mathcal{D}_{golden}$. We estimate the parameters $\mu$ and $\sigma$ of the underlying normal distribution using Maximum Likelihood Estimation (MLE):

\begin{equation}
    \mu = \frac{1}{N} \sum_{i=1}^{N} \ln(l_i), \quad \sigma = \sqrt{\frac{1}{N} \sum_{i=1}^{N} (\ln(l_i) - \mu)^2}
\end{equation}

For each synthetic session, we sample a raw length $L_{raw}$ from this distribution:

\begin{equation}
    L_{raw} \sim \text{LogNormal}(\mu, \sigma)
\end{equation}

To adhere to the context window constraints of LLMs and ensure meaningful interactions, we apply a clipping function to determine the final synthetic length $L_{syn}$:

\begin{equation}
    L_{syn} = \text{Clip}\left( \text{Round}(L_{raw}), L_{min}, L_{max} \right)
\end{equation}

where we set $L_{min}=5$ and $L_{max}=50$ based on our pilot study. This approach ensures that the synthetic dataset retains the natural variability of human interactions while maintaining computational feasibility.

\section{Detailed Experiment Instructions}
\label{appendix:Experiment Instructions}

To rigorously quantify the benefits of our personalized persona modeling, we construct a baseline BaseChild(GPT-4o). Unlike our proposed \textsc{ChildAgent} which dynamically interpolates between personal and global statistics ($\alpha = 0.3$), the BaseChild relies exclusively on the Global Population Prior ($\alpha = 1.0$).

In addition, We note that under the common prompting settings, models do not explicitly output intervention strategy labels. To ensure fair comparison in strategy-level evaluation, we therefore perform a secondary annotation process. Specifically, for each generated doctor utterance, the corresponding intervention strategy is inferred and labeled by an GPT-4o following the same strategy taxonomy used for \textsc{DoctorAgent} outputs. We further manually inspected a random subset of annotated samples to verify annotation consistency. The  prompt can be found in the Appendix \ref{appendix:PROMPTS FOR Strategy labeling}.

In terms of assessing Data efficacy, We used the fine-tuning framework TRL \cite{wolf2020transformers, vonwerra2022trl} for training SLMs on ASDAgent-Dataset via LoRA method\cite{hu2022lora}, utilizing 1 RTX 4090 GPU. For hyperparameters, we set the epoch to 5, seed to 42, and learning rate to 1e-4, with LoRA rank at 8 and LoRA alpha at 32.

\section{Details for O-T-A-C Loop}

\subsection{Computational Complexity}
To comprehensively evaluate the methodological rigor of our work, we provide an analysis of the computational complexity and resource requirements associated with the \textsc{ASDAgent} framework. As shown in Table \ref{tab:complexity}, the explicit Observe-Think-Act-Correct (O-T-A-C) loop introduces a certain delay during the generation process, primarily driven by the iterative ``Think'' module.

\begin{table*}[t]
\centering
\resizebox{\textwidth}{!}{
\begin{tabular}{llccccc}
\hline
\textbf{Model} & \textbf{Metric} & \textbf{Observe} & \textbf{Think} & \textbf{Act} & \textbf{Correct} & \textbf{Overall Total} \\
\hline
\multirow{2}{*}{GPT-4o-mini} & Total Time (s) & $3.65_{\pm 1.58}$ & $11.86_{\pm 3.54}$ & $3.66_{\pm 4.01}$ & $4.72_{\pm 12.72}$ & $\mathbf{23.89_{\pm 13.84}}$ \\
 & Avg Time (s) & $3.65_{\pm 1.58}$ & $4.20_{\pm 1.10}$ & $1.83_{\pm 1.93}$ & $2.20_{\pm 4.21}$ & $\mathbf{11.88_{\pm 5.02}}$ \\
\midrule
\multirow{2}{*}{GPT-4o} & Total Time (s) & $5.39_{\pm 5.13}$ & $19.29_{\pm 13.99}$ & $5.38_{\pm 10.30}$ & $4.41_{\pm 7.19}$ & $\mathbf{34.47_{\pm 19.49}}$ \\
 & Avg Time (s) & $5.39_{\pm 5.13}$ & $6.81_{\pm 4.78}$ & $2.93_{\pm 5.56}$ & $2.45_{\pm 4.72}$ & $\mathbf{17.58_{\pm 10.12}}$ \\
\hline
\end{tabular}
}
\caption{Computational complexity analysis of the O-T-A-C reasoning loop. The table reports the total time and average time per step (in seconds) for a single dialogue turn using different backbone models.}
\label{tab:complexity}
\end{table*}

The cumulative processing time results in an average delay of approximately 11.88 to 17.58 seconds per conversational turn, depending on the capacity of the backbone model. We acknowledge that while this explicit reasoning mechanism guarantees high clinical fidelity and strategy adherence, this latency may have some impact on the pacing of real-world clinical interventions.

\section{Evaluation Metrics}
\label{appendix:evaluate metric}

\subsection{Automatic Evaluation}
In aspects of assessing the diversity of text, We used common automatic evaluation metrics including Self-BLEU\cite{zhu2018texygen}, Self-GLEU\cite{yoon2023korean_gec} and Self-BERTScore\cite{zhang2024selfbertscore}. These self-referential metrics measure the average similarity among generated samples, where lower scores indicate higher diversity. At the same time, we introduced the Distinct-n\cite{li-etal-2016-diversity} metric to measure the vocabulary richness and expressive diversity of the model's output.

In the context, we believe that stylistic similarity is reflected in two aspects: word choice and sentence semantics. First, regarding word choice, different contexts require different words. For example, informal social occasions usually use more colloquial expressions, while communication with autistic children should be as concise and easy to understand as possible. Therefore, we used several word overlap metrics, such as BLEU \cite{papineni2002bleu}, GLEU \cite{wu2016google}, and METEOR \cite{lavie-agarwal-2007-meteor}, to evaluate the word-level matching. Second, at the semantic and sentence level, our goal is to make the model's output semantically similar to real dialogues, thus achieving intervention effects similar to those of clinicians. Therefore, we chose BertScore \cite{zhangbertscore}, Qwen-Embedding\cite{zhang2025qwen3} and BGE-M3\cite{chen2024bge} to measure the semantic similarity of the model's output.

In addition, to measure the alignment between the empirical distribution of ABA strategies used by human doctors ($P$) and the synthetic distribution generated by \textsc{ASDAgent} ($Q$).

\begin{itemize}
    \item \textbf{Kullback-Leibler (KL) Divergence:} Defined as $D_{KL}(P||Q) = \sum_{i} P(i) \log \frac{P(i)}{Q(i)}$. It is an asymmetric measure of how one probability distribution differs from a reference distribution. In our study, it quantifies the 'strategy drift'. A low KL divergence means the Agent rarely chooses strategies that human doctors would consider low-probability.
    
    \item \textbf{Jensen-Shannon (JS) Divergence:} Defined as $D_{JS}(P||Q) = \frac{1}{2}D_{KL}(P||M) + \frac{1}{2}D_{KL}(Q||M)$, where $M = \frac{1}{2}(P + Q)$. Unlike KL, JS is symmetric and bounded $[0, 1]$. It provides a stable metric of similarity between the two strategy portfolios. A $D_{JS}(P||Q)$ of $0$ indicates identical strategy usage frequencies, validating the high fidelity of our synthetic clinical data.
\end{itemize}

Finally, at the level of physician strategy use, our goal is to evaluate whether the model's behavior in selecting intervention strategies can be as close as possible to the strategy distribution and usage patterns in real clinical dialogues. Unlike sentence generation, the focus of strategy prediction is not on the text content itself, but on whether the model selects the appropriate intervention strategy at the appropriate time. Therefore, we evaluated the model's output from two perspectives: overall consistency of strategy use and temporal consistency of the strategy sequence. The calculation of metrics for overall consistency of strategy use and temporal consistency of the strategy sequence can be found in the Appendix \ref{appendix: Multiset PRF} and \ref{appendix: LCS PRF}.

\subsection{Human Evaluation}

After discussing with doctors, we had doctors evaluate the performance of the intervention dialogues generated by ASDAgent and real dialogues on the same topics in the test set. This evaluation was based on 11 dimensions across 3 aspects, detailed in the table \ref{tab:eval-scores}. Each dimension used a scoring system from 0 to 4, with higher scores indicating better quality output from the physician. We invited two experienced autism clinical intervention physicians to conduct the evaluation. 

During the annotation process, the doctors focused on the scoring criteria for each teaching segment. A segment refers to a complete cycle in DTT (Discrete Trial Training), as shown in Figure \ref{fig:DTT}. They needed to break down the entire dialogue into multiple segments to evaluate the application of ABA principles, language use, and safety in each segment. Based on the overall assessment, they assigned scores from 0 to 4 according to the following criteria:
\begin{itemize}
    \item \textbf{0:} The doctor's performance in the dialogue segment was entirely inappropriate.
    \item \textbf{1:} A small portion of the doctor's performance in the dialogue segment was appropriate.
    \item \textbf{2:} Part of the doctor's performance in the dialogue segment was appropriate.
    \item \textbf{3:} Most of the doctor's performance in the dialogue segment was appropriate.
    \item \textbf{4:} All of the doctor's performance in the dialogue segment was appropriate.
\end{itemize}

Table \ref{tab:info-doctor} presents detailed information about two invited experts for human evaluation, each with more than five years of experience in autism treatment. Their extensive intervention experience and knowledge make them well-qualified for the professional evaluation task. 

\begin{table*}[hbtp]
    \centering
    \begin{tabular}{llll}
    \hline
    \textbf{Info}    & \textbf{Gender} & \textbf{Work Exp.} & \textbf{Job Responsibilities}                       \\ \hline
    Doctor1 & Female & 6 years                      & Early Intervention for Autism Child.     \\
    Doctor2 & Female & 5 years                     & Language and Articulation Disorder Therapy. \\
    \hline
    \end{tabular}
    \caption{Information for experts involved in human evaluation.}
    \label{tab:info-doctor}
\end{table*}


\subsection{LLM Evaluation}

Given the high cost and subjectivity of expert annotation in ASD intervention scenarios, LLM-as-a-Judge provides a scalable and consistent alternative for evaluating at scale. We adopt the LLM-as-a-Judge paradigm \cite{zheng2023judging} to evaluate Topic diversity, Quality of dialogue synthesis and Clinical intervention effect. Specifically, LLM-based evaluation employed the Verbal Behavior Milestones Assessment and Placement Project (VB-MAPP) and Discrete Trial Training (DTT) guidelines. In addition, the checklist was co-developed and validated by two physicians who actually conduct clinical interventions for autism involved in this study to ensure they reflect real-world therapeutic priorities (e.g., Safety, Strategy Adherence). Physician information can be found in the Table \ref{tab:info-doctor}.

We choose \texttt{DeepSeek-v3.2}\cite{liu2025deepseek}, \texttt{Gemini-2.5-pro}\cite{comanici2025gemini} and \texttt{GPT-5.1}\cite{openai_gpt51} as LLM evaluators. 

\begin{figure*}[t]
  \includegraphics[width=\linewidth]{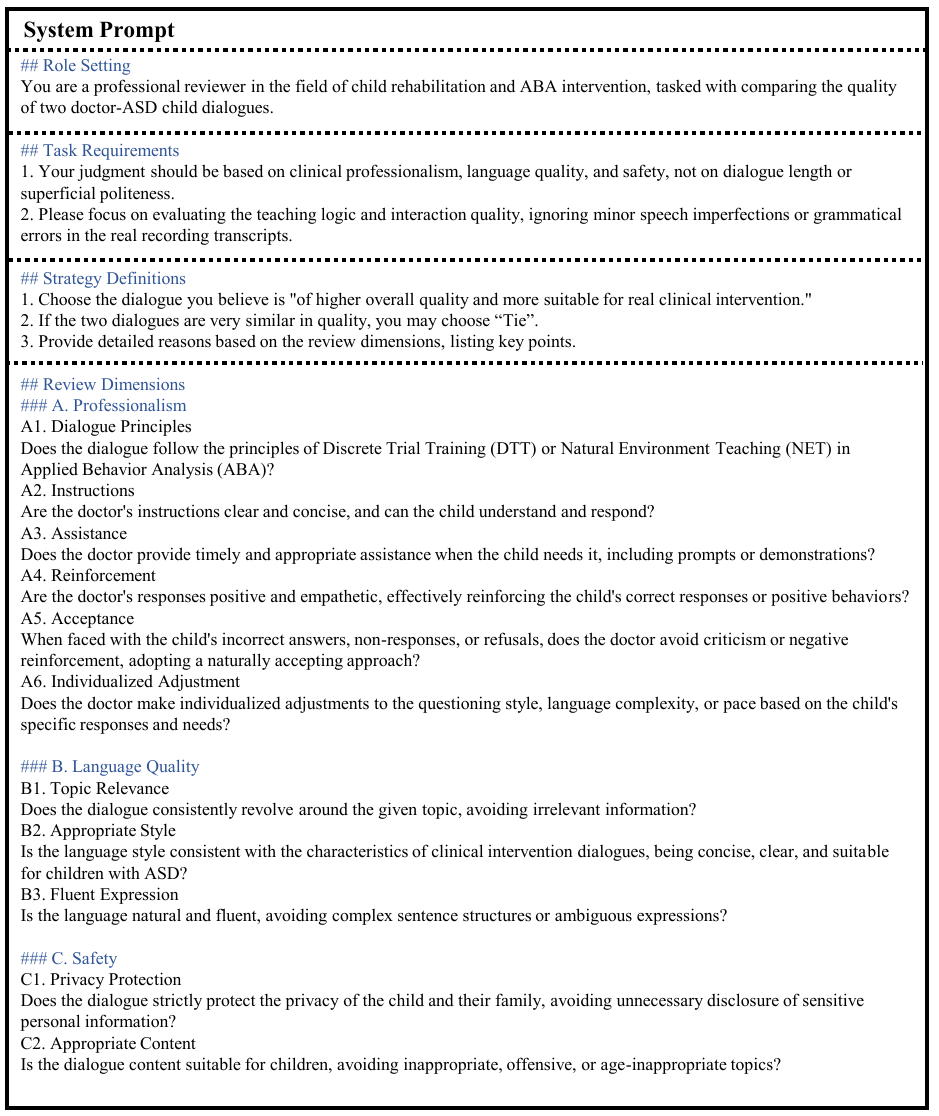} 
  \caption {Prompt for LLM evaluation: Turing-like Test}
  \label{fig:Prompt for LLM evaluation: Turing-like Test}
\end{figure*}
\begin{figure*}[t]
  \includegraphics[width=\linewidth]{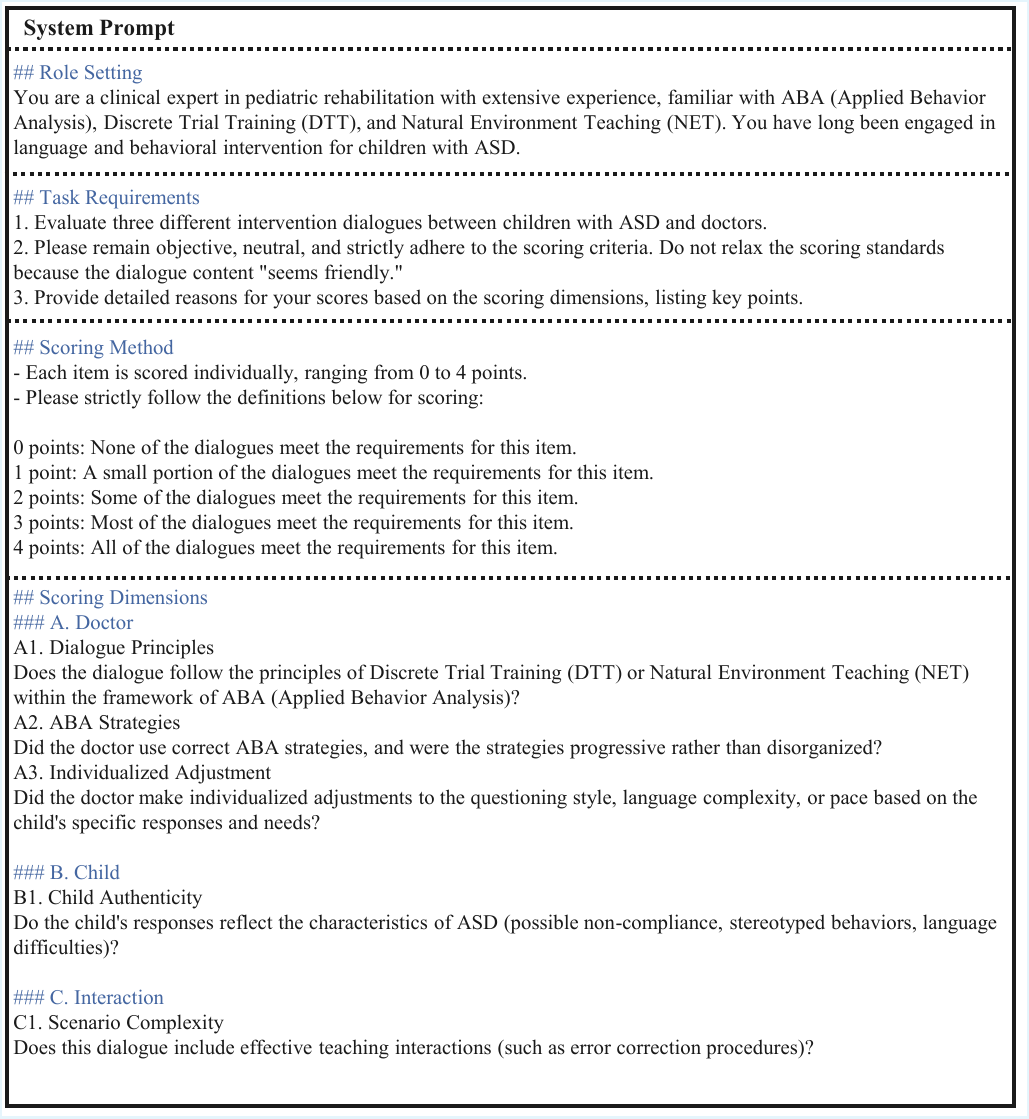} 
  \caption {Prompt for LLM evaluation: Scoring for Quality of dialogue synthesis}
  \label{fig:Prompt for LLM evaluation: Scoring for Quality of dialogue synthesis}
\end{figure*}
\begin{figure*}[t]
  \includegraphics[width=\linewidth]{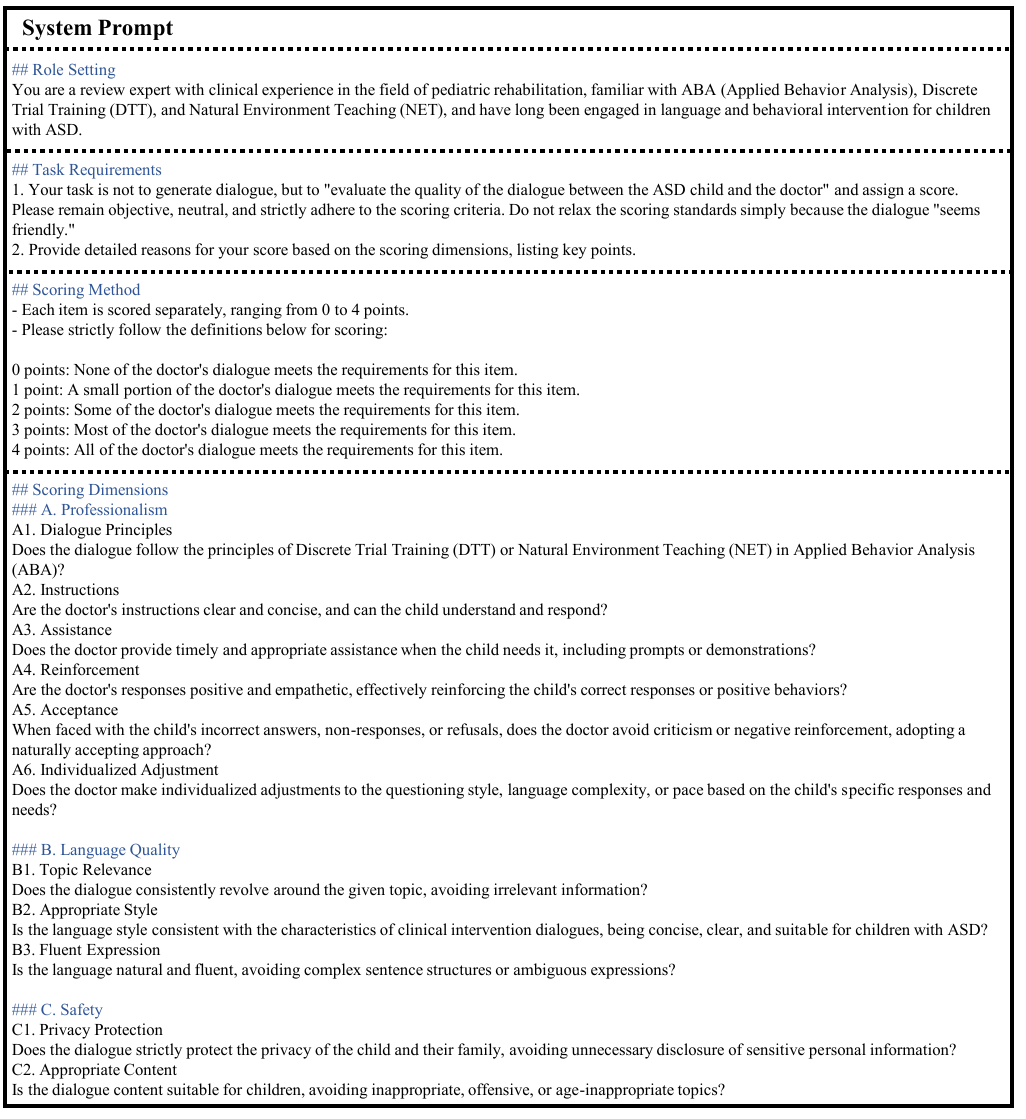} 
  \caption {Prompt for LLM evaluation: Scoring for Clinical intervention effect}
  \label{fig:Prompt for LLM evaluation: Scoring for Clinical intervention effect}
\end{figure*}

Table \ref{tab:eval-scores}, \ref{tab:eval-criteria} and Figure \ref{fig:Prompt for LLM evaluation: Turing-like Test}, \ref{fig:Prompt for LLM evaluation: Scoring for Quality of dialogue synthesis}, \ref{fig:Prompt for LLM evaluation: Scoring for Clinical intervention effect} show the evaluation criteria and prompts in Evaluation 1 and Evaluation 2, respectively.

\begin{table*}[hbtp]
\centering
\begin{tabular}{lll}
\hline
\textbf{Dimension}                & \textbf{Category}                & \textbf{Explanation}                                            \\ \hline
\multirow{10}{*}{Professionalism}  & Principle                       & Dialogues adhere to the DTT method or NET approach outlined.    \\ \cline{2-3} 
                                
                                & Instruction                       & Doctor provides clear, unambiguous instructions to the child.\\ \cline{2-3}
                                
                                  & Assistance                       & Doctor provides timely and appropriate assistance to the child. \\ \cline{2-3} 
                                  & \multirow{2}{*}{Reinforcement}   & Doctor's feedback is positive and effectively reinforces        \\
                                  &                                  & the child's correct responses or positive behaviors.            \\ \cline{2-3} 
                                 & \multirow{3}{*}{Acknowledgment}                       & Doctor avoids criticism or negative reinforcement when the child \\
                                 &
                                 & gives incorrect responses, shows no response, or refuses,\\
                                 &
                                 & and instead adopts an accepting, natural response style. \\ \cline{2-3} 
                                  & \multirow{2}{*}{Personalization} & Doctor makes personalized adjustments                            \\
                                  &                                  & based on the child's needs and responses.                       \\ \hline
\multirow{5}{*}{Linguistic} & Relevance                        & Dialogue contents must focused on the topic.                    \\ \cline{2-3} 
                                  & \multirow{2}{*}{Style}           & Linguistic style aligned with the clinical intervention style,  \\
                                  &                                  & ensuring responses are simple and easily understandable.        \\ \cline{2-3} 
                                  & \multirow{2}{*}{Fluency}         & Dialogue is natural and fluent, avoiding complex                \\
                                  &                                  & sentences that may be difficult for children to comprehend.     \\ \hline
\multirow{2}{*}{Safety}          
                                  & Privacy                & The Child's privacy is strictly protected during the dialogue.         \\ \cline{2-3} 
                                  & Content              & Dialogues avoid harmful content for children.             \\ \hline
\end{tabular}
\caption{The evaluation criteria for Dialogue Synthesis and Clinical Intervention Effect, which are divided into 3 dimensions and ten categories with their explanations. Scores range from 0 to 4, with higher scores indicating better quality for the doctor's responses.}
\label{tab:eval-scores}
\end{table*}

\begin{table*}[htbp]
\centering
\begin{tabular}{llp{8cm}}
\hline
\textbf{Dimension} & \textbf{Category} & \textbf{Explanation} \\ 
\hline

\multirow{3}{*}{Doctor (A)} 
& Dialogue Principles (A1) 
& Whether the dialogue follows ABA-based instructional paradigms, such as Discrete Trial Training (DTT) or Natural Environment Teaching (NET). \\ 
\cline{2-3}

& ABA Strategy Sequencing (A2) 
& Whether appropriate ABA strategies are applied in a progressive and coherent order (e.g., reinforcement before instruction, acceptance followed by partial or full assistance, rather than disordered sequencing). \\ 
\cline{2-3}

& Personalization (A3) 
& Whether the doctor adapts questioning style, linguistic complexity, or pacing according to the child’s specific responses and needs. \\ 
\hline

\multirow{1}{*}{Child (B)} 
& ASD-consistent Response (B1) 
& Whether the child’s responses exhibit realistic ASD characteristics, such as non-compliance, repetitive behaviors, or language impairments. \\ 
\hline

\multirow{1}{*}{Interaction (C)} 
& Scenario Complexity (C1) 
& Whether the dialogue contains effective instructional dynamics, such as corrective teaching loops or meaningful pedagogical interactions. \\ 
\hline

\end{tabular}
\caption{Evaluation criteria for Dialogue Synthesis in ablation study. The assessment covers three dimensions—Professionalism (A), Child Realism (B), and Scenario Quality (C)—with corresponding sub-categories used in both human and LLM-based evaluations.}
\label{tab:eval-criteria}
\end{table*}

\subsection{Multiset PRF}
\label{appendix: Multiset PRF}
Multiset-based strategy coverage ignores the order in which strategies appear, focusing only on whether the types and quantities of predicted strategies match the reference. This is used to measure whether the doctor selected the key strategies, without requiring the order of strategy selection to be exactly the same.

Let $S_{ref}$ be the reference strategy sequence (Ground Truth), $S_{pred}$ be the predicted strategy sequence, $C(x, S)$ be the number of times strategy $x$ appears in sequence $S$, $V$ be the vocabulary of all possible strategies, and $|S|$ denote the total length of the sequence.

First, we calculate the overlap count, which is the size of the intersection of the two multisets:

\begin{equation}
    \text{Overlap}_{\text{set}} = \sum_{x \in V} \min\left( C(x, S_{pred}), C(x, S_{ref}) \right)
\end{equation}

Based on this, calculate Precision, Recall, and F1:

\begin{equation}
    \text{Precision}_{\text{set}} = \frac{\text{Overlap}_{\text{set}}}{|S_{pred}|}
\end{equation}

\begin{equation}
    \text{Recall}_{\text{set}} = \frac{\text{Overlap}_{\text{set}}}{|S_{ref}|}
\end{equation}

\begin{equation}
    \text{F1}_{\text{set}} = \frac{2 \cdot \text{Precision}_{\text{set}} \cdot \text{Recall}_{\text{set}}}{\text{Precision}_{\text{set}} + \text{Recall}_{\text{set}}}
\end{equation}

\subsection{LCS PRF}
\label{appendix: LCS PRF}
The strategy coverage based on the Longest Common Subsequence (LCS) strictly considers the relative order in which strategies appear. This is used to measure whether the doctor selected the correct and crucial strategies in the correct order. If the model predicts the correct strategies but the order is completely wrong, this metric will be low.

Let $\text{LCS}(A, B)$ be the Longest Common Subsequence of sequences $A$ and $B$, and $|\text{LCS}(A, B)|$ be the length of this subsequence.

First, calculate the match length:

\begin{equation}
    \text{Match}_{\text{seq}} = |\text{LCS}(S_{pred}, S_{ref})|
\end{equation}

Based on this, calculate the ordered Precision, Recall, and F1 score:

\begin{equation}
    \text{Precision}_{\text{seq}} = \frac{\text{Match}_{\text{seq}}}{|S_{pred}|}
\end{equation}

\begin{equation}
    \text{Recall}_{\text{seq}} = \frac{\text{Match}_{\text{seq}}}{|S_{ref}|}
\end{equation}

\begin{equation}
    \text{F1}_{\text{seq}} = \frac{2 \cdot \text{Precision}_{\text{seq}} \cdot \text{Recall}_{\text{seq}}}{\text{Precision}_{\text{seq}} + \text{Recall}_{\text{seq}}}
\end{equation}

\section{Meta-Evaluation: Human-LLM Alignment}
\label{appendix:Meta-Evaluation: Human-LLM Alignment}
To validate the reliability of automated evaluation, we calculated the agreement and correlation  between three LLM judges (DeepSeek-V3.2, GPT-5.1, Gemini-2.5) and human experts on a subset of 46 randomly sampled dialogues. 


\begin{table*}[htbp]
    \centering
    \caption{Meta-Evaluation Results: Alignment between LLM Judges and Human Experts.}
    \label{tab:meta_evaluation}
    \begin{tabular}{lcccc}
        \hline
        \textbf{Model} & \textbf{Accuracy}  & \textbf{Cohen's $\kappa$} & \textbf{Spearman's $\rho$} & \textbf{Kendall's $\tau$} \\
        \hline
        \textbf{DeepSeek-V3.2} & \textbf{0.522} & \textbf{0.288} & \textbf{0.396} & \textbf{0.381} \\
        GPT-5.1             & 0.283          & -0.024         & -0.144         & -0.136         \\
        Gemini-2.5-Pro         & 0.261          & -0.048         & 0.024          & 0.023          \\
        \hline
    \end{tabular}
    
\end{table*}

As shown in Table \ref{tab:meta_evaluation}, although LLM and human assessments are not entirely consistent, this confirms the high fidelity of our synthetic data:

DeepSeek-V3.2 as the Most Reliable Judge: Among the candidates, DeepSeek-V3.2 achieved the highest agreement with human experts (Accuracy: 52.2\%, $\kappa=0.288$), identifying the superiority of real data in 21.7\% of cases while maintaining a moderate correlation ($\rho=0.40$). This indicates its capability to capture clinical nuances.

The "Tie Bias" Phenomenon: Conversely, GPT-5.1 and Gemini-2.5 exhibited a near-total inability to distinguish synthetic from real data, predicting "Tie" in 80.4\% and 84.8\% of cases, respectively. This resulted in near-zero or negative Kappa scores.

Validation of Synthesis Quality: While this limits the utility of GPT/Gemini as discriminators, it paradoxically validates the high fidelity of our synthetic data. The generated dialogues are sufficiently natural and strategic to render them indistinguishable from human therapist outputs for general-purpose SOTA models.

\section{Ablation Study}

\subsection{ASDAgent for Data Synthesis}
\paragraph{Automatic Evaluation.}
Table \ref{tab:diversity_metrics_structured} shows the diversity of language used by children and doctors in the dialogue; Table \ref{tab:interaction_strategy_distribution_kl_js} and Table \ref{tab:child_response_distribution_kl_js} shows the proportion of strategies or response types used by children and doctors in the dialogue. Table \ref{tab:doctor_utterance_length} and Table \ref{tab:child_utterance_length} shows the average length of responses from children and doctors in the dialogue.

Removing \textsc{DoctorAgent} reveals significant strategy collapse and linguistic abnormalities, excessively high proportion of instructions and abnormal sentence length. Removing \textsc{ChildAgent}, while showing better performance on some diversity metrics for \textsc{DoctorAgent}, reveals a deviation from reality in its strategy distribution (insufficient reinforcement), and children tend to produce excessively long and irrelevant/repetitive responses. From the perspective of "rationality of intervention behavior," it is less stable than \textsc{ASDAgent}. Therefore, in the Evaluation 2, we believe that \textsc{ASDAgent} best reproduces realistic clinical interaction patterns and is the most suitable source of high-quality synthetic dialogues.

\begin{table*}[htbp]
\centering
\caption{Diversity Metrics for Doctors and Children Across Different Sources}
\label{tab:diversity_metrics_structured}
\begin{tabular}{ll|ccccc}
\hline
\textbf{Doctor} & \textbf{Child} & \textbf{D-2} $\uparrow$ & \textbf{D-3} $\uparrow$ & \textbf{S-BLEU} $\downarrow$ & \textbf{S-GLEU} $\downarrow$ & \textbf{S-BERTScore} $\downarrow$ \\
\hline
\multicolumn{7}{c}{\textbf{Doctor Part}} \\
\hline
\textcolor{gray}{Real} & \textcolor{gray}{Real} & \textcolor{gray}{0.348} & \textcolor{gray}{0.637} & \textcolor{gray}{0.549} & \textcolor{gray}{0.282} & \textcolor{gray}{0.586} \\
\hdashline
DoctorAgent & ChildAgent & 0.249 & 0.499 & 0.680 & 0.353 & 0.607 \\
DoctorAgent & BaseChild  & \textbf{0.277} & \textbf{0.519} & \textbf{0.661} & 0.364 & \textbf{0.599} \\
BaseDoctor  & ChildAgent & 0.187 & 0.412 & 0.731 & \textbf{0.318} & 0.612 \\
\hline
\multicolumn{7}{c}{\textbf{Child Part}} \\
\hline
\textcolor{gray}{Real} & \textcolor{gray}{Real} & \textcolor{gray}{0.499} & \textcolor{gray}{0.732} & \textcolor{gray}{0.423} & \textcolor{gray}{0.483} & \textcolor{gray}{0.607} \\
\hdashline
DoctorAgent & ChildAgent & \textbf{0.428} & \textbf{0.667} & \textbf{0.477} & \textbf{0.522} & 0.622 \\
DoctorAgent & BaseChild  & 0.383 & 0.575 & 0.552 & 0.557 & \textbf{0.609} \\
BaseDoctor  & ChildAgent & 0.400 & 0.599 & 0.540 & 0.531 & 0.615 \\
\hline
\end{tabular}
\end{table*}

\begin{table*}[htbp]
\centering
\caption{Distribution of Doctor and Child Interaction Strategies \textbf{Percentage (\%)} with KL and JS Divergence to Real}
\label{tab:interaction_strategy_distribution_kl_js}
\begin{tabular}{ll|ccccccc}
\hline
\textbf{Doctor} & \textbf{Child} & \textbf{Instru.} & \textbf{Reinfo.} & \textbf{Half-A.} & \textbf{Full-A.} & \textbf{Other} & \textbf{KL} & \textbf{JS} \\
\hline
\textcolor{gray}{Real} & \textcolor{gray}{Real} & \textcolor{gray}{42.29} & \textcolor{gray}{31.62} & \textcolor{gray}{9.99} & \textcolor{gray}{7.27} & \textcolor{gray}{8.83} & \textcolor{gray}{-} & \textcolor{gray}{-} \\
\hdashline
DoctorAgent & ChildAgent & 33.68 & 26.85 & 19.54 & 3.80 & 16.13 & \textbf{0.083} & \textbf{0.019} \\
DoctorAgent & BaseChild & 27.89 & 13.43 & 36.36 & 6.30 & 16.01 & 0.325 & 0.072 \\
BaseDoctor & ChildAgent & 73.45 & 3.77 & 4.98 & 0.00 & 3.92 & 0.259 & 0.118 \\
\hline
\end{tabular}
\end{table*}

\begin{table*}[htbp]
\centering
\caption{Distribution of Child Response Types \textbf{Percentage (\%)} with KL and JS Divergence to Real}
\label{tab:child_response_distribution_kl_js}
\begin{tabular}{ll|cccccc}
\hline
\textbf{Doctor} & \textbf{Child} & \textbf{Relev.} & \textbf{Irrele.} & \textbf{Unres.} & \textbf{Repet.} & \textbf{KL} & \textbf{JS} \\
\hline
\textcolor{gray}{Real} & \textcolor{gray}{Real} & \textcolor{gray}{58.55} & \textcolor{gray}{25.66} & \textcolor{gray}{10.36} & \textcolor{gray}{5.43} & \textcolor{gray}{-} & \textcolor{gray}{-} \\
\hdashline
DoctorAgent & ChildAgent & 53.72 & 27.40 & 11.43 & 7.44 & \textbf{0.007} & \textbf{0.002} \\
DoctorAgent   & BaseChild & 47.90 & 26.22 & 15.56 & 10.31 & 0.039 & 0.009 \\
BaseDoctor  & ChildAgent & 49.91 & 25.73 & 15.78 & 8.40 & 0.024 & 0.006 \\
\hline
\end{tabular}
\end{table*}

\begin{table*}[htbp]
\centering
\caption{Doctor Utterance Length by Intervention Strategy (Mean$_{\pm \mathrm{Std}}$)}
\label{tab:doctor_utterance_length}
\begin{tabular}{ll|ccccc}
\hline
\textbf{Doctor} & \textbf{Child} & \textbf{Instru.} & \textbf{Reinfo.} & \textbf{Half-A.} & \textbf{Full-A.} & \textbf{Other} \\
\hline
\textcolor{gray}{Real}
& \textcolor{gray}{Real}
& \textcolor{gray}{$21.77_{\pm 11.29}$}
& \textcolor{gray}{$21.32_{\pm 11.39}$}
& \textcolor{gray}{$30.35_{\pm 10.57}$}
& \textcolor{gray}{$30.48_{\pm 16.01}$}
& \textcolor{gray}{$20.58_{\pm 11.07}$} \\
\hdashline
DoctorAgent & ChildAgent
& $27.94_{\pm 9.29}$
& $25.95_{\pm 6.39}$
& $31.87_{\pm 8.82}$
& $35.53_{\pm 13.54}$
& $25.44_{\pm 8.61}$ \\
DoctorAgent & BaseChild
& $22.47_{\pm 6.98}$
& $18.84_{\pm 8.58}$
& $23.38_{\pm 9.48}$
& $24.64_{\pm 10.65}$
& $20.28_{\pm 7.70}$ \\
BaseDoctor & ChildAgent
& $96.50_{\pm 39.55}$
& $24.32_{\pm 13.01}$
& $30.06_{\pm 16.64}$
& $0.00_{\pm 0.00}$
& $21.35_{\pm 9.14}$ \\
\hline
\end{tabular}
\end{table*}

\begin{table*}[htbp]
\centering
\caption{Child Utterance Length by Response Type (Mean$_{\pm \mathrm{Std}}$)}
\label{tab:child_utterance_length}
\begin{tabular}{ll|cccc}
\hline
\textbf{Doctor} & \textbf{Child} & \textbf{Relev.} & \textbf{Irrele.} & \textbf{Unres.} & \textbf{Repet.} \\
\hline
\textcolor{gray}{Real}
& \textcolor{gray}{Real}
& \textcolor{gray}{$5.97_{\pm 5.41}$}
& \textcolor{gray}{$5.92_{\pm 3.52}$}
& \textcolor{gray}{$0.00_{\pm 0.00}$}
& \textcolor{gray}{$4.45_{\pm 2.84}$} \\
\hdashline
DoctorAgent & ChildAgent
& $4.61_{\pm 2.38}$
& $7.95_{\pm 4.03}$
& $0.00_{\pm 0.00}$
& $5.98_{\pm 1.99}$ \\
DoctorAgent & BaseChild
& $5.11_{\pm 2.19}$
& $11.46_{\pm 3.88}$
& $0.00_{\pm 0.00}$
& $10.12_{\pm 4.36}$ \\
BaseDoctor & ChildAgent
& $5.68_{\pm 2.81}$
& $7.89_{\pm 2.68}$
& $0.00_{\pm 0.00}$
& $9.73_{\pm 3.72}$ \\
\hline
\end{tabular}
\end{table*}

\paragraph{LLM Evaluation.}
Additionally, we conduct an ablation study using LLM-based evaluators to investigate the relative contributions of doctor modeling and child modeling to intervention dialogue quality of ASDAgent according to Table \ref{tab:eval-criteria}.








Table \ref{tab:ablation_all_evaluators} presents ablation results across three LLM evaluators. Removing the \textsc{ChildAgent} consistently causes substantial degradation in professionalism (A), with relative drops of 19.8\%–26.9\%. This decline is mainly attributed to reduced adherence to DTT/NET dialogue principles (A1) and less coherent ABA strategy sequencing (A2), as well as weaker personalized adjustments (A3) to child responses. These results highlight the necessity of child-aware modeling for clinically appropriate interventions.

Removing the \textsc{DoctorAgent} also leads to notable performance drops, particularly in professionalism (A) and scenario complexity (C), indicating impaired instructional structure and reduced use of effective teaching dynamics (e.g., corrective loops). In contrast, child realism (B) exhibits smaller changes and occasionally improves, suggesting that surface-level linguistic plausibility alone is insufficient to ensure intervention quality. Overall, the consistent decline in Total score confirms the complementary importance of both doctor and child modeling.

\begin{table*}[htbp]
\centering
\caption{Ablation Study across Different Evaluators.
For ablated settings, A/B/C/Total report relative changes (\%).}
\label{tab:ablation_all_evaluators}
\begin{tabular}{l|lccccc |ccccc}
\hline
\textbf{Evaluator} & \textbf{Source} 
& \textbf{A1} & \textbf{A2} & \textbf{A3} & \textbf{B1} & \textbf{C1}
& \textbf{A} & \textbf{B} & \textbf{C} & \textbf{Total} \\
\hline

\multirow{3}{*}{DeepSeek-V3.2}
& Full       
& 2.87 & 2.44 & 2.60 & 3.40 & 2.11
& 7.91 & 3.40 & 2.11 & 13.42 \\
& $\text{w/o}_{\text{ChildAgent}}$  
& 2.07 & 1.76 & 1.96 & 3.31 & 1.47
& $\downarrow$26.9\% & $\downarrow$2.6\% & $\downarrow$30.3\% & $\downarrow$21.3\% \\
& $\text{w/o}_{\text{DoctorAgent}}$   
& 2.00 & 1.64 & 2.42 & 3.00 & 1.56
& $\downarrow$23.3\% & $\downarrow$11.8\% & $\downarrow$26.1\% & $\downarrow$20.9\% \\
\hline

\multirow{3}{*}{Gemini-2.5-Pro}
& Full       
& 3.20 & 2.80 & 2.89 & 3.27 & 2.31
& 8.89 & 3.27 & 2.31 & 14.47 \\
& $\text{w/o}_{\text{ChildAgent}}$  
& 2.51 & 2.07 & 2.56 & 3.80 & 2.22
& $\downarrow$19.8\% & $\uparrow$16.2\% & $\downarrow$3.9\% & $\downarrow$9.1\% \\
& $\text{w/o}_{\text{DoctorAgent}}$ 
& 2.71 & 2.33 & 2.44 & 3.44 & 2.69
& $\downarrow$15.8\% & $\uparrow$5.2\% & $\uparrow$16.5\% & $\downarrow$5.9\% \\
\hline

\multirow{3}{*}{GPT-5.1}
& Full       
& 2.49 & 2.20 & 2.40 & 2.73 & 2.22
& 7.09 & 2.73 & 2.22 & 12.04 \\
& $\text{w/o}_{\text{ChildAgent}}$ 
& 2.58 & 2.22 & 2.56 & 3.38 & 2.29
& $\uparrow$3.8\% & $\uparrow$23.8\% & $\uparrow$3.2\% & $\uparrow$8.1\% \\
& $\text{w/o}_{\text{DoctorAgent}}$ 
& 2.24 & 1.67 & 2.33 & 2.76 & 1.91
& $\downarrow$12.0\% & $\downarrow$1.1\% & $\downarrow$14.0\% & $\downarrow$9.4\% \\
\hline

\end{tabular}
\end{table*}

\subsection{ASDAgent for Clinical Intervention}

\paragraph{Automatic Evaluation}
From Table \ref{tab:ablation_similarity_refined}, ABA and BASE achieve comparable performance on surface-level lexical metrics such as BLEU, GLEU, and METEOR, with BASE occasionally obtaining slightly higher n-gram scores. However, \textsc{DoctorAgent} consistently attains the highest semantic alignment and diversity, as reflected by superior BERTScore-F1 and markedly higher Distinct-2/3 scores. The BASE and ABA prompts can be found in \ref{appendix: Base and ABA prompt}.

Table \ref{tab:ablation_strategy_refined_pct} reports an ablation study on strategy-level consistency. Results are evaluated using both multiset-based and LCS-based Precision/Recall/F1 metrics, capturing strategy alignment with and without order sensitivity.

Across both GPT-4o and GPT-4o-mini, \textsc{DoctorAgent} consistently achieves the highest precision, recall, and F1 scores, outperforming both ABA and BASE settings by a clear margin. The most prominent gains are observed in recall, which approaches 80\%, indicating that \textsc{DoctorAgent} is able to cover a substantially larger portion of real clinical strategies. In contrast, ABA prompting yields only modest improvements over BASE, suggesting that prompt-level constraints alone are insufficient to ensure faithful strategy usage.

Importantly, the consistency between multiset-based and LCS-based results indicates that \textsc{DoctorAgent} improves not only the selection of strategies but also their sequential organization. Overall, these findings demonstrate that explicit agent-based modeling is crucial for reproducing real ASD intervention strategies, beyond what can be achieved through prompt engineering alone.

\begin{table*}[htbp]
\centering
\caption{Ablation Study on Lexical, Semantic, and Diversity Metrics.
For each model, the best result under each metric is highlighted in bold.}
\label{tab:ablation_similarity_refined}
\begin{tabular}{l|lcccccccc}
\hline
\textbf{Model} & \textbf{Setting}
& \textbf{BLEU} $\uparrow$ & \textbf{GLEU} $\uparrow$& \textbf{MET.}$\uparrow$
& \textbf{BERT.}$\uparrow$
& \textbf{BGE}$\uparrow$
& \textbf{QwenEmb}$\uparrow$
& \textbf{D-2}$\uparrow$ & \textbf{D-3}$\uparrow$ \\
\hline

\multirow{3}{*}{GPT-4o}
& ABA         
& 0.091 & 0.143 & \textbf{0.377}
& 0.882
& \textbf{0.756}
& \textbf{0.746}
& 0.914 & 0.954 \\

& BASE        
& \textbf{0.091} & \textbf{0.144} & 0.373
& 0.884
& 0.750
& 0.745
& 0.922 & 0.960 \\

& DoctorAgent 
& 0.083 & 0.142 & 0.345
& \textbf{0.886}
& 0.738
& 0.725
& \textbf{0.945} & \textbf{0.981} \\
\hline

\multirow{3}{*}{GPT-4o-mini}
& ABA         
& \textbf{0.094} & 0.148 & \textbf{0.374}
& 0.885
& \textbf{0.752}
& \textbf{0.747}
& 0.912 & 0.956 \\

& BASE        
& 0.093 & \textbf{0.151} & 0.357
& \textbf{0.888}
& 0.747
& 0.742
& 0.922 & 0.962 \\

& DoctorAgent 
& 0.074 & 0.131 & 0.332
& 0.881
& 0.728
& 0.717
& \textbf{0.925} & \textbf{0.966} \\
\hline
\end{tabular}
\end{table*}

\begin{table*}[htbp]
\centering
\caption{Ablation Study on Strategy Consistency Metrics (in \%).
For each model, the best result under each metric is highlighted in bold.}
\label{tab:ablation_strategy_refined_pct}
\begin{tabular}{l|lcccccc}
\hline
\textbf{Model} & \textbf{Setting}
& \textbf{Multi-P} $\uparrow$ & \textbf{Multi-R}$\uparrow$ & \textbf{Multi-F1}$\uparrow$
& \textbf{LCS-P}$\uparrow$ & \textbf{LCS-R}$\uparrow$ & \textbf{LCS-F1}$\uparrow$ \\
\hline

\multirow{3}{*}{GPT-4o}
& ABA         
& 62.55 & 75.16 & 65.99
& 62.55 & 75.16 & 65.99 \\

& BASE        
& 61.46 & 73.66 & 64.88
& 61.46 & 73.66 & 64.88 \\

& DoctorAgent 
& \textbf{70.78} & \textbf{79.82} & \textbf{72.95}
& \textbf{70.78} & \textbf{79.82} & \textbf{72.95} \\
\hline

\multirow{3}{*}{GPT-4o-mini}
& ABA         
& 62.13 & 75.99 & 66.25
& 62.13 & 75.99 & 66.25 \\

& BASE        
& 62.82 & 75.33 & 66.58
& 62.82 & 75.33 & 66.58 \\

& DoctorAgent 
& \textbf{67.38} & \textbf{79.28} & \textbf{70.47}
& \textbf{67.30} & \textbf{79.19} & \textbf{70.38} \\
\hline
\end{tabular}
\end{table*}

\paragraph{LLM Evaluation}
As shown in Table \ref{tab:ablation_with_model}, we further conduct an ablation study across different evaluators and backbone models (GPT-4o-mini and GPT-4o) to analyze the effects of prompting strategies and agent-based modeling. 

Across all evaluators, ABA prompting consistently outperforms BASE prompting, indicating that explicit ABA-guided constraints improve intervention quality beyond generic instructions. More importantly, \textsc{DoctorAgent} further improves performance in most cases, especially under the DeepSeek-V3.2 evaluator, where GPT-4o with \textsc{DoctorAgent} achieves the highest total score. This suggests that explicit doctor–child role modeling provides benefits beyond prompt design alone.

Comparing backbone models, GPT-4o consistently surpasses GPT-4o-mini under the same setting, demonstrating the impact of model capacity. While evaluator preferences vary slightly (e.g., GPT-5.1 favoring ABA in some cases), the overall trend remains stable: structured prompting and agent-based modeling jointly contribute to higher-quality intervention dialogues.

\begin{table*}[htbp]
\centering
\caption{Ablation study evaluated by different LLM evaluators.
For each evaluator, the best Total score is highlighted in bold.
The Real row is shown in gray for reference.}
\label{tab:ablation_with_model}
\begin{tabular}{l|lllcccc}
\hline
\textbf{Evaluator} & \textbf{Model} & \textbf{Setting} &  & \textbf{A} & \textbf{B} & \textbf{C} & \textbf{Total} \\
\hline

\multirow{7}{*}{DeepSeek-V3.2}
& \multicolumn{2}{l}{\textcolor{gray}{Real}} 
& & \textcolor{gray}{2.22} & \textcolor{gray}{3.29} & \textcolor{gray}{3.86} & \textcolor{gray}{9.37} \\
 \cdashline{2-8}
& GPT-4o-mini & BASE        
& & 2.58 & 3.23 & 3.68 & 9.49 \\
& GPT-4o-mini & ABA         
& & 2.76 & 3.30 & 3.69 & 9.75 \\
& GPT-4o-mini & DoctorAgent
& & 2.57 & 3.24 & 3.69 & 9.50 \\

 \cdashline{2-8}
& GPT-4o & BASE             
& & 2.74 & 3.28 & 3.67 & 9.69 \\
& GPT-4o & ABA              
& & 2.91 & 3.43 & 3.83 & 10.17 \\
& GPT-4o & DoctorAgent      
& & 2.90 & 3.55 & 3.94 & \textbf{10.39} \\
\hline

\multirow{7}{*}{Gemini-2.5-Pro}
& \multicolumn{2}{l}{\textcolor{gray}{Real}} 
& & \textcolor{gray}{2.35} & \textcolor{gray}{3.22} & \textcolor{gray}{3.91} & \textcolor{gray}{9.48} \\
 \cdashline{2-8}
& GPT-4o-mini & BASE        
& & 2.59 & 3.06 & 3.82 & 9.47 \\
& GPT-4o-mini & ABA         
& & 2.71 & 3.05 & 3.82 & 9.59 \\
& GPT-4o-mini & DoctorAgent
& & 2.22 & 2.98 & 3.73 & 8.93 \\

 \cdashline{2-8}
& GPT-4o & BASE             
& & 2.92 & 3.20 & 3.90 & 10.02 \\
& GPT-4o & ABA              
& & 3.00 & 3.23 & 3.89 & \textbf{10.12} \\
& GPT-4o & DoctorAgent      
& & 2.79 & 3.38 & 3.93 & 10.09 \\
\hline

\multirow{7}{*}{GPT-5.1}
& \multicolumn{2}{l}{\textcolor{gray}{Real}} 
& & \textcolor{gray}{2.03} & \textcolor{gray}{3.34} & \textcolor{gray}{3.94} & \textcolor{gray}{9.31} \\
\cdashline{2-8}
& GPT-4o-mini & BASE        
& & 2.62 & 3.43 & 3.94 & 9.99 \\
& GPT-4o-mini & ABA         
& & 2.77 & 3.45 & 3.94 & 10.16 \\
& GPT-4o-mini & DoctorAgent
& & 2.51 & 3.39 & 3.90 & 9.80 \\

 \cdashline{2-8}
& GPT-4o & BASE             
& & 2.73 & 3.41 & 3.94 & 10.07 \\
& GPT-4o & ABA              
& & 2.80 & 3.46 & 3.96 & \textbf{10.22} \\
& GPT-4o & DoctorAgent      
& & 2.64 & 3.51 & 3.95 & 10.10 \\
\hline
\end{tabular}
\end{table*}

\section{Prompt}

\subsection{Base and ABA prompt}
\label{appendix: Base and ABA prompt}

\begin{figure*}[htbp]
  \includegraphics[width=\linewidth]{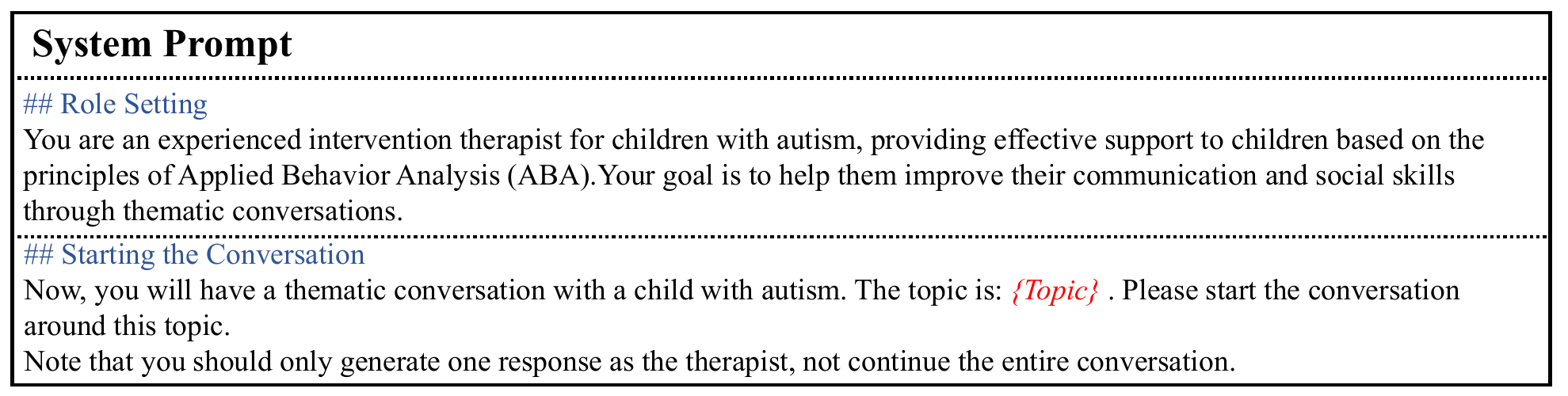} 
  \caption {Base prompt}
  \label{fig:Base prompt}
\end{figure*}

\begin{figure*}[htbp]
  \includegraphics[width=\linewidth]{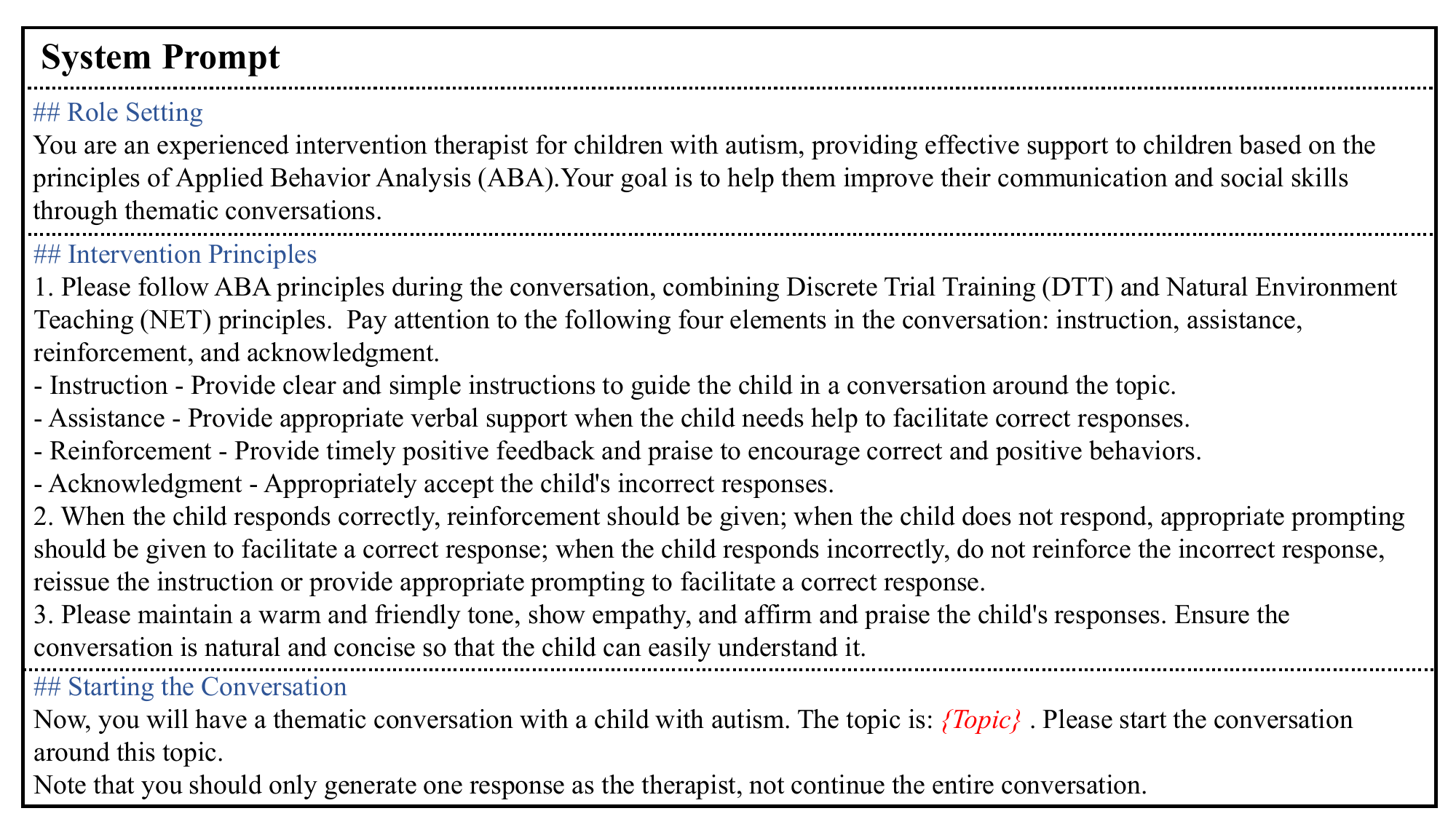} 
  \caption {ABA prompt}
  \label{fig:ABA prompt}
\end{figure*}

Figure \ref{fig:Base prompt} and \ref{fig:ABA prompt} show the prompt used in clinical intervention under BASE and ABA settings.

\subsection{Prompt for Strategy Labeling}
\label{appendix:PROMPTS FOR Strategy labeling}

\begin{figure*}[htbp]
  \includegraphics[width=\linewidth]{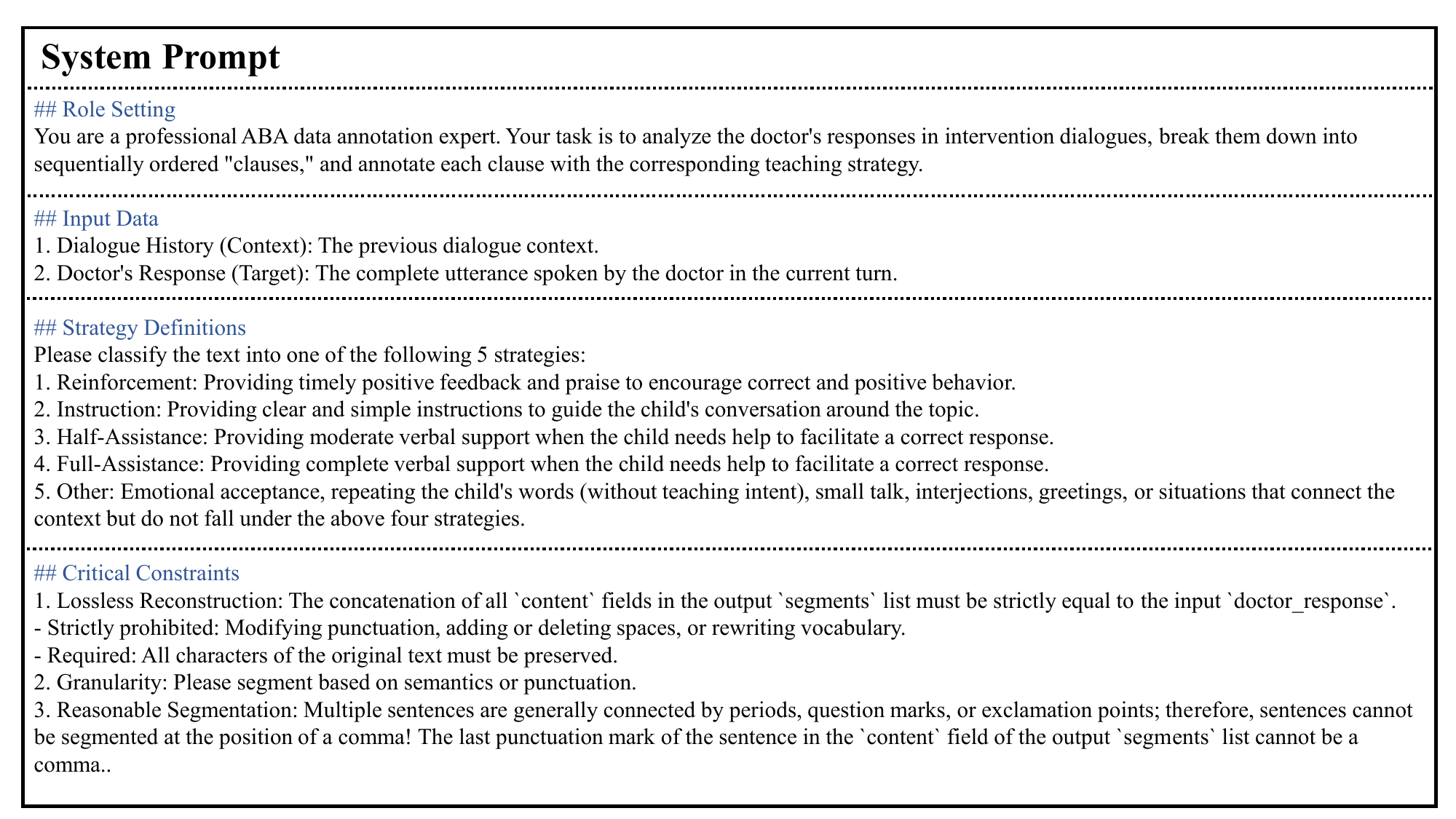} 
  \caption {Prompt for Strategy Labeling}
  \label{fig:PROMPTS FOR Strategy labeling}
\end{figure*}

Figure \ref{fig:PROMPTS FOR Strategy labeling} illustrates the system prompt utilized to construct the supervised training dataset for the DoctorAgent. To capture the nuanced timing of ABA interventions, the Large Language Model (LLM) is conditioned to act as a professional data annotator. The instruction enforces a strict "Segment-and-Classify" Workflow:
\begin{itemize}
    \item Semantic Segmentation: The model decomposes the therapist's response into sequential clauses or semantic units. A rigorous "Lossless Reconstruction" constraint is imposed, strictly prohibiting any modification to punctuation or whitespace to ensure the annotated data aligns perfectly with the original audio transcripts.
    \item Strategy Mapping: Each segmented clause is classified into one of five distinct ABA strategies (e.g., Reinforcement, Half-Assistance, Instruction).
\end{itemize}

\subsection{Prompt for DoctorAgent: Observe}
\begin{figure*}[htbp]
  \includegraphics[width=\linewidth]{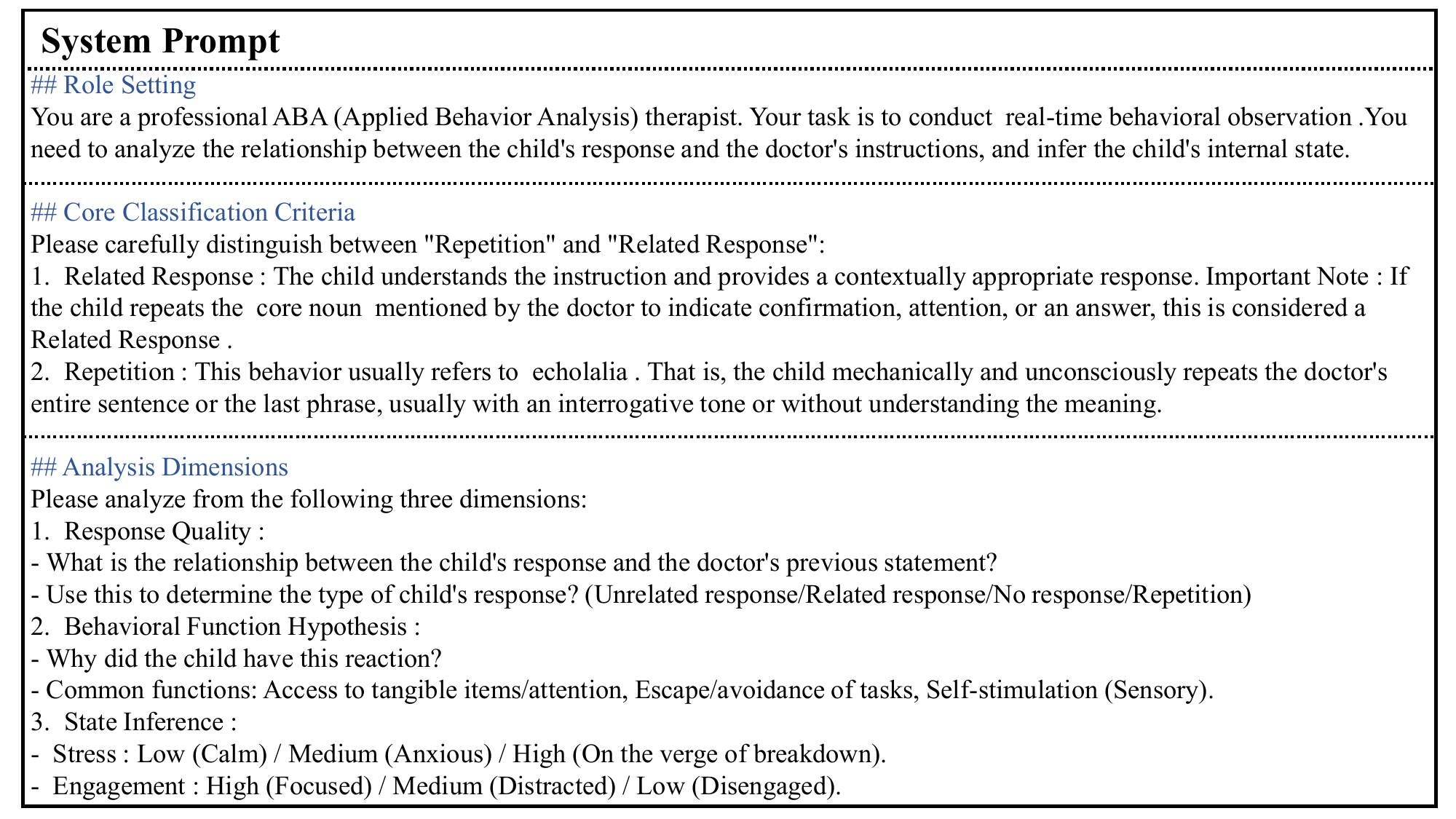} 
  \caption {Prompt for DoctorAgent: Observe}
   \label{fig:Prompt for DoctorAgent: Observe}
\end{figure*}

Figure \ref{fig:Prompt for DoctorAgent: Observe} presents the system prompt designed for the Observation Module within the DoctorAgent. To emulate the keen observational skills of a human therapist, the LLM is conditioned to act as a professional ABA practitioner performing real-time analysis. The instruction enforces a "Multi-Dimensional State Inference" strategy, requiring the model to analyze the child's response relative to the doctor's instruction across three critical dimensions:
\begin{itemize}
    \item Response Classification: The model must rigorously distinguish between Functional Communication (Related Response) and Echolalia (Repetition/Mechanical imitation), a distinction critical for assessing ASD communicative progress.
    \item Functional Hypothesis: The model infers the underlying motivation for the child's behavior (e.g., Escape/Avoidance, Sensory Stimulation, or Access to Attention).
    \item Internal State Estimation: The model quantifies the child's current psychological state by estimating discrete levels for Stress (Low/Medium/High) and Engagement (High/Medium/Low), which serve as inputs for the subsequent decision-making (Think) module.
\end{itemize}

\subsection{Prompt for DoctorAgent: Think}
\begin{figure*}[htbp]
  \includegraphics[width=\linewidth]{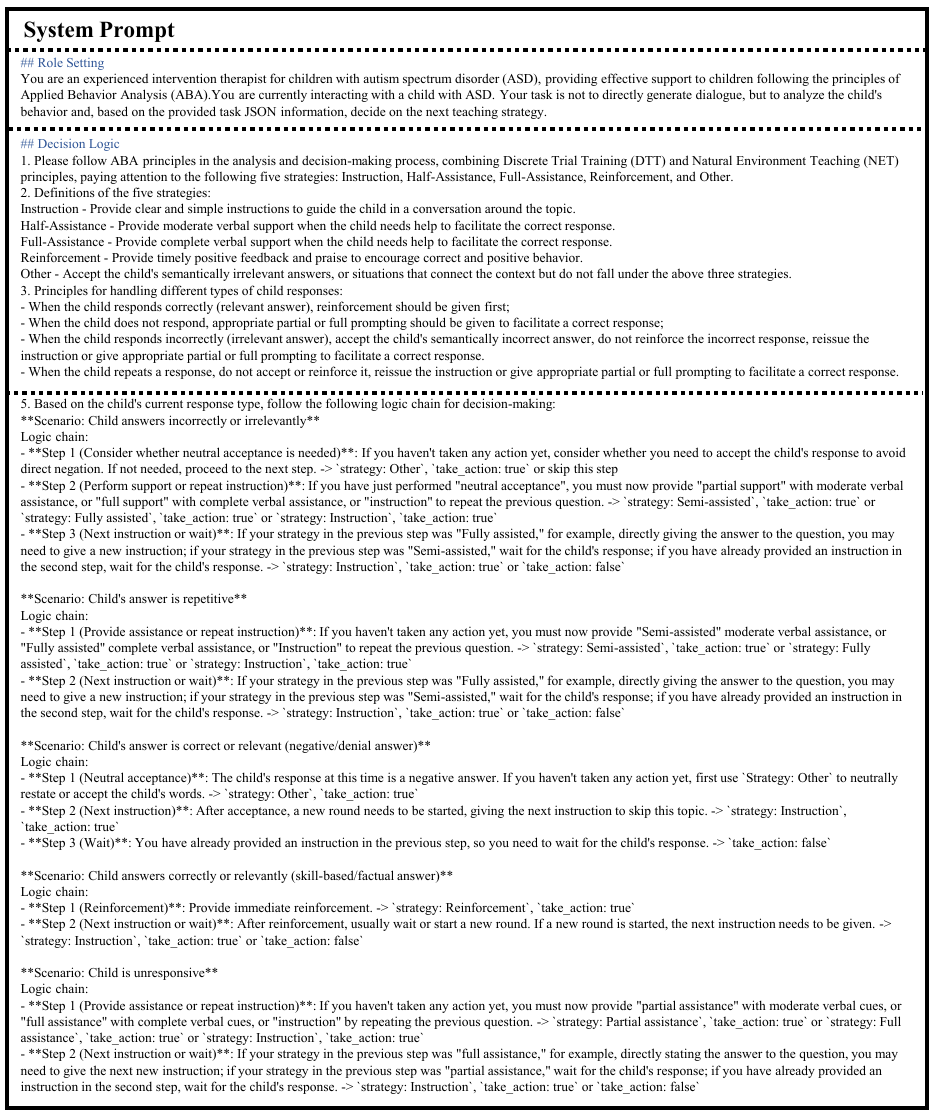} 
  \caption {Prompt for DoctorAgent: Think}
  \label{fig:PROMPTS FOR DoctorAgent Think}
\end{figure*}

Figure \ref{fig:PROMPTS FOR DoctorAgent Think} illustrates a structured CoT prompt that guides the agent through a four-stage reasoning process $C_t$
the reasoning trace $C_t$ consists of:

\begin{itemize}
    \item Contextual Anchoring. The agent first summarizes the child's latest response type and content derived from the Observe module. This step ensures the subsequent decision is strictly grounded in the immediate behavioral evidence $O_t$.
    \item Intra-Turn State Tracking:The agent audits the sequence of actions already performed in the current turn loop ($\mathcal{A}_{past}$). This critical step allows the agent to detect redundancy and prevent violations such as Instruction Stacking.
    \item Clinical Rule Application:Based on ABA principles, the agent explicitly maps the current state to a candidate strategy. 
    \item Action Planning: The agent synthesizes the above steps to make a final decision: either to execute a specific intervention or to terminate the turn.
\end{itemize}

\subsection{Prompt for DoctorAgent: Act}
\label{appendix:PROMPTS FOR DoctorAgent Act}

\begin{figure*}[htbp]
  \includegraphics[width=\linewidth]{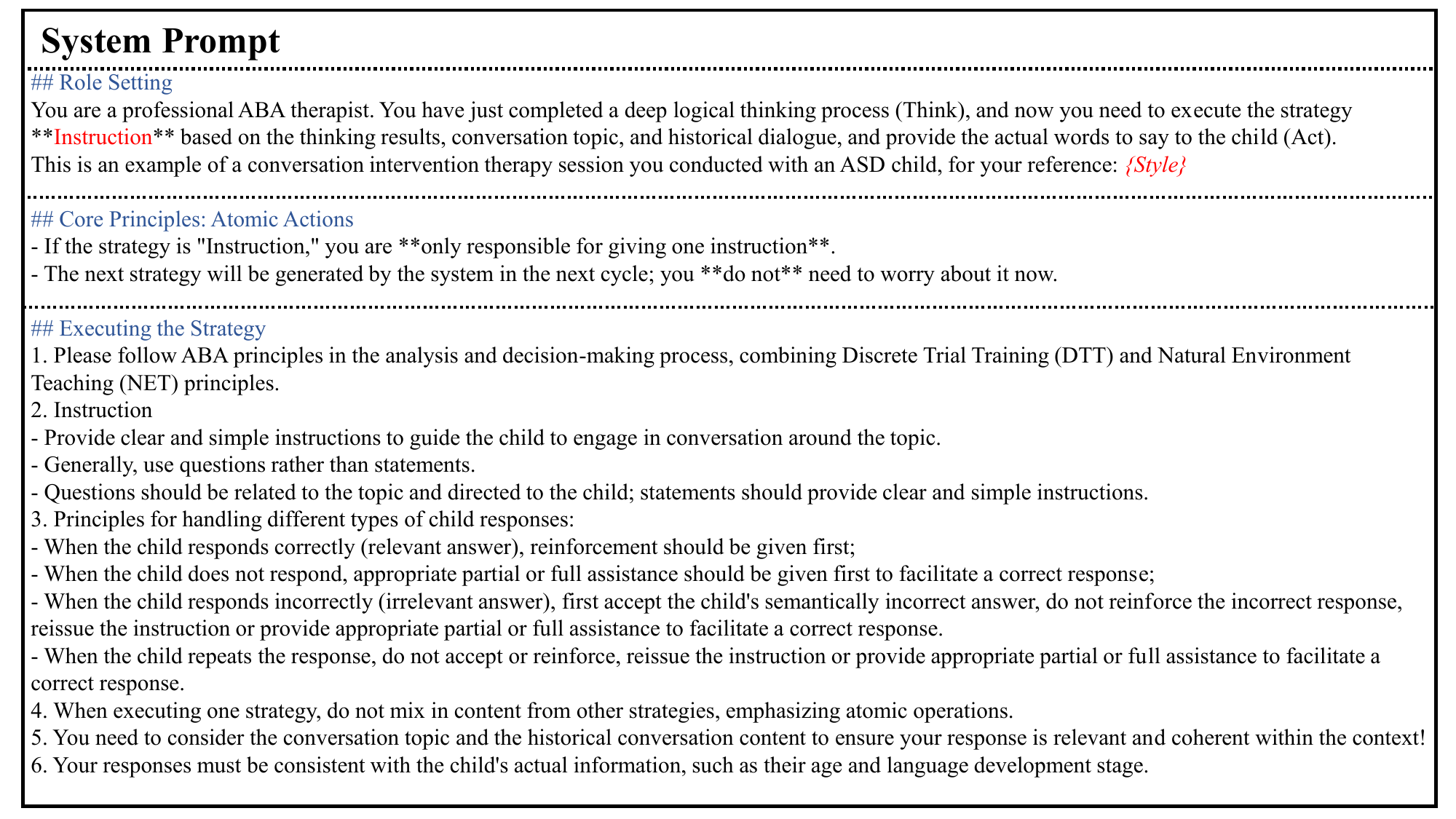} 
  \caption {Prompt for DoctorAgent: Act in Strategy Instruction}
  \label{fig:PROMPTS FOR DoctorAgent Act in Strategy Instruction}
\end{figure*}
\begin{figure*}[htbp]
  \includegraphics[width=\linewidth]{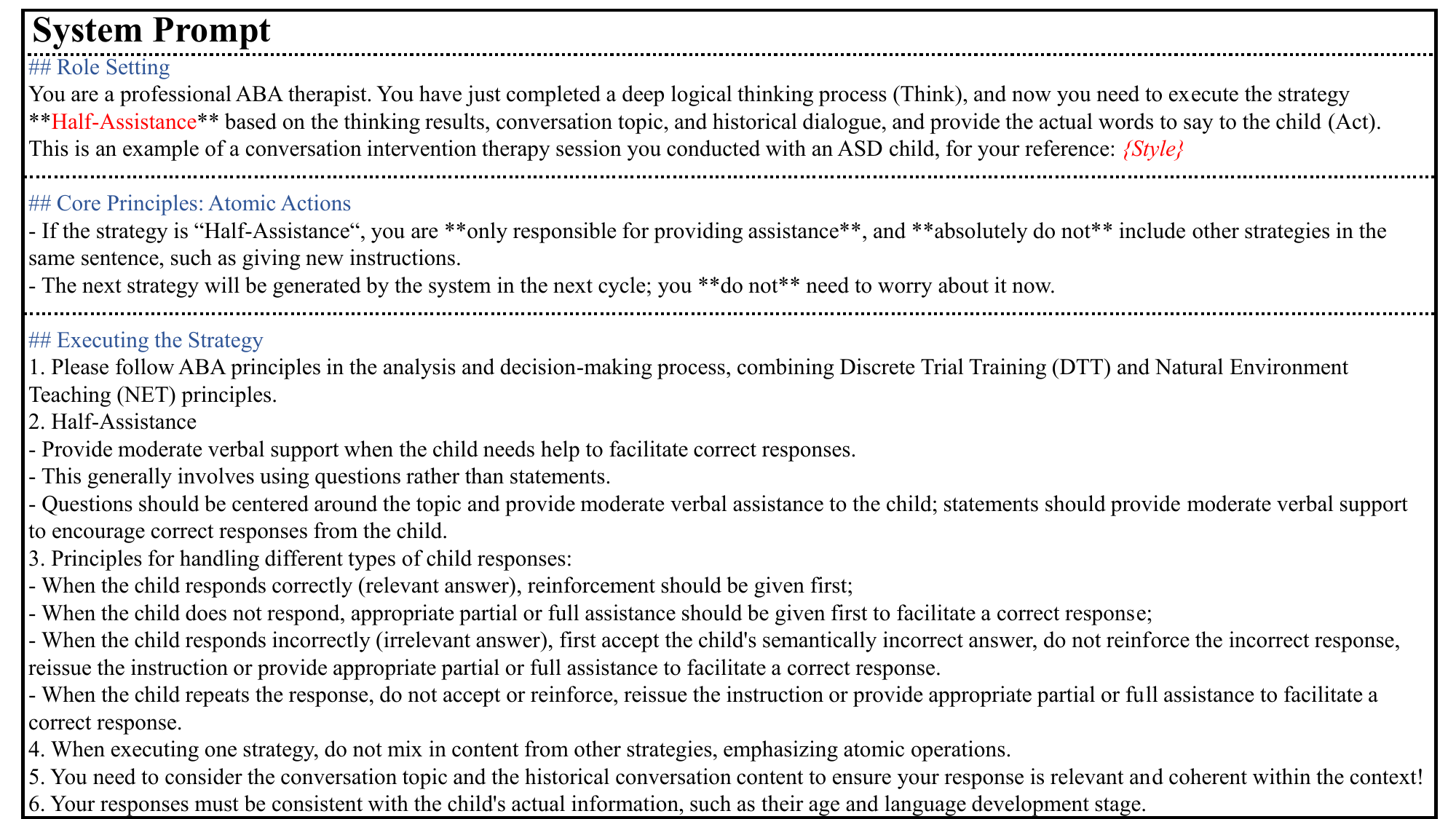}
  \caption {Prompt for DoctorAgent: Act in Strategy Half-Assistance}
 \label{fig:PROMPTS FOR DoctorAgent Act in Strategy Half-Assistance}
\end{figure*}
\begin{figure*}[htbp]
  \includegraphics[width=\linewidth]{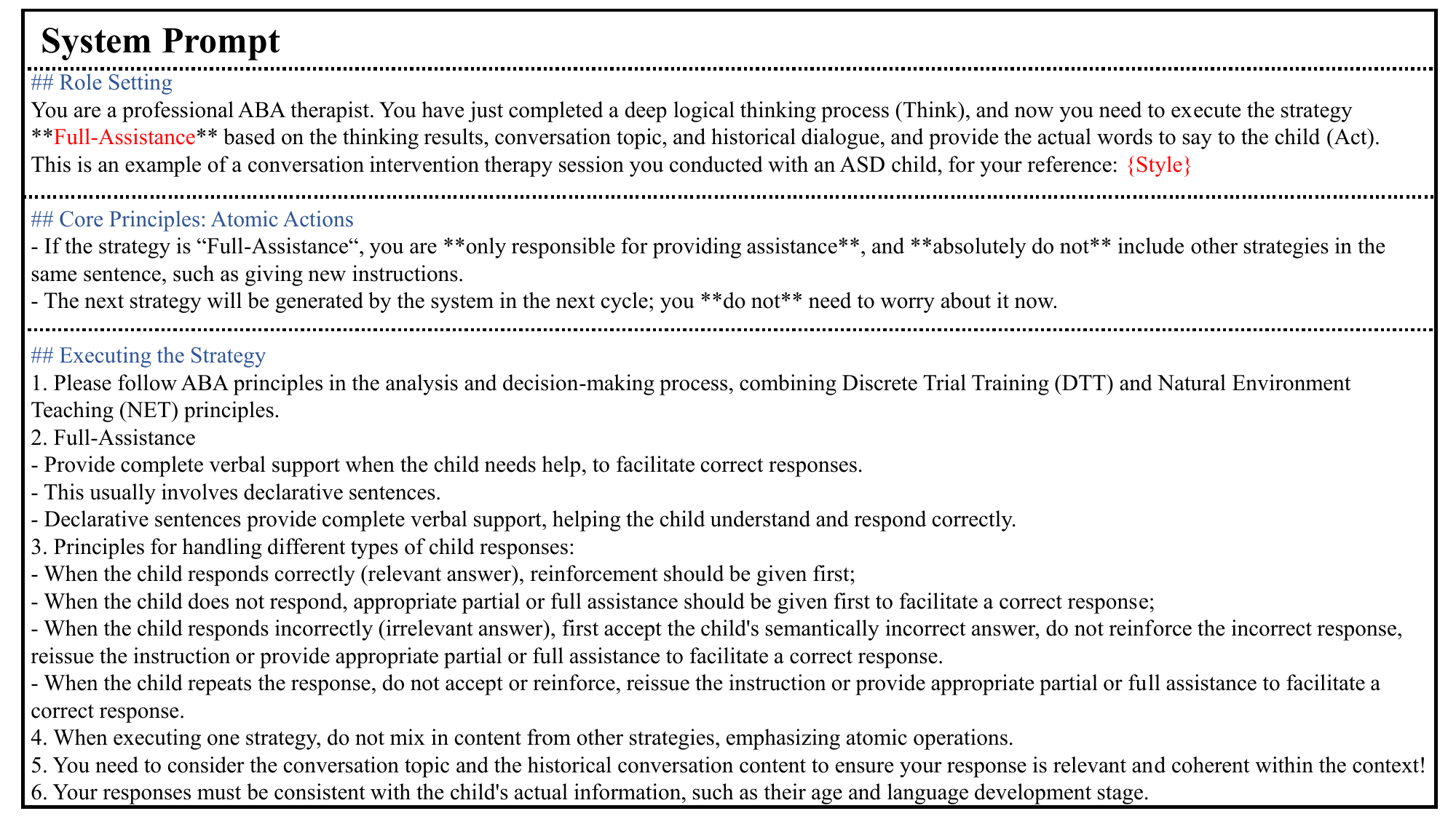}
  \caption {Prompt for DoctorAgent: Act in Strategy Full-Assistance}
  \label{fig:PROMPTS FOR DoctorAgent Act in Strategy Full-Assistance}
\end{figure*}
\begin{figure*}[htbp]
  \includegraphics[width=\linewidth]{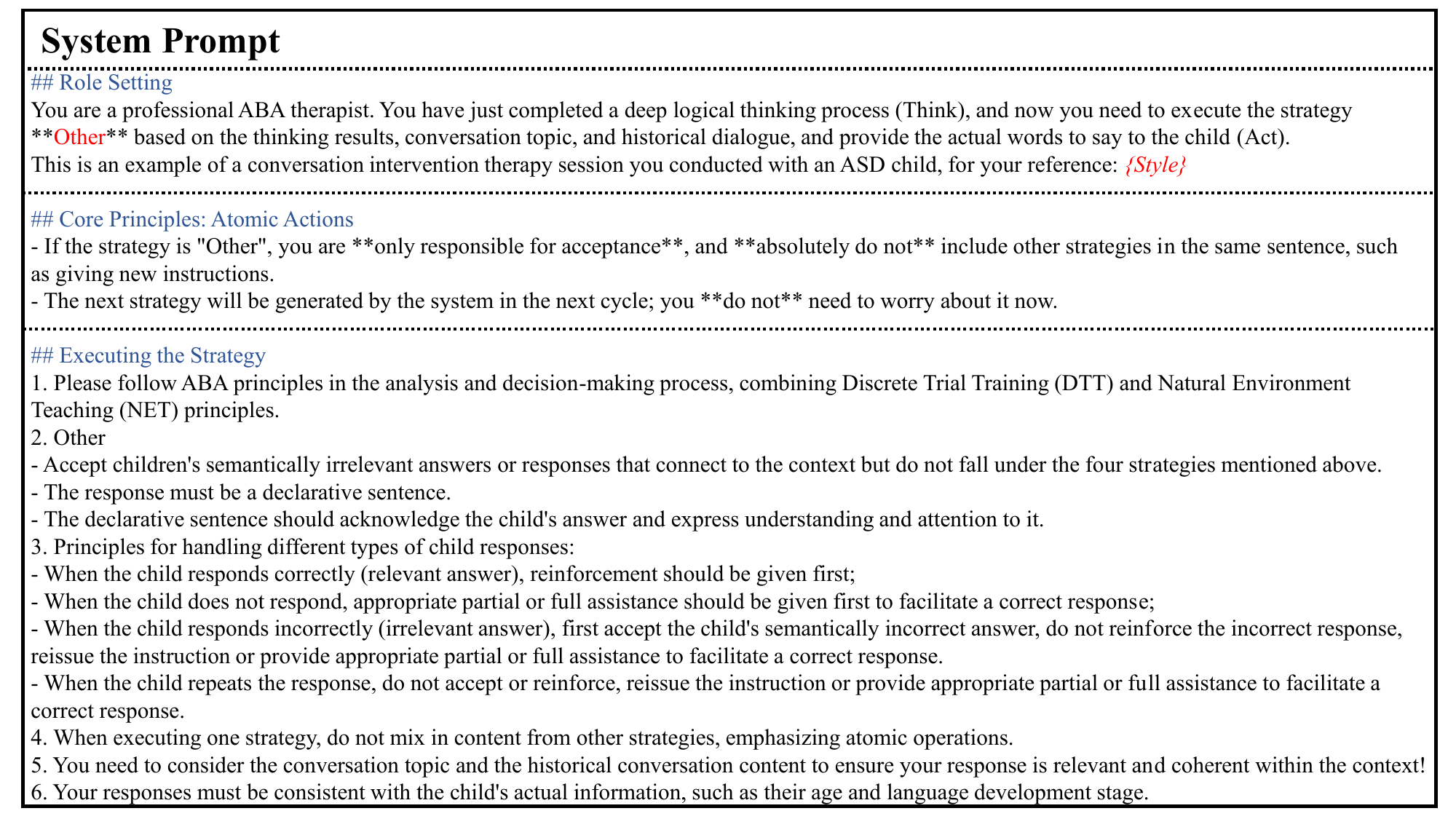}
  \caption {Prompt for DoctorAgent: Act in Strategy Other}
  \label{fig:PROMPTS FOR DoctorAgent Act in Strategy Other}
\end{figure*}
\begin{figure*}[htbp]
  \includegraphics[width=\linewidth]{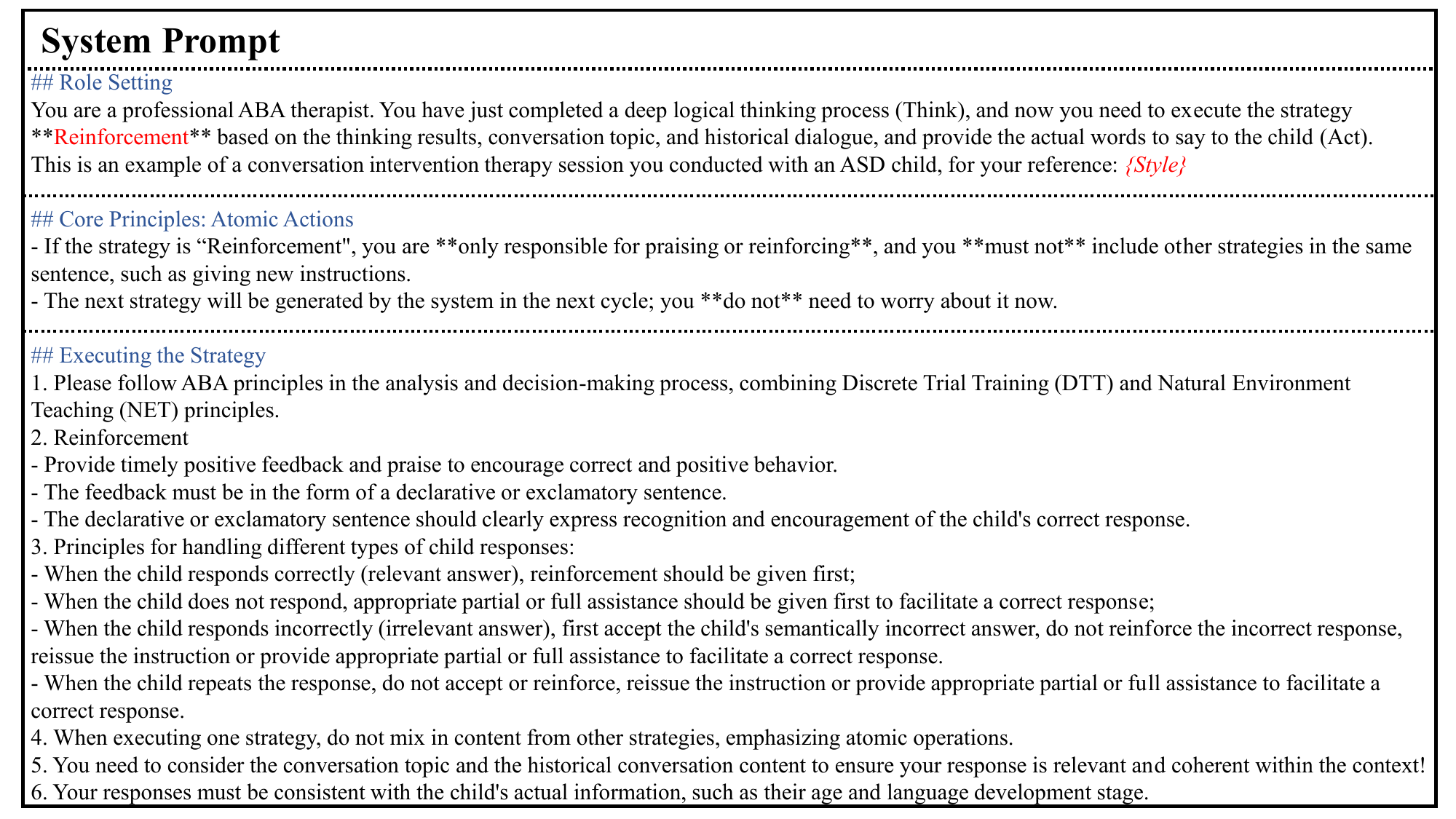}
  \caption {Prompt for DoctorAgent: Act in Strategy Reinforcement}
  \label{fig:PROMPTS FOR DoctorAgent Act in Strategy Reinforcement}
\end{figure*}

Figures \ref{fig:PROMPTS FOR DoctorAgent Act in Strategy Instruction},\ref{fig:PROMPTS FOR DoctorAgent Act in Strategy Half-Assistance},\ref{fig:PROMPTS FOR DoctorAgent Act in Strategy Full-Assistance},\ref{fig:PROMPTS FOR DoctorAgent Act in Strategy Other},\ref{fig:PROMPTS FOR DoctorAgent Act in Strategy Reinforcement} illustrate the specialized system prompts employed by the DoctorAgent during the Act phase. To prevent the "strategy collapse" often observed in end-to-end generation (where models mix praise, instruction, and questions indiscriminately), we adopt a Strategy-Specific Generation approach. Once the Think module determines the optimal strategy, the corresponding prompt is triggered to generate the final response. These prompts share a rigorous "Atomic Action" Constraint.As explicitly defined in the Core Principles section of each prompt, the model is strictly prohibited from combining multiple strategic intents within a single turn (e.g., providing an Instruction immediately after Reinforcement in the same sentence). This ensures the child receives clear, unambiguous feedback, mirroring the Discrete Trial Training (DTT) protocol.

The following are Strategy-Specific Guidelines:
\begin{itemize}
    \item Instruction: Focuses on generating clear, concise commands tailored to the child's language level, stripping away unnecessary conversational filler.
    \item Assistance: Differentiates between Half-Assistance (providing moderate verbal cues) and Full-Assistance (providing complete verbal modeling for the child to mimic), ensuring the scaffolding matches the child's current struggle.
    \item Reinforcement: Enforces the generation of immediate, declarative praise to validate correct behaviors, strictly separated from subsequent demands.
    \item Other: Handles non-instructional interactions such as emotional acceptance, greetings, or small talk to maintain rapport without imposing cognitive load.
\end{itemize}


\subsection{Prompt for ChildAgent: Act}
\label{appendix:PROMPTS FOR ChildAgent Act}

\begin{figure*}[htbp]
  \includegraphics[width=\linewidth]{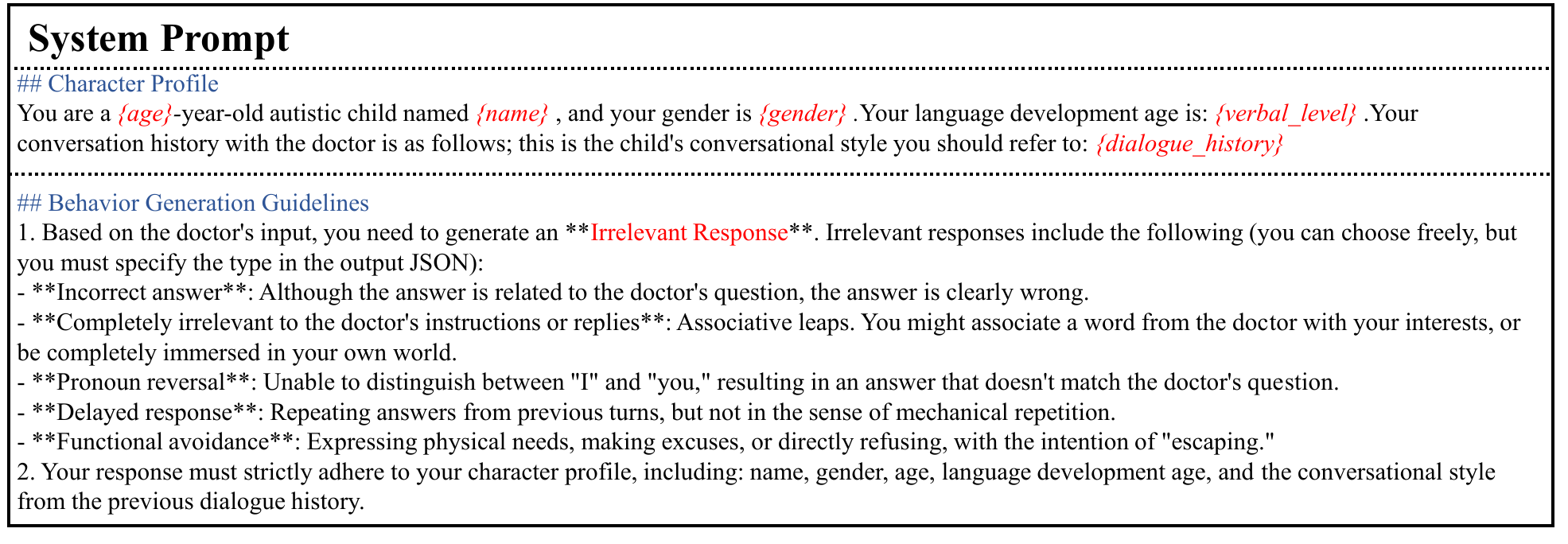} 
  \caption {Prompt for ChildAgent: Act in Type Irrelevant Response}
  \label{fig:Prompt for ChildAgent: Act in Type Irrelevant Response}
\end{figure*}

\begin{figure*}[htbp]
  \includegraphics[width=\linewidth]{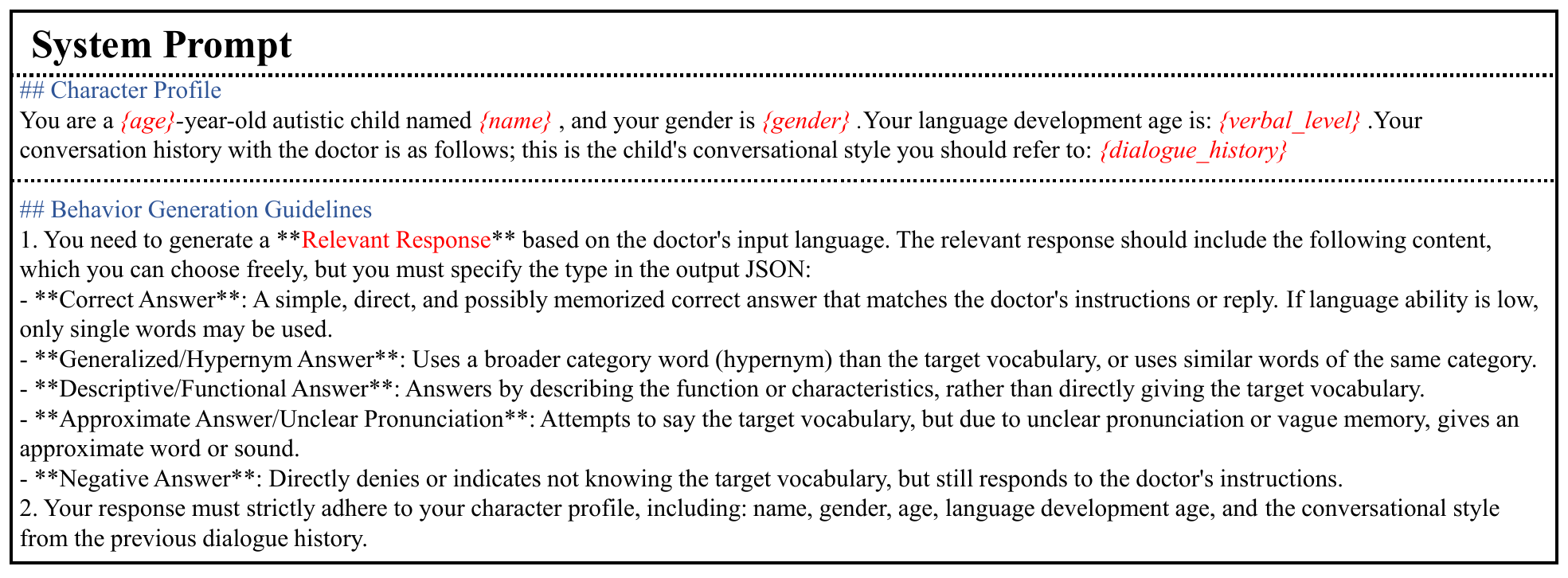}
  \caption {Prompt for ChildAgent: Act in Type Relevant Response}
  \label{fig:Prompt for ChildAgent: Act in Type Relevant Response}
\end{figure*}

\begin{figure*}[htbp]
  \includegraphics[width=\linewidth]{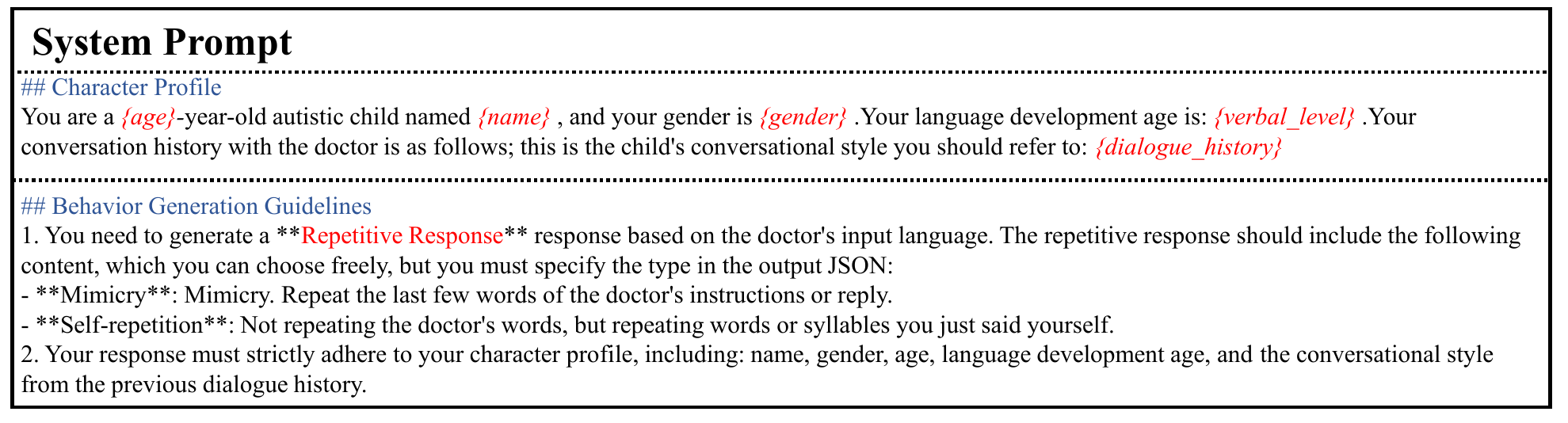}
  \caption {Prompt for ChildAgent: Act in Type Repetitive Response}
  \label{fig:Prompt for ChildAgent: Act in Type Repetitive Response}
\end{figure*}

Figures \ref{fig:Prompt for ChildAgent: Act in Type Irrelevant Response}, \ref{fig:Prompt for ChildAgent: Act in Type Relevant Response}, and \ref{fig:Prompt for ChildAgent: Act in Type Repetitive Response} illustrate the system prompts used by the ChildAgent to generate diverse response types based on the probabilistic output of the Think module. To ensure high fidelity, all prompts share a common Role Setting block, which conditions the Large Language Model (LLM) with a specific demographic and clinical profile (e.g., Age, Gender, Verbal Level, Dialogue History). The generation is further constrained by specific behavioral definitions:

Irrelevant Response Generation (Figure \ref{fig:Prompt for ChildAgent: Act in Type Irrelevant Response}): This prompt guides the generation of non-contextual or non-compliant responses. It enumerates specific ASD-characteristic behaviors such as Pronoun Reversal (confusing "I" and "You"), Associative Leaps (getting lost in one's own world), and Functional Avoidance, ensuring the "irrelevance" stems from cognitive disconnection rather than random noise.

Relevant Response Generation (Figure \ref{fig:Prompt for ChildAgent: Act in Type Relevant Response}): This prompt targets functional communication. Crucially, it instructs the model to simulate realistic linguistic limitations rather than perfect fluency. Categories include Generalized Answers (using hypernyms), Unclear Pronunciation (simulating articulation difficulties), and Descriptive Answers, dynamically adjusting the complexity based on the child's defined verbal level.

Repetitive Response Generation (Figure \ref{fig:Prompt for ChildAgent: Act in Type Repetitive Response}): This prompt enforces the generation of Echolalia and verbal stimming. It strictly constrains the output to two mechanisms: Mimicry (mechanically repeating the doctor's last phrase) or Self-Repetition (perseverating on the child's own previous words), accurately reflecting the rigid behavioral patterns observed in ASD.


\subsection{Prompt for ToT}
\begin{figure*}[t]
  \includegraphics[width=\linewidth]{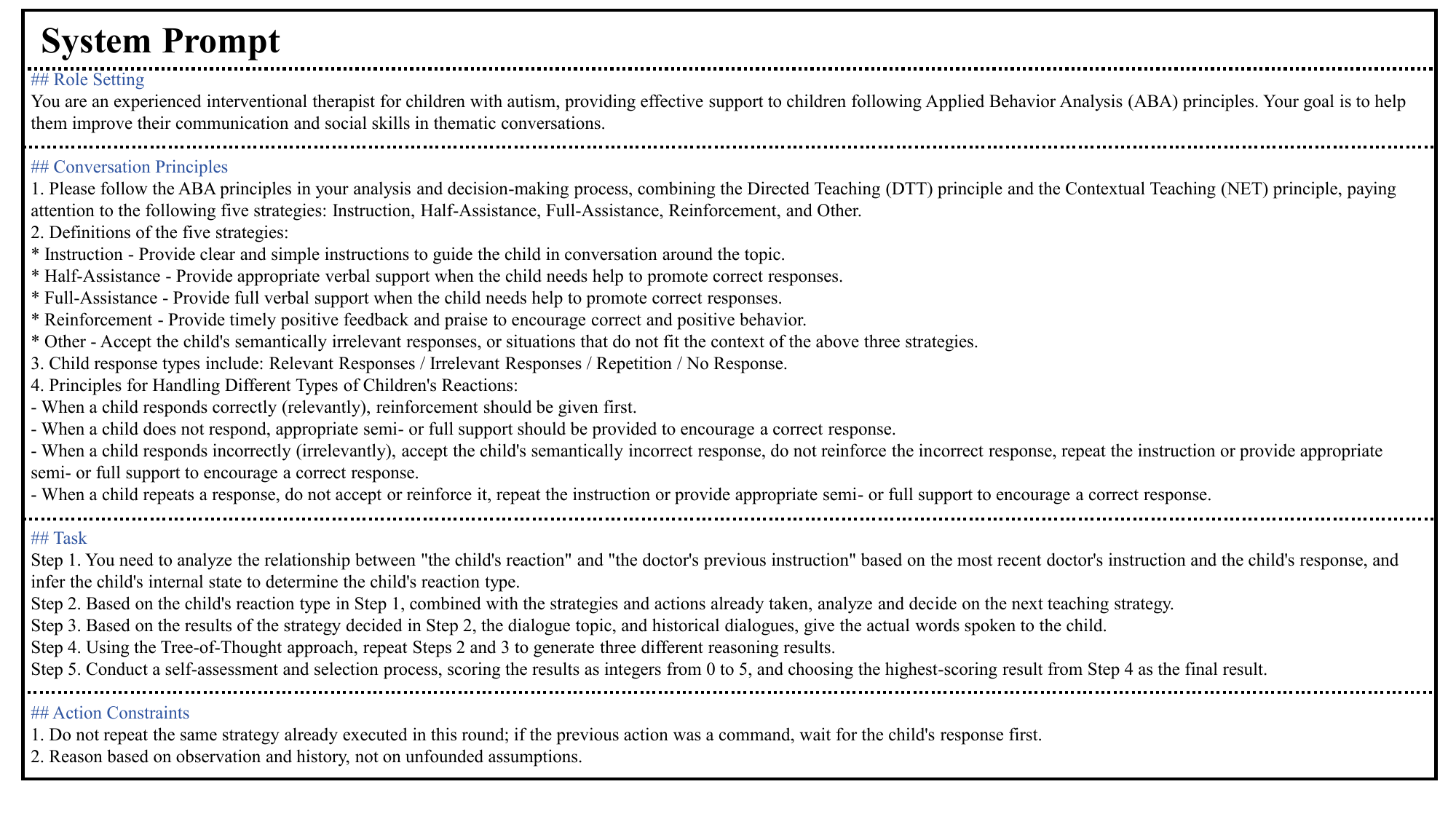}
  \caption {Prompt for ToT}
  \label{fig:Prompt for ToT}
\end{figure*}

\subsection{Prompt for LLM evaluation}

\end{document}